\documentclass{article}

\usepackage[margin=15mm]{geometry}
\usepackage{amssymb}
\usepackage{amsmath}
\usepackage{graphicx}
\usepackage[tight,footnotesize]{subfigure}
\usepackage{multirow}
\usepackage{color}
\usepackage{url}
\usepackage{colortbl}
\usepackage{rotating}
\usepackage{calc}

\begin{document}

\title{Information fusion in multi-task Gaussian processes}
\author{Shrihari Vasudevan, Arman Melkumyan and Steven Scheding}
\date{Australian Centre for Field Robotics, The University of Sydney, NSW 2006, Australia\\
  Email: shrihari.vasudevan@ieee.org $|$ a.melkumyan@acfr.usyd.edu.au $|$ s.scheding@acfr.usyd.edu.au}
\maketitle

\begin{abstract}
  This paper evaluates heterogeneous information fusion using multi-task Gaussian processes in the context of geological resource modeling. Specifically, it empirically demonstrates that information integration across heterogeneous information sources leads to superior estimates of all the quantities being modeled, compared to modeling them individually. Multi-task Gaussian processes provide a powerful approach for simultaneous modeling of multiple quantities of interest while taking correlations between these quantities into consideration. Experiments are performed on large scale real sensor data.\\
\end{abstract}

\section{Introduction}
\label{Sec:intro}
In applications such as space-exploration, mining or agriculture automation, modeling the underlying resource is a fundamental problem. For such applications, an efficient, flexible and high-fidelity representation of the geology is critical. The key challenges in realizing this are that of dealing with the problems of uncertainty and incompleteness. Uncertainty and incompleteness are virtually ubiquitous in any sensor based application as sensor capabilities are limited. The problem is magnified in a field automation scenario due to sheer scale of the application. Incompleteness is a major problem in any large scale resource modeling endeavor as sensors have limited range and applicability. A more significant contributor to this issue is that of cost - sampling and collecting such data is expensive. Geological data is typically collected through various sensors/processes of widely differing characteristics and consequently lead to different kinds of information. Often the resource is characterized by numerous quantities (for example, soil composition in terms of numerous elements). These quantities often are correlated. 

Given these issues, large scale geological resource modeling needs a representation that can handle spatially correlated, incomplete and uncertain data. Not only must the correlation between homogeneous quantities be modeled but also that between heterogeneous quantities. This paper uses a Gaussian process (GP) representation of resource data similar to that described in \cite{vasudevan_jfr2009}. GPs are ideally suited to handling spatially correlated data. This paper further uses an extension of the basic Gaussian process model, the multi-task Gaussian process (MTGP), to simultaneously model multiple quantities of interest. The proposed model not only captures spatial correlations between individual quantities with themselves (at different locations) but also that between totally different quantities that together quantify the resource. That the quantities modeled in this paper exhibit strong correlation is known from geological sciences. This paper presents an empirical evaluation to understand (1) if simultaneous modeling of multiple quantities of interest (i.e. modeling and using the correlations between them and hence performing data fusion) is better than modeling these quantities independently and (2) if the nonstationary kernels are more effective than stationary kernels at modeling geological data. Experiments are performed on large scale real sensor data.

\section{Related work}
\label{sec:related_work}

Gaussian processes (GPs) \cite{rasmussen2006} are powerful non-parametric Bayesian learning techniques that can handle correlated, uncertain and incomplete data. They have been used in a range of fields, the Gaussian process web-site\footnote{\url{http://www.gaussianprocess.org/}} lists several examples. GPs produce a scalable multi-resolution model of the entity under consideration. They yield a continuous domain representation of the data and hence can be sampled at any desired resolution. GPs incorporate and handle uncertainty in a statistically sound manner and represent spatially correlated data appropriately. They model and use the spatial correlation of the given data to estimate the values for other unknown points of interest. GPs basically perform a standard interpolation technique known as \textit{Kriging} \cite{matheron1963}.

The work \cite{vasudevan_jfr2009}, modeled large scale terrain modeling using GPs. It proposed the use of non-stationary kernels (neural network) to model large scale discontinuous spatial data. A performance comparison between GPs based on stationary (squared exponential) and non-stationary (neural network) kernels as well as several other standard interpolation methods applicable to alternative representations of terrain data, was reported. The non-stationary neural network kernel was found to be superior to the stationary squared exponential kernel and at least as good as most standard interpolation techniques for a range of terrain (in terms of sparsity/complexity/discontinuities). The work presented in this paper builds on this GP representation. However, it addresses the problem of simultaneous modeling multiple heterogeneous quantities of interest, in the context of geological resource modeling. This requires the modeling and usage of the correlations between these quantities towards improving predictions of each of them - an instance of data fusion using Gaussian processes.

Data fusion in the context of Gaussian processes is necessitated by the presence of multiple, multi-sensor, multi-attribute, incomplete and/or uncertain data sets of the entity being modeled. Two preliminary attempts towards addressing this problem include \cite{ebeltagy2001} and \cite{msmith2005}. The former bears a ``hierarchical learning'' flavor to it in that it demonstrates how a GP can be used to model an expensive process by (a) modeling a GP on an approximate or cheap process and (b) using the many input-output data from the approximate process and the few samples available of the expensive process together in order to learn a GP for the latter. The work \cite{msmith2005} attempts to generalize arbitrary transformations on GP priors through linear transformations. It hints at how this framework could be used to introduce heteroscedasticity (random variables with non-constant variance) and how information from different sources could be fused. However, specifics on how the fusion can actually be performed are beyond the scope of the work. 

Girolami in \cite{girolami2006} integrated heterogeneous feature types within a Gaussian process classification setting, in a protein fold recognition application domain. Each feature representation is represented by a separate GP. The fusion uses the idea that individual feature representations are considered independent and hence a composite covariance function would be defined in terms of a linear sum of Gaussian process priors. A recent work by Reece et al. \cite{reece2011} integrated ``hard'' data obtained from sensors with ``soft'' information obtained from human sources within a Gaussian process classification framework. This problem/approach is different from the work presented here. It uses heterogeneous information domains (i.e. kinds of information) as mutually independent sources of information that are transformed into the kernel representation (a kernel for each kind of information) and combined using a product rule (a linear sum in Girolami's work). The focus thus, is on encoding or representing different kinds of information in a common mathematical framework using kernels. This paper is concerned with a ``higher level'' data fusion problem of heterogeneous-source information integration \emph{after} it has been represented using kernel methods. The experiments of this paper demonstrate the case when information from each source is itself from a homogeneous domain - e.g. the heterogeneous input data are all real numbers. The approach presented in this paper improves the estimate of several different quantities being simultaneously modeled by explicitly modeling the correlation between multiple heterogeneous information sources. If this is not the case (e.g. input data is made up of qualitative and quantitative data dimensions), each of heterogeneous information types can be represented by separate kernels and these can be combined using a sum or product as has been done in \cite{girolami2006,reece2011}. Simpler data fusion approaches, based on GPs, heteroscedastic GPs and their variants (see \cite{vasudevan2012}), may be applied. However, the application of the approach presented in this paper, based on multi-output or multi-task GPs, will require a non-trivial derivation of auto and cross covariances for kernels applied on heterogeneous information types.

Examples of related works that use multiple sources of the same kind of information within a single GP representation framework include \cite{thompson2008} and \cite{dragiev2011}. Whereas the former uses single output GPs to incorporate in-situ surface spectra information and remotely sensed spectra information into a kilometer scale map of the environment, the latter uses a GP implicit surface representation of an object that has to be grasped and manipulated. The representation incorporates visual, haptic and laser data into a single representation of the object. Data from each of these sensor modalities conditions the GP prior based on the implied surface at that point (on/outside/inside the object).

Two recent approaches demonstrating data fusion with Gaussian processes in the context of large scale terrain modeling were based on heteroscedastic GPs \cite{vasudevan_icra2010} and dependent GPs \cite{vasudevan_iros2010,vasudevan_icra2011}. These address the problem of fusing multiple, multi-sensor data sets of a single quantity of interest. This paper describes the framework for extending this concept to multiple heterogeneous quantities of interest. The work \cite{vasudevan_icra2010} treated the data-fusion problem as one of combining different noisy samples of a common entity (terrain) being modeled. In the Machine Learning community, this idea is referred to as heteroscedastic GPs \cite{goldberg1998,wabha2004,le2005,kersting2007}. The works \cite{vasudevan_iros2010} and \cite{vasudevan_icra2011} treated the data fusion problem as one of improving GP regression through modeling the spatial correlations (auto and cross covariances) between several dependent GPs representing the respective data sets. This idea has been inspired by recent machine learning contributions in multi-task or multi-output or dependent GP modeling including \cite{Bonilla2007} and \cite{Boyle2004}, the latter being based on \cite{Higdon2002}.  In Kriging terminology, this idea is akin to Co-kriging \cite{wackernagel2003}. The work \cite{vasudevan2012} performed a model complexity analysis of multiple approaches to data fusion using GPs, applied in the context of large scale terrain modeling. The work presented in this paper, focuses on the most generic of these approaches in the context of geological resource modeling. The significantly stronger evaluation, the discussion of ``big-picture'' issues relating to the application of the approach in practical problems, the fusion of heterogeneous data, the use of more kernels and the tying together of different prior works that have studied this approach \cite{vasudevan_iros2010,vasudevan_icra2011,melkumyan2011} are enhancements presented in this work.

The work \cite{melkumyan2011} provided preliminary findings to geological resource modeling using various combinations of stationary kernel including the squared exponential (SQEXP), Matern 3/2 and a sparse covariance function \cite{melkumyan2009}. For a geological resource modeling data set taken from a mine, it found the Matern 3/2 - Matern 3/2 - SQEXP kernel combination provided best performance in terms of the prediction error. This paper reports a detailed multi-metric benchmarking experiment, using cross validation methods, performed on a multi-task GP, an equivalent set of GPs and a set of independently optimized GPs, to provide for an exact and an independent comparison between them. The objective is to quantify the benefit (if any) of simultaneous modeling of the multiple quantities by modeling and using the correlations between them as against modeling each of these quantities separately. This paper also compares data fusion using multiple stationary and nonstationary kernels in the context of modeling geological data.

An extensive review of kernel methods applied in modeling vector valued functions was presented in a recent survey paper \cite{alvarez2012}. The paper discusses different approaches to develop kernels for multi-task applications and draws parallels between regularization perspective of this problem and a Bayesian one. The latter perspective is discussed through Gaussian processes. The work presented in this paper focuses on one of the approaches reviewed in \cite{alvarez2012}; specifically, it addresses modeling and information fusion of multi-task geological data using Gaussian processes developed using the process convolution approach. The paper presents a detailed empirical study of the approach applied to a large scale real world problem in order to evaluate its efficacy for information fusion, to understand the modeling capabilities of different kernels (chosen apriori) with such data and to understand broader approach-related questions from an application perspective. The paper also ties together past works of the authors within the process convolution theme.

\section{Approach}
\label{sec:approach}

\subsection{Gaussian processes}
\label{sec:gpintro}

Gaussian processes \cite{rasmussen2006} (GPs) are stochastic processes wherein any finite subset of random variables is jointly Gaussian distributed. They may be thought of as a Gaussian probability distribution in function space. They are characterized by a mean function $m(\mathbf{x})$ and the covariance function $k(\mathbf{x},\mathbf{x}')$ that together specify a distribution over functions. In the context of geological resource modeling, each $\mathbf{x} \;\equiv\; (east, north, depth)$ (3D coordinates) and $f(\mathbf{x}) \;\equiv\; z$, the concentration of the quantity being modeled. Although not necessary, the mean function $m(\mathbf{x})$ may be assumed to be zero by scaling/shifting the data appropriately such that it has an empirical mean of zero. 

The covariance function or kernel models the relationship between the random variables corresponding to the given data. It can take numerous forms \cite[chap. 4]{rasmussen2006}. The stationary squared exponential (or Gaussian) kernel (SQEXP) is given by  
\begin{equation}
  k_{SQEXP}(\mathbf{x},\mathbf{x}',\Sigma) = \sigma_f^2 \,.\, \exp \left( -\frac{1}{2}(\mathbf{x}-\mathbf{x}')^T \Sigma (\mathbf{x}-\mathbf{x}') \right) \,,
  \label{eqn:sqexp}
\end{equation}
\noindent where $k$ is the covariance function or kernel; $\Sigma = diag[\; l_{east} \;,\; l_{north} \;,\; l_{depth} \;]^{-2}$ is a $d$ x $d$ diagonal length-scale matrix ($d$ = dimensionality of input = 3 in this case), a measure of how quickly the modeled function changes in the east, north and depth directions; $\sigma_f^2$ is the signal variance. The set of parameters $\{\;l_{east} \;,\; l_{north} \;,\; l_{depth} \;,\; \sigma_f \;\}$ are referred to as the kernel hyperparameters. 

The non-stationary neural network (NN) kernel \cite{Neal1996,Williams1998a,Williams1998b} takes the form 
\begin{equation}
  k_{NN}(\mathbf{x},\mathbf{x}',\Sigma) \,=\, \sigma_f^2 \,.\, \frac{2}{\pi}\;\arcsin \left( \frac{2\tilde{\mathbf{x}}^T \Sigma \tilde{\mathbf{x}}'}{\sqrt{(1 \,+\,2 \tilde{\mathbf{x}}^T \Sigma \tilde{\mathbf{x}})(1 \,+\, 2\tilde{\mathbf{x}}'^T \Sigma \tilde{\mathbf{x}}')}} \right) \,,
  \label{eqn:nn}
\end{equation}
\noindent where $\tilde{\mathbf{x}}$ and $\tilde{\mathbf{x}}'$ are augmented input vectors (each point is augmented with a 1), $\Sigma$ is a $(d+1)$ x $(d+1)$ diagonal length-scale matrix given by $\Sigma = diag[\; \beta \;,\, l_{east} \;,\; l_{north} \;,\; l_{depth} \;]^{-2}$, $\beta$ being a bias factor and $d$ being the dimensionality of the input data. The variables $\{\;\beta \;,\; l_{east} \;,\; l_{north} \;,\; l_{depth} \;,\; \sigma_f \;\}$ constitute the kernel hyperparameters. The NN kernel represents the covariance function of a neural network with a single hidden layer between the input and output, infinitely many hidden nodes and using a Sigmoidal transfer function \cite{Williams1998a} for the hidden nodes. Hornik, in \cite{Hornik1993}, showed that such neural networks are universal approximators and Neal, in \cite{Neal1996}, observed that the functions produced by such a network would tend to a Gaussian process. Prior work in \cite{vasudevan_jfr2009} found the NN kernel to be more effective than the SQEXP kernel at modeling discontinuous data. 

The Matern 3/2 kernel is another stationary kernel differing from the SQEXP kernel in that the latter is infinitely differentiable and consequently tends to have a strong smoothing nature, which is argued as being detrimental to modeling physical processes \cite{rasmussen2006}. It takes the form specified in Equation \ref{eqn:matern3}.
\begin{equation}
  k_{MATERN3}(x,x',\Sigma) \,=\, \sigma_f^2 \,.\, {\prod_{1 \leq k \leq d} (1 + \frac{\sqrt{3}r_k}{l_{k}}) \exp{\left(-\frac{\sqrt{3}r_k}{l_{k}}\right)}}
  \label{eqn:matern3}
\end{equation}
where $k \;\epsilon\; 1 \ldots d$ is the dimension of the input data ($d$ = dimensionality of input = 3 in this case), $\Sigma = [\; l_{east} \;,\; l_{north} \;,\; l_{depth} \;]$ is a $1$ x $d$ length-scale matrix, a measure of how quickly the modeled function changes in the east, north and depth directions; $\sigma_f^2$ is the signal variance. The set of parameters $\{\;l_{east} \;,\; l_{north} \;,\; l_{depth} \;,\; \sigma_f \;\}$ is referred to as the kernel hyperparameters. 

Regression using GPs uses the fact that any finite set of training (evaluation) data and test data of a GP are jointly Gaussian distributed. Assuming noise free data, this idea is shown in Expression \ref{eqn:gpmodel} (hereafter referred to as Equation \ref{eqn:gpmodel}). This leads to the standard GP regression equations yielding an estimate (the mean value, given by Equation \ref{eqn:gpmean}) and its uncertainty (Equation \ref{eqn:gpcov}). 
\begin{equation}
  \left[ \begin{array}{c}
      \mathbf{z} \\
      f_*
    \end{array}
  \right]
  \;\sim\;
  N \left( 0 \,,\,
    \left[
      \begin{array}{c c}
        K(X,X)   & K(X,X_*) \\
        K(X_*,X) & K(X_*,X_*)
      \end{array}
    \right]
  \right)    
  \label{eqn:gpmodel}
\end{equation}

\begin{align}
  \bar{f}_* = {} & K(X_*,X) \; K(X,X)^{-1} \; \mathbf{z}  \label{eqn:gpmean} \\
  \mathrm{cov}(f_*) = {} & K(X_*,X_*) - K(X_*,X) K(X,X)^{-1} K(X,X_*) \label{eqn:gpcov}
\end{align}

For $n$ training points $(X,\mathbf{z}) \,=\, (\mathbf{x_i},z_i)_{i=1 \ldots n}$ and $n_*$ test points $(X_*,f_*)$, $K(X,X_*)$ denotes the $n \times n_*$ matrix of covariances evaluated at all pairs of training and test points. The terms $K(X,X)$, $K(X_*,X_*)$ and $K(X_*,X)$ are defined likewise. In the event that the data being modeled is noisy, a noise hyperparameter ($\sigma$) is also learnt with the other GP hyperparameters and the covariance matrix of the training data $K(X,X)$ is replaced by $[ K(X,X)+\sigma^2I ]$ in Equations \ref{eqn:gpmodel}, \ref{eqn:gpmean} and \ref{eqn:gpcov}. GP hyperparameters may be learnt using various techniques such as cross validation based approaches \cite{rasmussen2006} and maximum-a-posteriori approaches using Markov Chain Monte Carlo techniques \cite{rasmussen2006,Williams1998b} and maximizing the marginal likelihood of the observed training data \cite{rasmussen2006,vasudevan_jfr2009}. This paper adopts the latter most approach based on the intuition that it may be more suited for large data sets. The marginal likelihood to be maximized is described in Equation \ref{eqn:gplml}.
\begin{equation}
  \log\; p(\mathbf{z}|X,\theta) \,=\, -\frac{1}{2} \mathbf{z}^T K(X,X)^{-1}\mathbf{z} - \frac{1}{2} \log|K(X,X)| -\frac{n}{2} \log(2\pi) \label{eqn:gplml}
\end{equation}

\subsection{Multi-task Gaussian processes (MTGPs)}
\label{sec:dgp}

The problem being addressed in this paper can be described as follows. The objective is to model multiple heterogeneous quantities (e.g. concentrations of various elements) of the entity in consideration (e.g. land mass). The data fusion aspect of this problem is the improved estimation of each one of these quantities by integration or use of all other quantities of interest. If each quantity is modeled using a separate GP, the objective is to improve one GPs prediction estimates given all other GP models.

\begin{figure*}[htbp]
  \begin{center}
    \subfigure{\includegraphics[width=0.85\columnwidth]{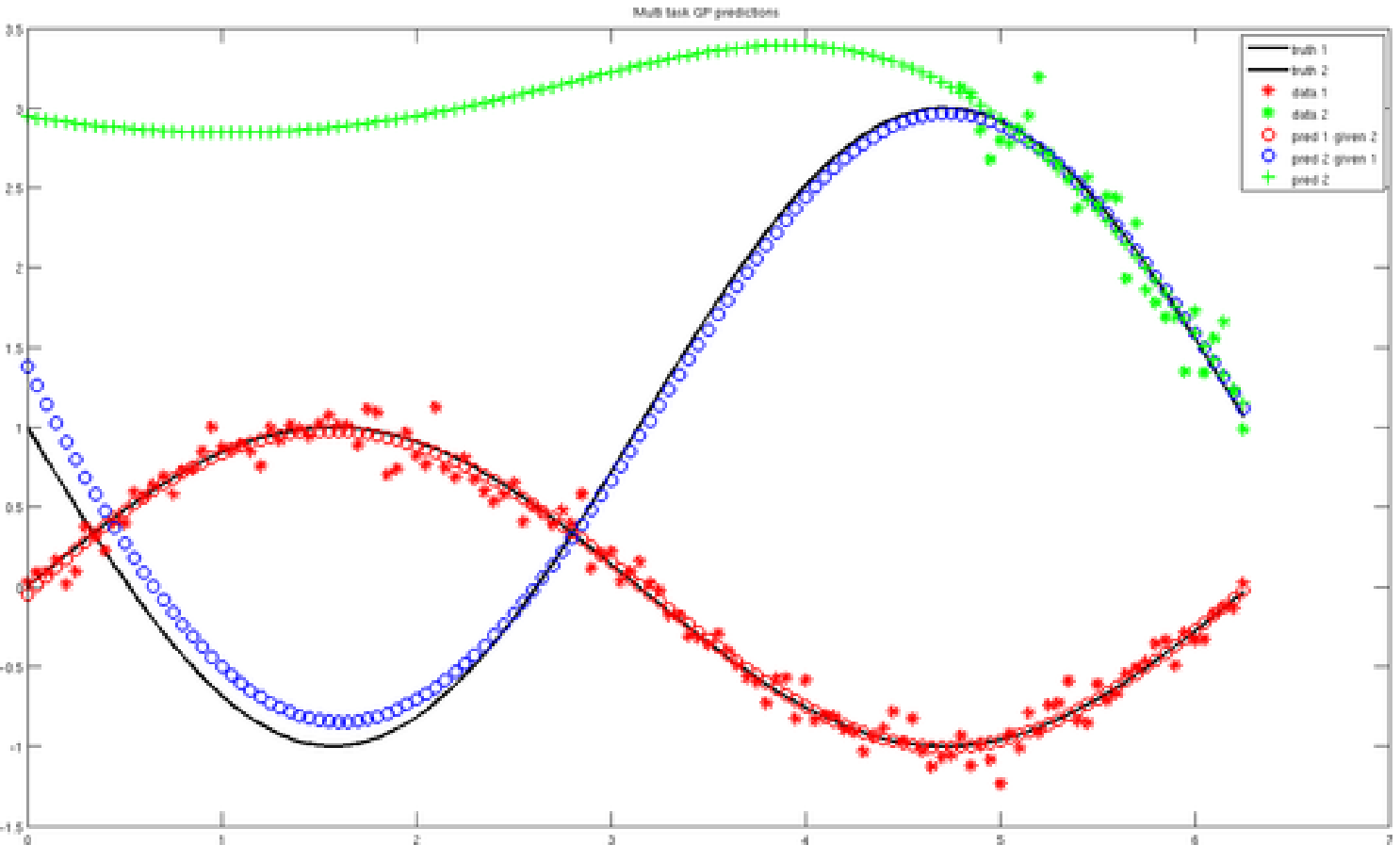}}
    \subfigure{\includegraphics[width=0.85\columnwidth]{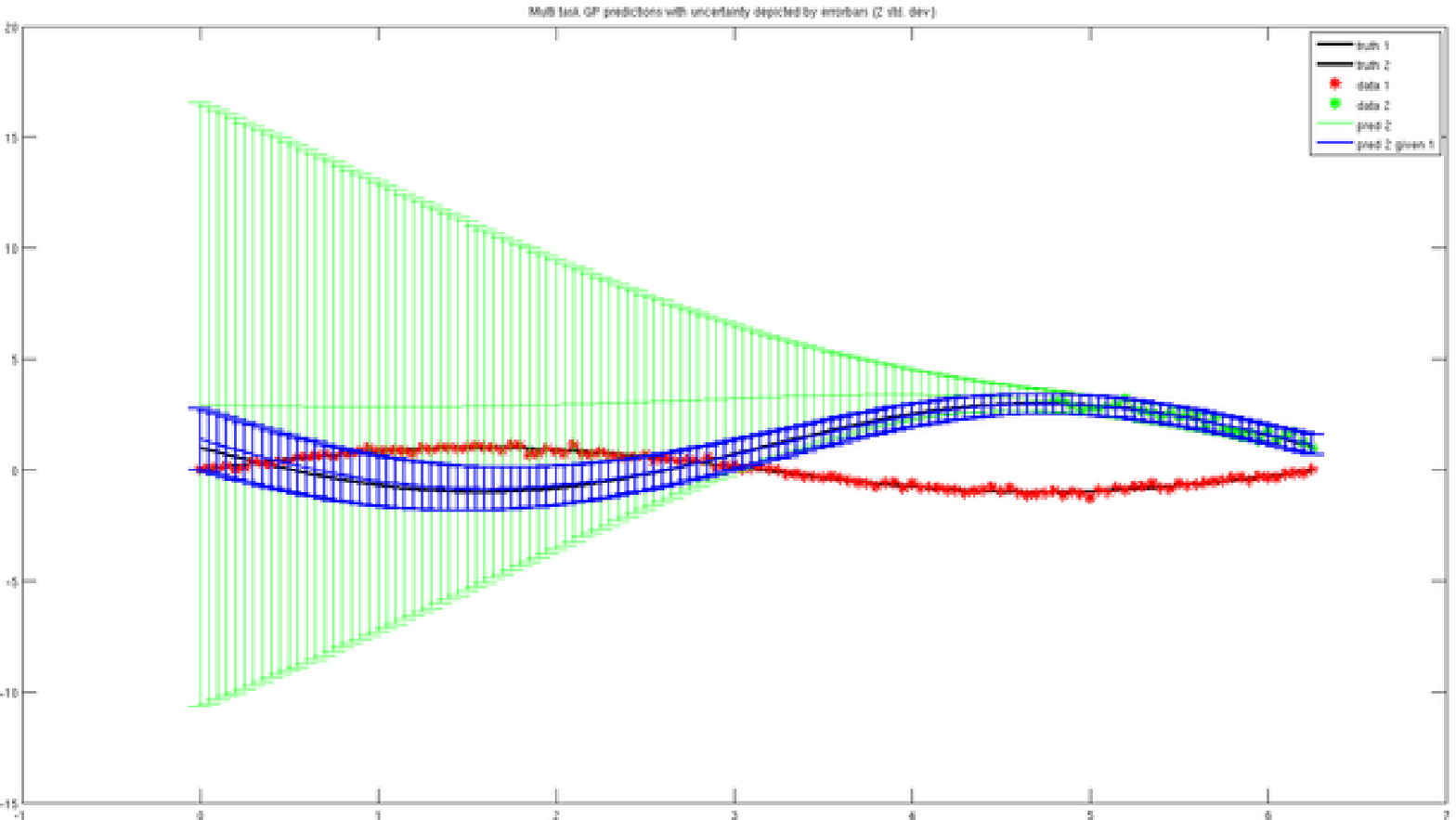}}
  \end{center}
  \caption{A simple demonstration of the MTGP/DGP concept demonstrating data fusion. Two sine waves (black) are to be modeled. One is an inverted function of the other. Noisy samples are available all over one of them (red) whereas the other one has noisy samples only in one part of it (green). Merely using these few green samples would result in a poor prediction of the sine wave in the areas devoid of samples. Using the spatial correlation with the red sampled sine wave enables the MTGP approach to improve the prediction of the green sampled sine wave. The figure above shows the predictions of the GPs given the other GP (red/blue circles) and that of the second GP taken alone (green plus marks). The figure below shows the uncertainty in predictions (error bars of two standard deviations about mean) of the second GP taken alone (green) and that when taken together with the first GP (blue) - a clear reduction in uncertainty is observed.}
  \label{fig:mogp}
\end{figure*}
Multi-task Gaussian processes (MTGPs or multi-output GPs or Dependent GPs) extend Gaussian processes to handle multiple correlated outputs simultaneously. The main advantage of this technique is that the model exploits not only the spatial correlation of data corresponding to one output but also those of the other outputs. This improves GP regression/prediction of an output given the  others, thus performing data fusion. Figure \ref{fig:mogp} shows a simulated example of this concept.

Let the number of outputs/tasks that need to be simultaneously modeled be denoted by $nt$. Equations \ref{eqn:gpmodel}, \ref{eqn:gpmean} and \ref{eqn:gpcov} represent respectively the MTGP data fusion model, the regression estimates and their uncertainties, subject to the following modifications to the basic notation. The set 
$$\mathbf{z} = [\;\mathbf{z}_1\;,\;\mathbf{z}_2\;,\;\mathbf{z}_3\;,\;...\;,\;\mathbf{z}_{nt}\;]'$$ represents the output values of the selected training data from the individual $nt$ tasks that need to be simultaneously modeled. The term $$X = [\;X_1\;,\;X_2\;,\;X_3\;,\;...\;,\;X_{nt}\;]$$ denotes the input location values (east, north, depth) of the selected training data from the individual data sets. Any kernel \cite{rasmussen2006} may be used and even different kernel could be used for different data sets using the technique demonstrated in \cite{melkumyan2011} (for stationary kernel) or the convolution process technique demonstrated in \cite{Higdon2002,Boyle2004,vasudevan_iros2010,vasudevan_icra2011} and in this paper (for both stationary and nonstationary kernel). The covariance matrix of the training data is given by 
$$  K(X,X) \,\equiv\,   \left[ \begin{array}{c c c c}
    K^Y_{11}    & K^Y_{12}  & \ldots & K^Y_{1\,nt} \\
    K^Y_{21}    & \ldots   & \ldots & \vdots    \\
    \vdots     & \vdots   & \vdots & \vdots    \\
    K^Y_{nt\,1}  & \ldots   & \ldots & K^Y_{nt\,nt} \\
  \end{array} \right] $$ 
where
\begin{eqnarray}
  K_{ii}^Y & = & K_{ii}^U(X_i,X_i) \,+\, \sigma_i^2I \nonumber\\
  K_{ij}^Y & = & K_{ij}^U(X_i,X_j) \,. \nonumber
\end{eqnarray}
Here, $K_{ii}^Y$ represents the auto-covariance of the $i^{th}$ data set with itself and $K_{ij}^Y$ represents the cross covariance between the $i^{th}$ and $j^{th}$ data sets. These terms model the covariance between the noisy observed data points ($z$ values). Thus, they also take the noise components of the individual data sets / GPs into consideration. The corresponding noise free terms are respectively given by $K_{ii}^U$ and $K_{ij}^U$. These are derived by using the process convolution approach to formulating Gaussian processes; details of this follow in the next subsection. The covariance matrix between the test points and training points is given by 

$$K(X_*,X) = [ K^U_{i1}(X_*,X_1), K^U_{i2}(X_*,X_2), \ldots, K^U_{i\,nt}(X_*,X_{nt}) ] \;,$$

where $i \,\epsilon\, \{1\,\ldots\,nt\}$ is the GP that is being evaluated given all other GPs. The matrix $K(X,X_*)$ is defined likewise. Finally, the covariance of the test points is given by $$K(X_*,X_*) \,=\, K^U_{ii}(X_*,X_*)+\sigma_i^2I \;,$$ assuming the $i^{th}$ GP needs to be evaluated for the particular test point. The mean and variance of the concentration estimate can thus be obtained by applying Equations \ref{eqn:gpmean} and \ref{eqn:gpcov}, after incorporating multiple outputs/tasks, multiple GP/noise hyperparameters and deriving appropriate auto and cross covariances functions that model the spatial correlation between the individual data sets. Data fusion is thus achieved in the MTGP approach by correlating individual heterogeneous outputs/tasks and using this correlation information to improve the prediction estimates of each of them.

\subsection{Derivation of the auto and cross covariance terms}
\label{sec:pca}

The main challenge in the use of multi-task GPs is the derivation of closed form cross (and auto) covariance functions. The process convolution approach to modeling GPs, proposed in \cite{Higdon2002}, can address this problem. The cited paper (1) modeled a GP as the convolution of a ``smoothing kernel'' and a Gaussian white noise process, (2) expressed a relationship between the ``smoothing kernel'' and the corresponding covariance function through the Fourier transform, (3) noted that for stationary isotropic kernels, there existed a one-to-one relationship between the covariance function and its smoothing kernel and that for non-isotropic and/or non-stationary kernels, there was no unique solution to the smoothing kernel and (4) hinted at how this approach may be used to develop GP models with complex properties (e.g. nonstationarity). As a consequence of this approach, modeling the GP amounted to modeling the hyperparameters of the smoothing kernel. For the second point above, the paper suggested that the smoothing kernel for a covariance function could be obtained as the Inverse Fourier Transform of the square root of the spectrum (Fourier transform) of the covariance function. The process convolution approach to MTGPs has been used with the stationary SQEXP kernel in \cite{Boyle2004,alvarez2008,vasudevan_iros2010} and the nonstationary NN kernel in \cite{vasudevan_icra2011,vasudevan2012}. Once the smoothing kernel is identified for a covariance function, the cross-covariance between two covariance functions can be derived as a kernel correlation between the respective smoothing kernels \cite{Boyle2004}. The following mathematical formalism is based on \cite{Higdon2002} and \cite{Boyle2004}.

\begin{align}
  Y_i(s) = {} & U_i(s) \,+\, W_i(s) \label{eqn:pc1} \\
  U_i(s) = {} & \int_s k_i(s,\lambda) \,\star\, X(\lambda) \,\, d\lambda  \label{eqn:pc2}
\end{align}

\begin{align}
  K_{ij}^{U}(s_{a},s_{b}) = {} & E\left\{ U_{i}(s_{a})\, U_{j}(s_{b})\right\} \nonumber \\
  = {} & E\left\{ \int k_{i}(s_{a},\alpha).X(\alpha)d\alpha\;\int k_{j}(s_{b},\beta).X(\beta)d\beta\right\}\nonumber\\
  = {} & \int k_{i}(s_{a},\alpha)\, k_{j}(s_{b},\alpha)\: d\alpha \label{eqn:ccov} \\
  K_{ii}^{U}(s_{a},s_{b}) = {} & \int k_{i}(s_{a},\alpha)\, k_{i}(s_{b},\alpha)\: d\alpha \label{eqn:acov}
\end{align}

Mathematically, if $Y_i(s)$ represents the observed data in Equation \ref{eqn:pc1}, it is expressed as a combination of a noise-free GP $U_i(s)$ and Gaussian white noise process $W_i(s)$. The GP $U_i(s)$ is further modeled as a convolution of a smoothing kernel $k_i(s,\lambda)$ and a Gaussian white noise process $X(\lambda)$, as shown in Equation \ref{eqn:pc2}. A stationary and/or isotropic smoothing kernel would take the form $k_i(s-\lambda)$ as it would be a function of the distance between the input points. If two covariance functions (corresponding to two GPs $U_i(s)$ and $U_j(s)$) have smoothing kernels $k_i(s_a,\lambda)$ and $k_j(s_b,\lambda)$ respectively, then the cross covariance between them can be derived as shown in Equation \ref{eqn:ccov}. The auto covariance can be deduced from the cross covariance expression and take the form shown in Equation \ref{eqn:acov}. The smoothing kernel $k_i$ and $k_j$ need to be finite energy kernels i.e. $\int|\;k_{i}(x_{a},\alpha)\;|^{2}\: d\alpha\;<\infty$. This can be intrinsically true of some kernel (e.g. squared exponential kernel) or can be true subject to the bounded application of the kernel (e.g. neural network kernel).

The work \cite{melkumyan2011} suggested that if a covariance function could be written as a convolution of its ``basis functions'' (the form specified in Equation \ref{eqn:acov}), then a cross-covariance between two covariance functions could be derived as a kernel correlation of their respective basis functions (the form specified in Equation \ref{eqn:ccov}). The paper proved that the resulting cross-covariance would be positive definite. In order to find the basis function for a particular covariance function, the paper derived an expression in terms of its Fourier transform. This relationship is identical to that suggested by \cite{Higdon2002} and valid for stationary kernels only. The paper also derives closed form cross-covariance functions for different combinations of stationary kernels including the squared exponential, Matern 3/2 and a sparse covariance function developed by the authors in \cite{melkumyan2009}. 

This paper argues that both of these methods using the ``smoothing kernel'' \cite{Higdon2002} and the ``basis functions'' \cite{melkumyan2011} are actually equivalent with the former providing a sound basis to explain the latter as well as a powerful framework to develop other complex GP models such as space-time models and nonstationary GPs. The key insight obtained here is in the methodology of identifying the smoothing kernel for the process convolution approach. If the covariance function is a stationary kernel, there is an exact one-to-one relationship between the covariance function and the smoothing kernel as pointed out in \cite{Higdon2002} and whose expression is derived in \cite{melkumyan2011}. If the covariance function is nonstationary, several possible smoothing kernels may lead to the same covariance function, as pointed out in \cite{Higdon2002}. However, attempting to express the kernel in a separable form (e.g. as the correlation of two identically formed basis functions) and thereby identifying the smoothing kernel would be \emph{one} possible approach, if the form of the kernel form allowed for such separation. Needless to say, this idea would be applicable only in a restricted class of covariance functions and finding a universal approach to identifying the smoothing kernel for other nonstationary kernel remains an open question. Given the smoothing kernel of the covariance functions in consideration, the cross-covariance terms can be derived as a kernel correlation as demonstrated in \cite{Higdon2002,Boyle2004,vasudevan_icra2011,vasudevan2012,melkumyan2011}.

Assume two GPs $N(0, k_i)$ and $N(0, k_j)$, with with length scale matrices $\Sigma_i$ and $\Sigma_j$. Based on \cite{Boyle2004}, the cross and auto covariances for the stationary SQEXP kernel are given by Equations \ref{eqn:ccov_sqexp} and \ref{eqn:acov_sqexp} respectively. The corresponding expressions for the nonstationary NN kernel are derived in \cite{vasudevan_icra2011,vasudevan2012} and given in Equations \ref{eqn:ccov_nn} and \ref{eqn:acov_nn} respectively. For the Matern 3/2 kernel, the expressions for the cross covariance and auto covariance are derived in \cite{melkumyan2011} and given in Equations \ref{eqn:ccov_matern3} and \ref{eqn:acov_matern3} respectively. Also based on \cite{melkumyan2011}, the cross covariance function between an SQEXP and a Matern 3/2 kernel is given by Equation \ref{eqn:ccov_sqexp_matern3}.

\noindent \fbox{\begin{minipage}{\textwidth-0.4\parindent}
    \begin{equation}
      K_{ij}^U(x,x') = K_f(i,j) \; \frac{(2\pi)^{\frac{d}{2}}}{|\Sigma_i \,+\, \Sigma_j|^{\frac{1}{2}}} \; \exp{\left( -\frac{1}{2} (x-x')^T \Sigma_{ij} (x-x') \right)} \label{eqn:ccov_sqexp}
    \end{equation}
    where  $$\Sigma_{ij} \,=\, \Sigma_i(\Sigma_i+\Sigma_j)^{-1}\Sigma_j \,=\, \Sigma_j(\Sigma_i+\Sigma_j)^{-1}\Sigma_i$$    
    \begin{equation}
      K_{ii}^U(x,x') = K_f(i,i) \; \frac{(\pi)^{\frac{d}{2}}}{|\Sigma_i|^{\frac{1}{2}}} \; \exp{\left(-\frac{1}{4}(x-x')^T\Sigma_i(x-x')\right)} \label{eqn:acov_sqexp}
    \end{equation}    
  \end{minipage}}

\noindent \fbox{\begin{minipage}{\textwidth-0.4\parindent}
    \begin{equation}
      K_{ij}^U(x,x') = K_f(i,j) \; 2^{\frac{d+1}{2}} \frac{|\Sigma_i|^{\frac{1}{4}} |\Sigma_j|^{\frac{1}{4}}}{|\Sigma_i \;+\; \Sigma_j|^{\frac{1}{2}}} \; k_{NN}(\mathbf{x},\mathbf{x'},\Sigma_{ij}) \label{eqn:ccov_nn}
    \end{equation}
    where $$\Sigma_{ij} \,=\, 2\;\Sigma_i\;(\Sigma_i\;+\;\Sigma_j)^{-1}\;\Sigma_j$$    
    \begin{equation}
      K_{ii}^U(x,x') \,=\, K_f(i,i) \; k_{NN}(\mathbf{x},\mathbf{x'},\Sigma_i) \label{eqn:acov_nn}
    \end{equation}
\end{minipage}}

\noindent \fbox{\begin{minipage}{\textwidth-0.4\parindent}
    \begin{equation}
      K_{ij}^U(x,x') \,=\, K_f(i,j) \; {\prod_{1 \leq k \leq d} \frac{2 l_{ik}^\frac{1}{2} l_{jk}^\frac{1}{2}}{l_{ik}^2-l_{jk}^2}%
        \left(l_{ik}e^{-\sqrt{3}\frac{r_k}{l_{ik}}}-l_{jk}e^{-\sqrt{3}\frac{r_k}{l_{jk}}}\right)} \label{eqn:ccov_matern3}
    \end{equation}
    where $k \;\epsilon\; 1 \ldots d$ is the dimension of the input data, $l_i$ and $l_j$ are the length scales for the two Matern 3/2 kernel based GPs $i$ and $j$, $l_{ik}$ and $l_{jk}$ are the $k^{th}$ length scales (corresponding to the $k^{th}$ dimensions) of these GPs and $r_k\,=\,|x_k-x'_k|$ is the distance in the $k^{th}$ dimension between the input data.    
    \begin{equation}
      K_{ii}^U(x,x') \,=\, K_f(i,i) \; k_{MATERN3}(\mathbf{x},\mathbf{x'},\Sigma_i) \label{eqn:acov_matern3}
    \end{equation}
\end{minipage}}

\vspace{5mm}
\noindent \fbox{\begin{minipage}{\textwidth-0.4\parindent}
    \begin{align}
      K_{ij}^U(x,x') \;=\; K_f(i,j) & {\prod_{1 \leq k \leq d} \sqrt{\lambda_k} \left(\frac{\pi}{2}\right)^{1/4} e^{\lambda_k^{2}} \Bigg[ 2\cosh \left(\frac{\sqrt{3}r_k}{l_{Mk}}\right) } - \nonumber \\
      & e^{\frac{\sqrt{3}r_k}{l_{Mk}}} \operatorname{erf} \left(\lambda_k +\frac{r_k}{l_{SEk}}\right) - e^{-\frac{\sqrt{3}r_k}{l_{Mk}}} \operatorname{erf} \left(\lambda_k -\frac{r_k}{l_{SEk}}\right) \Bigg] \label{eqn:ccov_sqexp_matern3}
    \end{align}
    where $\lambda_k =\frac{\sqrt{3}}{2}\frac{l_{SEk}}{l_{Mk}}$, $\operatorname{erf}\left(x\right)=\frac{2}{\sqrt{\pi}}\int_{0}^{x}e^{-t^{2}}\textrm{d}t$, $k \;\epsilon\; 1 \ldots d$ is the dimension of the input data, $l_{SE}$ and $l_M$ are the respective length scales for the SQEXP and Matern 3/2 kernel based GPs $i$ and $j$, $l_{SEk}$ and $l_{Mk}$ are the $k^{th}$ length scales (corresponding to the $k^{th}$ dimensions) of these GPs and $r_k\,=\,|x_k-x'_k|$ is the distance in the $k^{th}$ dimension between the input data.
  \end{minipage}}    
\vspace{5mm}

In Equations \ref{eqn:ccov_nn} and \ref{eqn:acov_nn}, the term, $k_{NN}(\mathbf{x},\mathbf{x'},\Sigma_{ij})$, is the NN kernel for two data $\mathbf{x}$, $\mathbf{x'}$ and length scale matrix $\Sigma_{ij}$. It is given by Equation \ref{eqn:nn}, excluding the signal variance term ($\sigma_f^2$). Likewise, in Equation \ref{eqn:acov_matern3}, $k_{MATERN}(\mathbf{x},\mathbf{x'},\Sigma_i)$ refers to the Matern 3/2 kernel for two data $\mathbf{x}$, $\mathbf{x'}$ and length scale matrix $\Sigma_{ij}$, given by Equation \ref{eqn:matern3} (excluding the $\sigma_f^2$ term). The $K_f$ terms in Equations \ref{eqn:ccov_sqexp}, \ref{eqn:acov_sqexp}, \ref{eqn:ccov_nn} and \ref{eqn:acov_nn} are inspired by \cite{Bonilla2007}. This term models the task similarity between individual tasks. Incorporating it in the auto and cross covariances provides additional flexibility to the multi-task GP modeling process. It is a symmetric matrix of size $nt$ x $nt$ and is learnt along with the other GP hyperparameters. Thus, the hyperparameters of the system that need to be learnt include $(nt.(nt+1))/2$ task similarity values, $nt \,.\, 2$ or $nt \,.\,3$ length scale values respectively for the individual SQEXP/MATERN3 or NN kernels and $nt$ noise values corresponding to the noise in the observed data sets. Learning these hyperparameters by adapting the GP learning procedure described before (Equation \ref{eqn:gplml}) for multiple outputs/tasks \cite{vasudevan_iros2010,vasudevan_icra2011}.

%%%%%%%%%%%%%%%%%%%%%%%%%%%%%%%%%%%%%%%%%%%%%%%%%%%%%%%%%%%%%%%%%%%%%%%%

\section{Experiments}
\label{sec:experiments}

\begin{figure}[htbp]
  \begin{center}
    \subfigure[Element-1 (E1) concentration]{\includegraphics[width=\columnwidth]{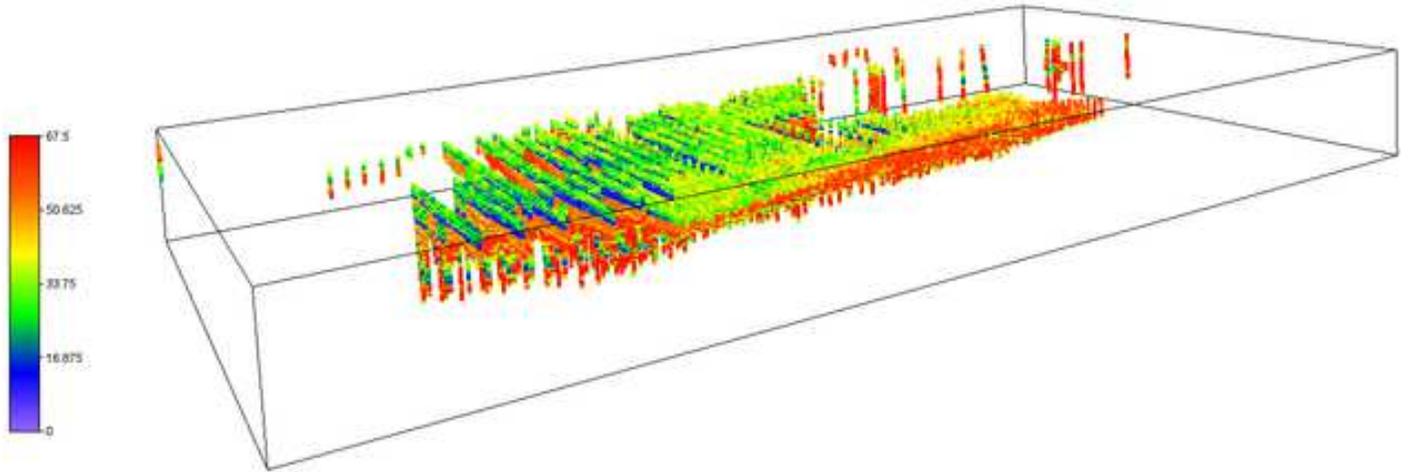}\label{fig:e1}}
    \subfigure[Element-2 (E2) concentration]{\includegraphics[width=\columnwidth]{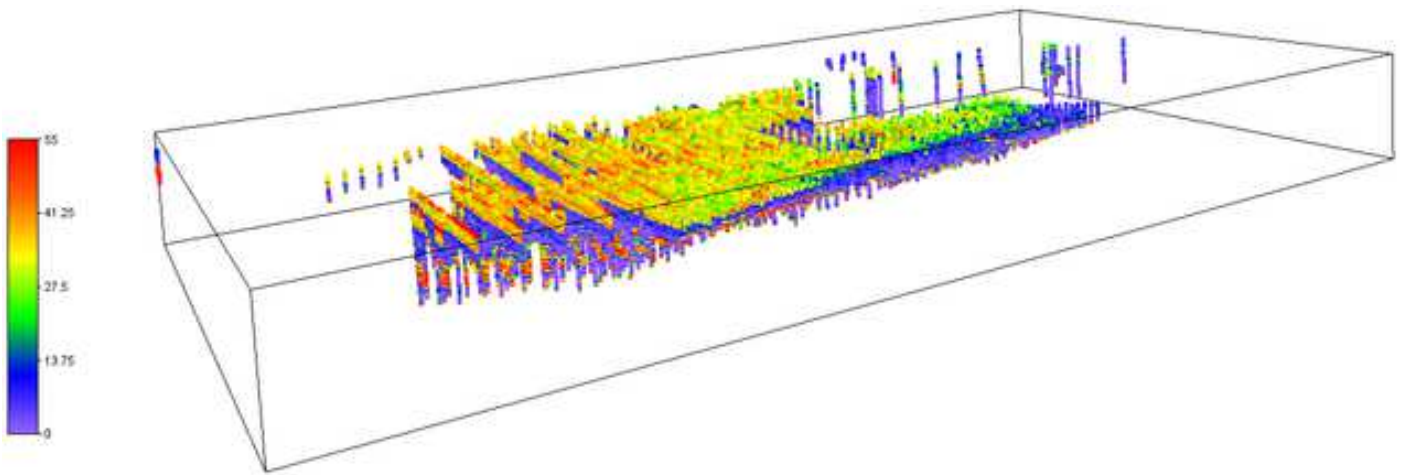}\label{fig:e2}}
    \subfigure[Element-3 (E3) concentration]{\includegraphics[width=\columnwidth]{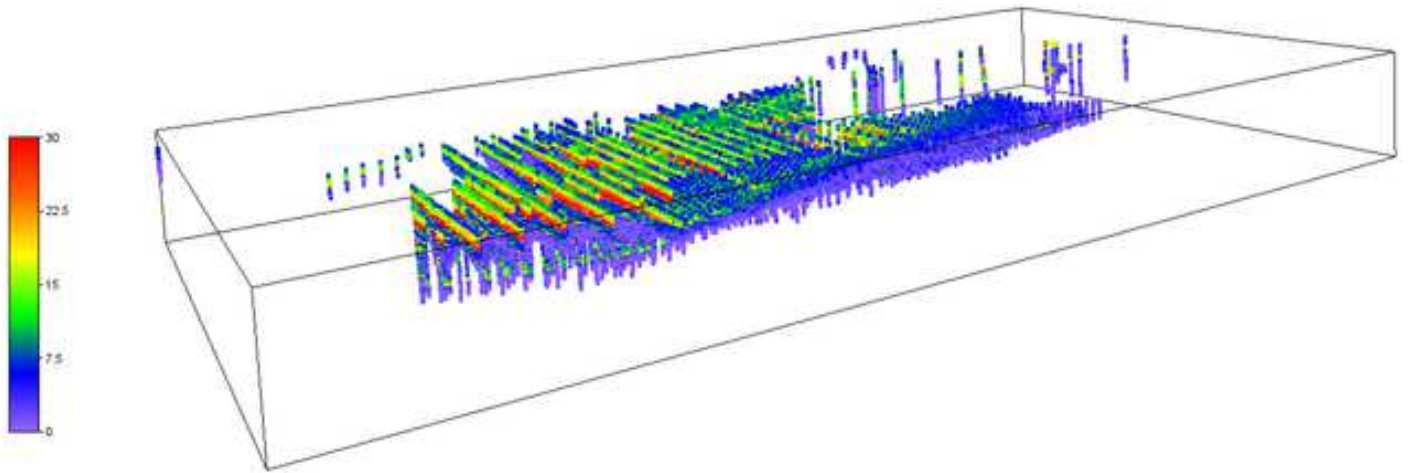}\label{fig:e3}}
  \end{center}
  \caption{The geological resource data set. Figures \ref{fig:e1}, \ref{fig:e2} and \ref{fig:e3} respectively show the concentrations of three elements over the region of interest. The central region of points is surrounded by sparse sets of points which are not pre-filtered when applying the proposed algorithm.}
  \label{fig:data}
\end{figure}

Experiments were conducted on a large scale geological resource data set made up of real sensor data. The data consists of 63,667 measurements from a 3478.4 m x 1764.6 m x 345.9 m region in Australia that has undergone drilling and chemical assays to determine its composition. The holes are generally 25-100m apart and tens to hundreds of meters deep. Within each hole, data is collected at an interval of 2m. The measurements include the (east, north, depth) position data along with the concentrations of three elements, Element-1, Element-2 and Element-3, hereafter denoted as E1, E2 and E3 respectively. These three quantities are known to be correlated and hence the objective is to use each of their GP models to improve the others' prediction estimates by capturing the correlation between these quantities. The data set is shown in Figure \ref{fig:data}. The methodology of testing is described in Section \ref{sec:exp:testprocedure}. Multiple metrics have been used to evaluate the methods, these are described in Section \ref{sec:exp:metrics}. Results obtained are then presented and discussed in Section \ref{sec:exp:results}. Outputs of the data fusion process provided by the best performing model as suggested by the evaluation are also presented.

\subsection{Testing procedure}
\label{sec:exp:testprocedure}

The objective of the experiment was to compare the multi-task GP approach with a conventional GP approach and quantify if the data fusion in the MTGP actually improves estimation. A second objective of the experiments was to compare the nonstationary NN kernel with the stationary SQEXP kernel, the Matern 3/2 kernel and a combination of them that proved effective in prior testing \cite{melkumyan2011}. Towards these aims, a ten fold cross validation experiment was performed on the data set, with each of the kernels. This was motivated by the work \cite{kohavi1995study}, which suggests a ten fold stratified (similar number of samples in each fold) cross validation as the best way of testing the estimation accuracy of machine learning methods on real world data sets.

The MTGP and simple GP approaches each require an optimization step for model learning. The optimization step in each method can result in different local minima in each trial (and with each kernel). Thus, to do a one-on-one comparison between the two approaches and quantify their relative performances, an exact comparison is required.  The benchmarking experiment presented in this paper provides an \emph{exact} comparison between the MTGP and GP approaches. To do this, 
\begin{itemize}
\item The best available MTGP parameters were found for each kernel. From this, appropriate subsets of the parameters were chosen for the GP approach.
\item The approaches were compared on identical test points and identical training/evaluation points selected for each of the test points.
\item It is also necessary that the covariance function for the simple GP approach \emph{must} be identical to the auto-covariance function of the DGP approach. For this reason, the auto-covariance function (for both kernels) is used as the covariance function for the GP approach to data fusion.
\end{itemize}
In addition to this, three independent GPs (denoted as GPI here after) were optimized for E1, E2 and E3 and their estimates for the same set of test points were also compared. Thus the effect of information integration in the context of the geological resource modeling can be seen in terms of both an exact comparison (MTGP vs GP) and an independent comparison (MTGP vs GPI).

\begin{figure}[htbp]
  \begin{center}
    \includegraphics[width=0.9\textwidth]{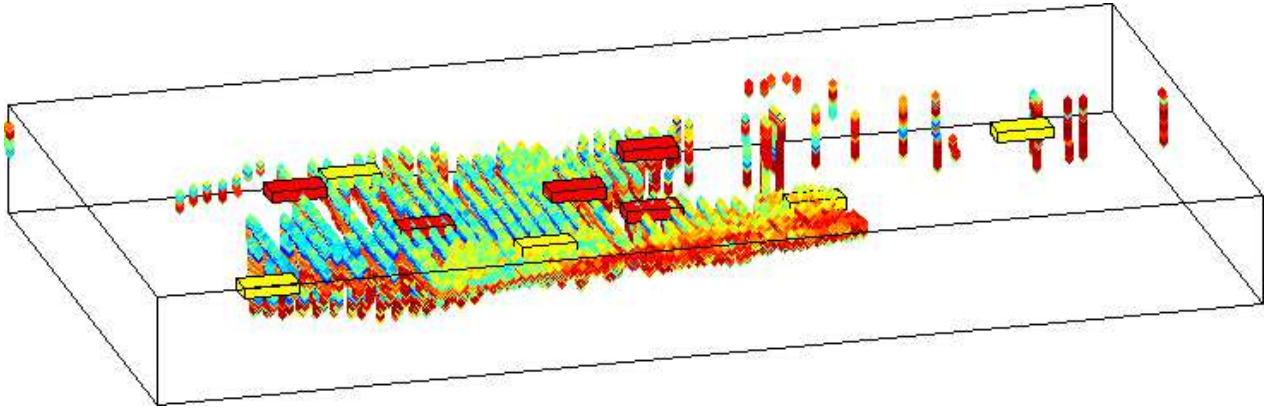}
  \end{center}
  \caption{Example of 3D block sampling of a geological resource data set. Blocks may be sampled of different sizes. The red and yellow blocks represent blocks from two of the ten folds used in cross validation testing. Test points within these blocks have ``support'' data away from them, outside the blocks. This sampling method is therefore a stronger test of the robustness of an approach to estimating the quantity of interest, as compared to uniformly sampling test points. The estimation errors however, will be higher than that obtained for a uniformly sampled set of points.}
  \label{fig:bs}
\end{figure}

For the cross validation, a ``block'' sampling technique (see Figure \ref{fig:bs}) was used, a 3D version of the ``patch'' sampling method used in \cite{vasudevan_jfr2009}. The idea was that rather than selecting test points uniformly, blocks of data test the robustness of the approach better as the support points to the query point are situated farther away than in uniform point selection. The data set is gridded into blocks of different sizes. Collections of blocks represent individual folds. In each cross validation test, one fold was designated as a test fold and points from it were used exclusively for testing. All other folds together constituted the evaluation data, a small subset of which were labeled as the training data. Note that this technique of testing will naturally lead to larger errors. For the test fold, the E1, E2 and E3 concentrations (and error metrics defined in the following section) are estimated first using the MTGP approach, then with the GP approach using parameters from the optimized MTGP parameters and finally, with an independently optimized GP for each of the three quantities. The result of a 10 fold cross validation test is a 63,667 point evaluation in tougher test conditions than what would be attainable with uniform sampling (e.g. every tenth point) of test points. 

Block sizes were chosen empirically, in proportion (arbitrarily rounded up or down) to the dimensions of the whole data set and with a view of performing a stratified cross validation test. The block sizes chosen and the resulting implications on the cross validation testing are shown in Table \ref{tab:cvp_bs}. The smaller block size of 22m x 11m x 2m results in each fold having a similar number of points (i.e. numbers of points in folds with min/max test points are similar) and thus results in the most stratified cross validation test. With increasing block size, prediction error increases (support data is farther away), stratification is reduced and hence, variance in prediction error also increases. Uniform sampling of test points may be considered as a limiting case of block sampling with the smallest block size possible.

\begin{table}[htb]
  \begin{center}
    \caption{10 fold cross validation with block sampling; 63667 points in data set spread over 3478.4 m x 1764.6 m x 345.9 m; block sizes tested vs relative implications on results}
    \begin{tabular}{| c | c | c | c | c | c | c | c |}
      \hline
      \hline
      Block size     & Number of points & Number of points & Comments on           \\
      (m)            & in fold with MIN & in fold with MAX & cross validation test \\
                     & test points      & test points      &                        \\
      \hline
      22 x 11 x 2    %
      &    6209          &      6454        & Most stratified cross validation \\
      &                  &                  & Least prediction error           \\
      \hline
      44 x 22 x 4    %
      &    6183          &      6456        & stratification $\downarrow$ prediction error $\uparrow$ \\
      \hline
      87 x 45 x 9    %
      &    5807          &      6739        & stratification $\downarrow$ prediction error $\uparrow$ \\
      \hline
      174 x 89 x 18  %
      &    5133          &      7549        & stratification $\downarrow$ prediction error $\uparrow$ \\
      \hline
      348 x 177 x 35 %
      &    4976          &      9662        & stratification $\downarrow$ prediction error $\uparrow$\\
      \hline
      696 x 353 x 70 %
      &    1204          &      10371       & Least Stratified cross validation \\
      &                  &                  & Highest prediction error          \\
      \hline
    \end{tabular}
    \vspace{-4mm}
    \label{tab:cvp_bs}
  \end{center}
\end{table}

\clearpage
\subsection{Metrics}
\label{sec:exp:metrics}

Multiple metrics have been used to understand the various methods being tested. They are briefly described below. These are evaluated for each test point in each fold of the cross validation test. The result would then be represented by the mean and standard deviations of all values across all folds. 

\begin{enumerate}
\item \emph{Squared Error (SE):}
  This represents the squared difference between the predicted concentration and the known concentrations for the set of test points. The mean over the set of all test points (Mean Squared Error or MSE) is the most popular metric for the context of this paper. Referring Equations \ref{eqn:gpmean} and \ref{eqn:gpcov}, for the $i^{th}$ test point,
  $$
  SE(i) \;=\; (\bar{f}_*(i) - z_i)^2
  $$
\item \emph{Variance (VAR):}
  This represents the variance (uncertainty) in the predicted concentrations for the set of test points. a lower VAR is a good outcome, only if the SE is also low. A model that has high SE and low VAR would be a poor model as this result would suggest that the model is confident of its inaccurate estimates. A better outcome would be a model with high SE and correspondingly high VAR i.e. a model that has inaccurate predictions but is also uncertain about these predictions.
\item \emph{Negative log probability / Log loss (NLP):}
  Inspired by \cite{rasmussen2006} (see page 23), this is a measure of the extent to which the model (including the GP model, kernel, parameters and evaluation data) explain the current test point. The lower the value of this metric, the better the model. 
  For the $i^{th}$ test point,
  $$
  NLP(i) \;=\; \frac{1}{2}log(2\pi\sigma_*^2) +  \frac{(\bar{f}_*(i) - z_i)^2}{2\sigma_*(i)^2}
  $$
\end{enumerate}

\subsection{Results}
\label{sec:exp:results}

\definecolor{Gray}{gray}{0.8}
\newcolumntype{g}{>{\columncolor{Gray}}c}
\begin{sidewaystable}
  \normalsize%\small%\footnotesize%\scriptsize
  \caption{E1 concentration estimation; 10 fold cross validation results using block sampling of various block sizes; Multi-task GP (MTGP) vs GP derived from MTGP (GP) vs Independently optimized GP (GPI) using Neural Network (NN), Matern 3/2 (MM), Squared exponential (SQEXP) and a Matern 3/2 - Matern 3/2 - Squared Exponential (MS) kernel combination on identical test data. The error metrics are expressed in squared units (squ).}
  \centering
  \begin{tabular}{|c*{13}{|g}|}
    \hline
    \hline
    \rowcolor{white}
    Block size & Method & \multicolumn{3}{|c|}{NN kernel} & \multicolumn{3}{|c|}{MM kernel} & \multicolumn{3}{|c|}{SQEXP kernel} & \multicolumn{3}{|c|}{MS kernel}\\
    \hline
    \rowcolor{white}
    (m) & & SE  & VAR & NLP & SE & VAR & NLP & SE & VAR & NLP & SE & VAR & NLP \\
    \rowcolor{white}
    & & (squ) & (squ) &  & (squ) & (squ) &  & (squ) & (squ) &  & (squ) & (squ) &  \\
    \rowcolor{white}
    & & mean & mean & mean & mean & mean & mean & mean & mean & mean & mean & mean & mean \\
    \rowcolor{white}
    & & (std) & (std) & (std) & (std) & (std) & (std) & (std) & (std) & (std) & (std) & (std) & (std) \\
    \hline
    & MTGP & 1.59  & 1.47 & 1.68 & 2.66 & 0.96 & 2.35 & 23.14 & 0.39 & 31.63 & 34.08 & 33.48 & 3.18 \\ 
    \rowcolor{white}
    &      & (7.25) & (0.28) & (2.75) & (8.66) & (0.15) & (5.01) & (55.62) & (0.10) & (76.20) & (76.16) & (1.48) & (1.14) \\
    22 x 11 x 2%
    & GP   & 36.52 & 14.93 & 3.56 & 41.42 & 5.96 & 5.77 & 44.43 & 0.51 & 45.37 & 43.49 & 36.98 & 3.30 \\ 
    \rowcolor{white}
    &      & (86.50) & (8.23) & (3.25) & (96.79) & (5.50) & (9.97) & (122.12) & (0.74) & (103.28) & (95.17) & (5.59) & (1.27) \\ 
    & GPI  & 41.28 & 73.23 & 3.34 & 45.26 & 76.44 & 3.38 & 52.96 & 86.80 & 3.45 & 45.26 & 76.44 & 3.38 \\ 
    \rowcolor{white}
    &      & (90.35) & (7.30) & (0.60) & (96.79) & (5.23) & (0.62) & (107.62) & (4.74) & (0.61) & (96.79) & (5.23) & (0.62) \\
    \hline
    & MTGP & 1.86 & 1.81 & 1.74 & 2.79 & 1.17 & 2.19 & 24.29 & 0.48 & 28.69 & 39.49  & 34.57  & 3.26 \\ 
    \rowcolor{white}
    &      & (9.49) & (0.59) & (2.66) & (10.51) & (0.24) & (4.53) & (57.54) & (0.20) & (68.94) & (87.09) & (1.71) & (1.27) \\
    44 x 22 x 4%
    & GP   & 52.75 & 26.55 & 3.60 & 65.00 & 13.91 & 5.05 & 76.89 & 1.07 & 50.71 & 57.55  & 40.75 & 3.46 \\ 
    \rowcolor{white}
    & & (124.18) & (17.09) & (2.72) & (149.17) & (14.52) & (7.27) & (235.51) & (3.92) & (113.15) & (124.37) & (10.77) & (1.48) \\
    & GPI  & 55.81 & 81.28 & 3.45 & 58.74 & 81.80 & 3.47 & 65.30 & 89.95 & 3.53 & 58.74  & 81.80  & 3.47 \\ 
    \rowcolor{white}
    &      & (119.69) & (11.10) & (0.71) & (124.07) & (8.02) & (0.74) & (132.02) & (6.61) & (0.72) & (124.07) & (8.02) & (0.74) \\
    \hline
    & MTGP & 3.38 & 3.24 & 1.91 & 7.50 & 1.65 & 3.04 & 30.08 & 0.85 & 21.35 & 52.13  & 36.90  & 3.42 \\ 
    \rowcolor{white}
    &      & (22.92) & (5.06) & (2.56) & (28.75) & (0.47) & (6.07) & (70.98) & (0.53) & (49.50) & (109.11) & (2.07) & (1.47) \\
    84 x 45 x 9%
    & GP   & 85.88 & 58.60 & 3.71 & 114.05 & 44.66 & 4.44 & 265.25 & 8.06 & 45.35 & 91.75  & 57.17  & 3.71 \\ 
    \rowcolor{white}
    & & (187.61) & (39.70) & (2.00) & (242.32) & (37.46) & (4.35) & (853.00) & (25.34) & (95.72) & (190.74) & (33.92) & (1.63)\\
    & GPI  & 85.53 & 100.04 & 3.63 & 86.81 & 97.42 & 3.64 & 91.63 & 101.22 & 3.67 & 86.81  & 97.42  & 3.64 \\ 
    \rowcolor{white}
    & & (171.85) & (21.34) & (0.81) & (175.15) & (15.97) & (0.85) & (179.65) & (13.42) & (0.85) & (175.15) & (15.97) & (0.85)\\
    \hline
    & MTGP & 14.60 & 10.97 & 2.24 & 32.05 & 2.24 & 7.29 & 52.96 & 1.54 & 19.27 & 83.79  & 39.05  & 3.81 \\ 
    \rowcolor{white}
    &      & (88.59) & (27.66) & (2.70) & (96.26) & (0.61) & (15.78) & (122.09) & (0.78) & (41.43) & (166.79) & (2.41) & (2.09) \\
    174 x 89 x 18%
    & GP   & 128.39 & 113.03 & 3.84 & 156.56 & 95.10 & 4.23 & 701.62 & 34.69 & 28.36 & 154.34  & 104.53  & 3.94 \\ 
    \rowcolor{white}
    & & (261.10) & (88.51) & (1.60) & (306.22) & (56.46) & (3.05) & (1787.63) & (59.68) & (59.58) & (319.96) & (106.83) & (1.58)\\
    & GPI  & 124.52 & 129.20 & 3.79 & 124.93 & 121.09 & 3.80 & 128.86 & 122.29 & 3.82 & 124.93  & 121.09  & 3.80 \\ 
    \rowcolor{white}
    & & (235.22) & (45.85) & (0.88) & (240.35) & (26.43) & (0.95) & (244.59) & (23.44) & (0.96) & (240.35) & (26.43) & (0.95)\\
    \hline
    & MTGP & 73.42 & 64.36 & 3.00 & 112.99 & 2.74 & 19.76 & 114.47 & 2.45 & 24.18 & 155.36  & 40.86  & 4.64 \\ 
    \rowcolor{white}
    &      & (213.89) & (86.94) & (2.09) & (206.86) & (0.64) & (32.27) & (214.94) & (0.98) & (42.85) & (257.28) & (2.50) & (3.08) \\
    348 x 177 x 35%
    & GP   & 204.57 & 249.34 & 4.06 & 215.40 & 153.09 & 4.25 & 1091.93 & 124.02 & 16.84 & 290.14  & 329.34  & 4.23 \\ 
    \rowcolor{white}
    & & (387.09) & (232.37) & (1.39) & (373.05) & (71.73) & (2.51) & (2801.29) & (94.50) & (43.56) & (541.45) & (376.16) & (1.21)\\
    & GPI  & 189.14 & 199.86 & 3.98 & 189.21 & 151.49 & 4.00 & 190.92 & 155.24 & 4.01 & 189.21  & 151.49  &  4.00\\ 
    \rowcolor{white}
    & & (335.76) & (120.66) & (0.90) & (327.56) & (45.31) & (1.04) & (331.84) & (44.34) & (1.03) & (327.56) & (45.31) & (1.04)\\
    \hline
    & MTGP & 180.64 & 173.67 & 3.61 & 206.18 & 2.97 & 33.83 & 214.11 & 2.89 & 37.05 & 243.80  & 41.95  & 5.66 \\ 
    \rowcolor{white}
    & & (368.05) & (154.95) & (1.56) & (349.73) & (0.52) & (54.10) & (357.84) & (0.84) & (58.21) & (380.23) & (2.21) & (4.46)\\
    696 x 353 x 70%
    & GP   & 325.98 & 562.94 & 4.31 & 301.60 & 192.63 & 4.43 & 871.72 & 185.80 & 12.58 & 460.23 & 976.69 & 4.55 \\ 
    \rowcolor{white}
    & & (546.05) & (523.54) & (1.19) & (452.75) & (64.46) & (2.29) & (2481.90) & (93.03) & (39.08) & (775.98) & (915.85) & (1.01)\\
    & GPI  & 291.72 & 362.43 & 4.19 & 282.28 & 180.05 & 4.23 & 283.43 & 183.92 & 4.23 & 282.28  & 180.05  & 4.23 \\ 
    \rowcolor{white}
    & & (465.37) & (271.48) & (0.80) & (428.51) & (49.34) & (1.12) & (430.84) & (49.45) & (1.11) & (428.51) & (49.34) & (1.12)\\
    \hline
    \hline
  \end{tabular}
  \label{tab:cvp_e1}
\end{sidewaystable}

\definecolor{Gray}{gray}{0.8}
\newcolumntype{g}{>{\columncolor{Gray}}c}
\begin{sidewaystable}
  \normalsize%\small%\footnotesize%\scriptsize
  \caption{E2 concentration estimation; 10 fold cross validation results using block sampling of various block sizes; Multi-task GP (MTGP) vs GP derived from MTGP (GP) vs Independently optimized GP (GPI) using Neural Network (NN), Matern 3/2 (MM), Squared exponential (SQEXP) and a Matern 3/2 - Matern 3/2 - Squared Exponential (MS) kernel combination on identical test data. The error metrics are expressed in squared units (squ).}
  \centering
  \begin{tabular}{|c*{13}{|g}|}
    \hline
    \hline
    \rowcolor{white}
    Block size & Method & \multicolumn{3}{|c|}{NN kernel} & \multicolumn{3}{|c|}{MM kernel} & \multicolumn{3}{|c|}{SQEXP kernel} & \multicolumn{3}{|c|}{MS kernel}\\
    \hline
    \rowcolor{white}
    (m) & & SE & VAR & NLP & SE & VAR & NLP & SE & VAR & NLP & SE & VAR & NLP \\
    \rowcolor{white}
    & & (squ) & (squ) &  & (squ) & (squ) &  & (squ) & (squ) &  & (squ) & (squ) &  \\
    \rowcolor{white}
    & & mean & mean & mean & mean & mean & mean & mean & mean & mean & mean & mean & mean \\
    \rowcolor{white}
    & & (std) & (std) & (std) & (std) & (std) & (std) & (std) & (std) & (std) & (std) & (std) & (std) \\
    \hline
    & MTGP & 3.85 & 3.13 & 2.10 & 3.88 & 2.02 & 2.23 & 52.32 & 0.04 & 650.87 & 36.09  &  39.09 & 3.20 \\ 
    \rowcolor{white}
    &      & (17.59) & (0.53) & (2.87) & (18.47) & (0.36) & (4.74) & (231.51) & (0.10) & (1848.02) & (102.25) & (2.48) & (1.27) \\
    22x11x2%
    & GP   & 36.86 & 19.44 & 3.34 & 40.79 & 8.45 & 4.37 & 75.67 & 0.05 & 668.93 & 46.60  &  43.34 & 3.31 \\ 
    \rowcolor{white}
    & & (118.99) & (8.89) & (3.10) & (129.77) & (6.63) & (7.41) & (869.23) & (0.30) & (1880.52) & (134.94) & (6.65) & (1.36) \\
    & GPI  & 49.79 & 93.47 & 3.44 & 53.61 & 97.26 & 3.47 & 60.25 & 85.95 & 3.48 &  53.61 & 97.26  & 3.47 \\ 
    \rowcolor{white}
    &      & (143.90) & (7.40) & (0.71) & (151.69) & (5.97) & (0.72) & (161.56) & (5.68) & (0.86) & (151.69) & (5.97) & (0.72) \\
    \hline
    & MTGP & 4.79 & 3.80 & 2.20 & 4.72 & 2.53 & 2.29 & 88.25 & 0.10 & 657.09 & 42.82  & 40.31  & 3.28 \\ 
    \rowcolor{white}
    &      & (22.21) & (1.06) & (2.87) & (22.80) & (0.58) & (4.43) & (352.70) & (0.30) & (1840.03) & (117.40) & (2.88) & (1.40) \\
    44x22x4%
    & GP   & 55.51 & 32.14 & 3.49 & 64.20 & 18.19 & 4.19 & 181.27 & 0.26 & 694.20 & 64.50  & 47.76  & 3.46 \\ 
    \rowcolor{white}
    & & (174.24) & (18.07) & (2.70) & (192.62) & (17.10) & (5.58) & (1385.73) & (2.48) & (1901.88) & (186.52) & (12.97) & (1.58) \\
    & GPI  & 68.69 & 101.11 & 3.55 & 71.21 & 102.80 & 3.56 & 77.86 & 89.38 & 3.57 & 71.21  & 102.80  & 3.56 \\ 
    \rowcolor{white}
    & & (195.85) & (10.86) & (0.86) & (201.30) & (8.82) & (0.88) & (211.96) & (9.01) & (1.04) & (201.30) & (8.82) & (0.88)\\
    \hline
    & MTGP & 8.04 & 6.22 & 2.37 & 10.97 & 3.69 & 2.81 & 211.05 & 0.56 & 461.55 & 56.72  & 42.90  & 3.44 \\ 
    \rowcolor{white}
    &      & (40.48) & (6.00) & (2.71) & (44.78) & (1.12) & (5.02) & (753.57) & (1.15) & (1290.67) & (146.64) & (3.41) & (1.64) \\
    84x45x9%
    & GP   & 95.98 & 66.23 & 3.71 & 116.98 & 54.56 & 4.05 & 1140.18 & 4.72 & 532.52 & 105.69  & 67.37  & 3.69 \\ 
    \rowcolor{white}
    & & (274.95) & (41.41) & (2.27) & (318.33) & (43.24) & (3.56) & (8156.99) & (22.50) & (1420.66) & (288.37) & (42.25) & (1.70)\\
    & GPI  & 105.20 & 119.02 & 3.72 & 105.27 & 119.32 & 3.72 & 115.96 & 103.63 & 3.74 & 105.27  & 119.32  & 3.72 \\ 
    \rowcolor{white}
    & & (277.85) & (20.10) & (1.03) & (279.95) & (17.19) & (1.06) & (301.71) & (23.63) & (1.22) & (279.95) & (17.19) & (1.06)\\
    \hline
    & MTGP & 21.49 & 15.88 & 2.66 & 37.85 & 5.09 & 4.85 & 402.24 & 1.92 & 228.66 & 90.32  & 45.49  & 3.79 \\ 
    \rowcolor{white}
    &      & (102.60) & (29.16) & (2.61) & (117.51) & (1.44) & (8.77) & (1144.48) & (2.06) & (752.97) & (211.07) & (3.73) & (2.23)\\
    174x89x18%
    & GP   & 142.62 & 123.42 & 3.88 & 165.30 & 112.93 & 4.07 & 3510.60 & 25.86 & 312.01 & 170.93  & 125.06 & 3.91 \\ 
    \rowcolor{white}
    & & (356.16) & (91.33) & (1.88) & (394.82) & (64.77) & (2.70) & (14425.39) & (59.06) & (914.26) & (420.58) & (134.64) & (1.59) \\
    & GPI  & 148.71 & 146.71 & 3.88 & 147.83 & 145.39 & 3.88 & 164.86 & 133.30 & 3.92 & 147.83  & 145.39  & 3.88 \\ 
    \rowcolor{white}
    & & (347.79) & (41.34) & (1.09) & (351.45) & (28.92) & (1.12) & (376.69) & (41.47) & (1.25) & (351.45) & (28.92) & (1.12)\\
    \hline
    & MTGP & 82.02 & 72.71 & 3.23 & 119.86 & 6.26 & 10.28 & 419.72 & 4.61 & 134.53 & 167.10 & 48.08 & 4.54 \\ 
    \rowcolor{white}
    & & (233.18) & (90.66) & (2.04) & (236.67) & (1.52) & (16.06) & (1196.06) & (2.90) & (599.11) & (320.09) & (4.15) & (3.23)\\
    348x177x35%
    & GP   & 219.37 & 265.84 & 4.09 & 232.89 & 178.16 & 4.17 & 8397.84 & 114.38 & 164.10 & 314.92  & 414.05 & 4.23 \\ 
    \rowcolor{white}
    & & (484.76) & (239.30) & (1.71) & (475.45) & (81.64) & (2.19) & (28045.42) & (102.39) & (633.41) & (689.64) & (500.88) & (1.24) \\
    & GPI  & 213.44 & 213.16 & 4.05 & 216.33 & 182.88 & 4.06 & 234.24 & 194.54 & 4.11 & 216.33 & 182.88 & 4.06 \\ 
    \rowcolor{white}
    & & (442.90) & (107.76) & (1.09) & (443.91) & (51.20) & (1.18) & (459.31) & (59.86) & (1.17) & (443.91) & (51.20) & (1.18) \\
    \hline
    & MTGP & 196.72 & 189.18 & 3.75 & 227.42 & 6.86 & 17.23 & 420.95 & 6.14 & 94.13 & 273.23 & 49.71 & 5.58 \\ 
    \rowcolor{white}
    & & (379.37) & (162.78) & (1.48) & (370.75) & (1.26) & (24.68) & (1016.76) & (2.62) & (458.35) & (440.21) & (4.07) & (4.37)\\
    696x353x70%
    & GP   & 340.01 & 594.95 & 4.34 & 331.10 & 223.12 & 4.37 & 5910.35 & 189.13 & 105.40 & 493.28  & 1309.98 & 4.59 \\ 
    \rowcolor{white}
    & & (603.14) & (544.29) & (1.33) & (534.42) & (73.21) & (1.77) & (23857.18) & (106.28) & (514.02) & (924.16) & (1309.73) & (0.96)\\
    & GPI  & 314.51 & 365.16 & 4.24 & 317.37 & 218.58 & 4.27 & 331.32 & 236.80 & 4.32 & 317.37  & 218.58  & 4.27 \\ 
    \rowcolor{white}
    & & (529.84) & (250.80) & (0.91) & (519.71) & (58.44) & (1.15) & (531.79) & (60.61) & (1.13) & (519.71) & (58.44) & (1.15)\\
    \hline
    \hline
  \end{tabular}
  \label{tab:cvp_e2}
\end{sidewaystable}

\definecolor{Gray}{gray}{0.8}
\newcolumntype{g}{>{\columncolor{Gray}}c}
\begin{sidewaystable}
  \normalsize%\small%\footnotesize%\scriptsize
  \caption{E3 concentration estimation; 10 fold cross validation results using block sampling of various block sizes; Multi-task GP (MTGP) vs GP derived from MTGP (GP) vs Independently optimized GP (GPI) using Neural Network (NN), Matern 3/2 (MM), Squared exponential (SQEXP) and a Matern 3/2 - Matern 3/2 - Squared Exponential (MS) kernel combination on identical test data.  The error metrics are expressed in squared units (squ).}
  \centering
  \begin{tabular}{|c*{13}{|g}|}
    \hline
    \hline
    \rowcolor{white}
    Block size & Method & \multicolumn{3}{|c|}{NN kernel} & \multicolumn{3}{|c|}{MM kernel} & \multicolumn{3}{|c|}{SQEXP kernel} & \multicolumn{3}{|c|}{MS kernel}\\
    \hline
    \rowcolor{white}
    (m) & & SE & VAR & NLP & SE & VAR & NLP & SE & VAR & NLP & SE & VAR & NLP \\
    \rowcolor{white}
    & & (squ) & (squ) &  & (squ) & (squ) &  & (squ) & (squ) &  & (squ) & (squ) &  \\
    \rowcolor{white}
    & & mean & mean & mean & mean & mean & mean & mean & mean & mean & mean & mean & mean \\
    \rowcolor{white}
    & & (std) & (std) & (std) & (std) & (std) & (std) & (std) & (std) & (std) & (std) & (std) & (std) \\
    \hline
    & MTGP & 1.60 & 1.29 & 1.64 & 1.29 & 0.98 & 1.57 & 7.30 & 0.64 & 6.50 & 19.15 & 36.32  & 2.98 \\ 
    \rowcolor{white}
    &      & (6.82) & (0.35) & (2.49) & (5.42) & (0.14) & (2.84) & (20.47) & (0.10) & (16.50) & (45.80) & (0.17) & (0.63) \\
    22 x 11 x 2%
    & GP   & 9.09 & 3.84 & 2.83 & 9.65 & 2.36 & 3.57 & 10.47 & 0.75 &  8.21 & 19.18  & 36.47  & 2.98 \\ 
    \rowcolor{white}
    &      & (26.21) & (1.76) & (3.67) & (27.41) & (1.44) & (6.46) & (29.99) & (0.40) & (21.19) & (45.75) & (0.19) & (0.63) \\
    & GPI  & 9.69 & 14.64 & 2.58 & 11.19 & 16.90 & 2.66 & 12.06 & 18.46 & 2.70 & 12.06  & 18.46  &  2.70\\ 
    \rowcolor{white}
    &      & (27.69) & (3.24) & (0.90) & (31.47) & (1.16) & (0.94) & (33.46) & (1.15) & (0.91) & (33.46) & (1.15) & (0.91) \\
    \hline
    & MTGP & 1.90 & 1.68 & 1.71 & 1.50 & 1.19 & 1.62 & 7.99 & 0.73 & 6.49 & 22.53  & 36.35  & 3.03 \\ 
    \rowcolor{white}
    &      & (8.29) & (0.66) & (2.17) & (7.41) & (0.23) & (2.92) & (21.94) & (0.19) & (16.04) & (52.33) & (0.20) & (0.72) \\
    44 x 22 x 4%
    & GP   & 12.80 & 6.38 & 2.91 & 14.47 & 4.52 & 3.50 & 15.59 & 1.07 & 9.56 & 22.56  & 36.51  &  3.03\\ 
    \rowcolor{white}
    &      & (37.03) & (3.66) & (3.17) & (40.66) & (3.53) & (5.34) & (45.59) & (1.45) & (23.82) & (52.27) & (0.23) & (0.72) \\
    & GPI  & 12.88 & 16.79 & 2.69 & 14.16 & 18.14 & 2.76 & 14.92 & 19.41 & 2.79 & 14.92  & 19.41  & 2.79 \\ 
    \rowcolor{white}
    &      & (36.55) & (4.65) & (1.00) & (39.23) & (1.79) & (1.08) & (40.78) & (1.73) & (1.05) & (40.78) & (1.73) & (1.05) \\
    \hline
    & MTGP & 2.87 & 2.79 & 1.88 & 3.08 & 1.65 & 1.99 & 9.19 & 1.09 & 5.46 & 29.82  & 36.43  & 3.13 \\ 
    \rowcolor{white}
    &      & (13.66) & (1.97) & (2.06) & (12.59) & (0.42) & (3.53) & (24.11) & (0.48) & (12.14) & (63.83) & (0.31) & (0.88) \\
    84 x 45 x 9%
    & GP   & 20.35 & 13.22 & 3.01 & 24.26 & 12.01 & 3.33 & 35.53 & 3.73 & 9.73 & 29.82  & 36.61  & 3.13 \\ 
    \rowcolor{white}
    &      & (55.84) & (8.54) & (2.45) & (64.08) & (8.52) & (3.63) & (99.80) & (7.16) & (21.78) & (63.64) & (0.36) & (0.87) \\
    & GPI  & 19.56 & 21.69 & 2.86 & 20.50 & 21.61 & 2.92 & 21.21 & 22.73 & 2.94 & 21.21  & 22.73  &  2.94\\ 
    \rowcolor{white}
    &      & (51.95) & (8.26) & (1.08) & (53.15) & (3.60) & (1.19) & (54.63) & (3.76) & (1.17) & (54.63) & (3.76) & (1.17) \\
    \hline
    & MTGP & 6.63 & 6.00 & 2.17 & 8.67 & 2.19 & 3.07 & 14.56 & 1.73 & 5.51 & 39.96  & 36.66  & 3.27 \\ 
    \rowcolor{white}
    &      & (35.07) & (8.73) & (2.24) & (24.92) & (0.54) & (4.93) & (34.46) & (0.70) & (10.69) & (80.47) & (0.54) & (1.10) \\
    174 x 89 x 18%
    & GP   & 29.91 & 25.03 & 3.13 & 32.57 & 23.31 & 3.28 & 73.85 & 12.06 & 7.34 & 39.81  & 36.89  & 3.26 \\ 
    \rowcolor{white}
    &      & (81.55) & (19.65) & (2.01) & (79.36) & (12.64) & (2.58) & (189.74) & (14.90) & (14.06) & (79.82) & (0.66) & (1.08) \\
    & GPI  & 28.27 & 29.28 & 3.01 & 28.87 & 26.65 & 3.07 & 29.50 & 28.35 & 3.09 & 29.50  & 28.35  & 3.09 \\     
    \rowcolor{white}
    &      & (69.54) & (16.01) & (1.10) & (69.91) & (6.01) & (1.27) & (72.11) & (6.17) & (1.24) & (72.11) & (6.17) & (1.24) \\
    \hline
    & MTGP & 22.76 & 23.57 & 2.64 & 25.66 & 2.62 & 5.81 & 27.36 & 2.58 & 6.50 & 54.46 & 37.62 & 3.46 \\ 
    \rowcolor{white}
    &      & (76.19) & (36.17) & (1.91) & (50.80) & (0.57) & (8.42) & (55.29) & (0.89) & (10.74) & (99.06) & (1.59) & (1.32) \\
    348 x 177 x 35%
    & GP   & 48.12 & 55.21 & 3.32 & 47.08 & 34.95 & 3.39 & 96.50 & 33.92 & 5.49 & 54.31 & 38.13 & 3.45 \\ 
    \rowcolor{white}
    & & (113.43) & (53.40) & (1.82) & (100.71) & (15.85) & (2.25) & (228.76) & (22.17) & (10.17) & (98.51) & (2.22) & (1.30)\\
    & GPI  & 43.19 & 46.90 & 3.20 & 42.96 & 32.36 & 3.25 & 43.24 & 34.15 & 3.25 & 43.24 & 34.15  & 3.25 \\ 
    \rowcolor{white}
    &      & (97.09) & (36.64) & (1.15) & (92.63) & (10.28) & (1.40) & (93.69) & (10.70) & (1.35) & (93.69) & (10.70) & (1.35) \\
    \hline
    & MTGP & 50.93 & 81.32 & 3.10 & 43.57 & 2.86 & 8.55 & 46.94 & 3.00 & 8.98 & 66.54 & 40.95 & 3.58 \\ 
    \rowcolor{white}
    &      & (123.14) & (99.31) & (1.43) & (77.06) & (0.48) & (12.36) & (81.94) & (0.77) & (13.24) & (109.25) & (5.93) & (1.32) \\
    696 x 353 x 70%
    & GP   & 73.71 & 129.87 & 3.57 & 64.20 & 43.42 & 3.58 & 93.45 & 47.21 & 4.85 & 67.03 & 43.80 & 3.57 \\ 
    \rowcolor{white}
    & & (146.92) & (126.46) & (1.65) & (112.93) & (14.04) & (2.05) & (205.91) & (21.02) & (8.96) & (109.94) & (11.83) & (1.27)\\
    & GPI  & 65.75 & 83.71 & 3.41 & 61.34 & 38.04 & 3.48 & 61.26 & 39.36 & 3.47 & 61.26  & 39.36 & 3.47 \\ 
    \rowcolor{white}
    & & (122.28) & (72.39) & (1.03) & (108.31) & (10.97) & (1.39) & (108.52) & (10.98) & (1.34) & (108.52) & (10.98) & (1.34)\\
    \hline
    \hline
  \end{tabular}
  \label{tab:cvp_e3}
\end{sidewaystable}

\begin{figure}[htb]
  \begin{center}
    \subfigure[Predicted E1 concentrations over entire region superimposed with input data]{\includegraphics[width=\columnwidth]{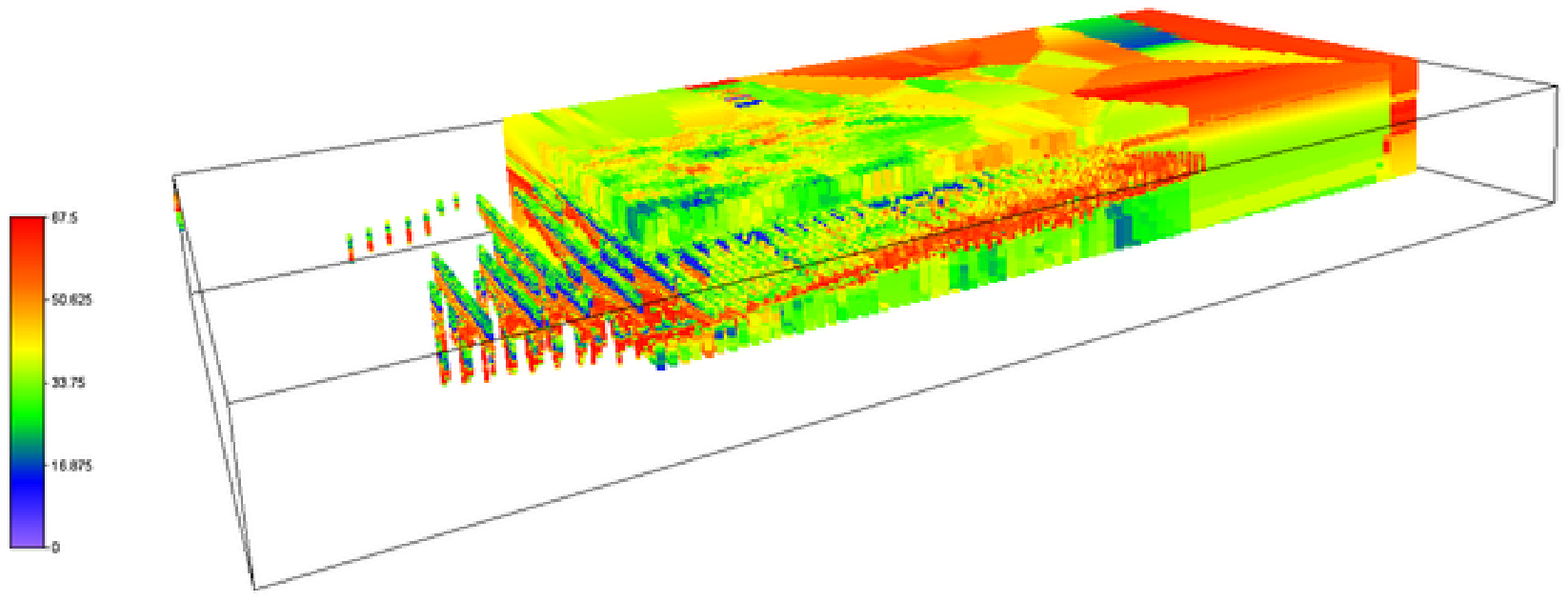}\label{fig:ope11}}
    \subfigure[2D section view of the predicted E1 concentrations]{\includegraphics[width=\columnwidth]{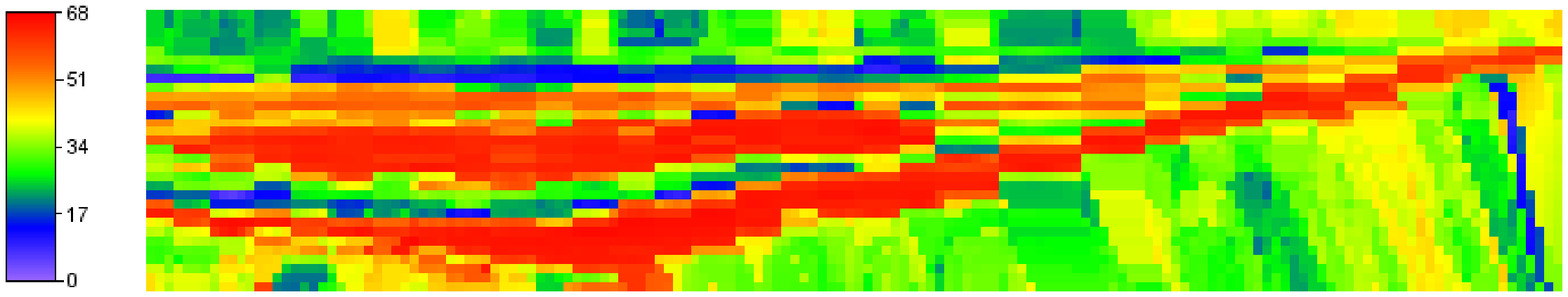}\label{fig:ope12}}
    \subfigure[Uncertainty in E1 predictions constituting the 2D section view]{\includegraphics[width=\columnwidth]{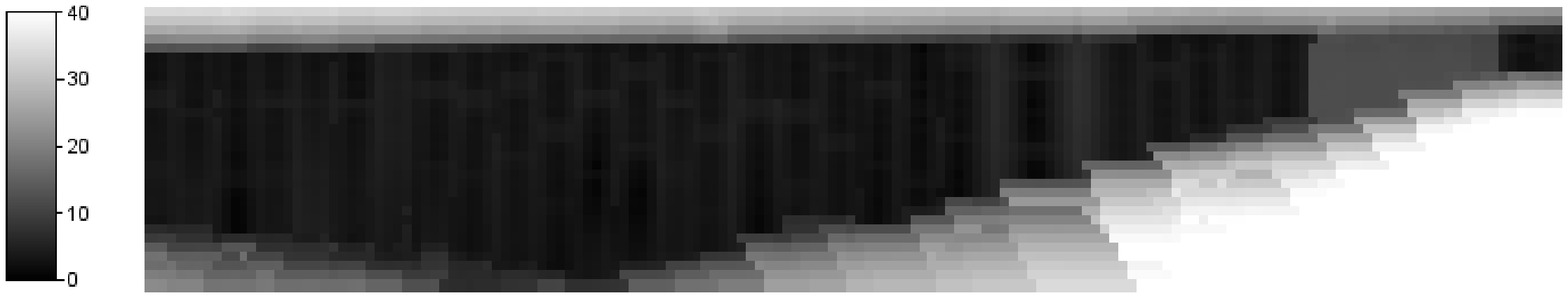}\label{fig:ope13}}
  \end{center}
  \caption{Figures \ref{fig:ope11}, \ref{fig:ope12} and \ref{fig:ope13} respectively show the predicted E1 concentrations (over the entire region) superimposed with the input data, a 2D section-view of the output data and the uncertainty in the predicted concentrations for the 2D view. Expectedly, the uncertainty is low around regions where input/given data exist and rapidly rises for predictions away from such areas - typically, the fringe areas. The 2D section view shows two red regions corresponding two regions of high E1 concentration. The corresponding regions in Figures \ref{fig:ope2} and \ref{fig:ope3} show low E2 and E3 concentrations respectively.}
  \label{fig:ope1}
\end{figure}

\begin{figure}[htb]
  \begin{center}
    \subfigure[Predicted E2 concentrations over entire region superimposed with input data]{\includegraphics[width=\columnwidth]{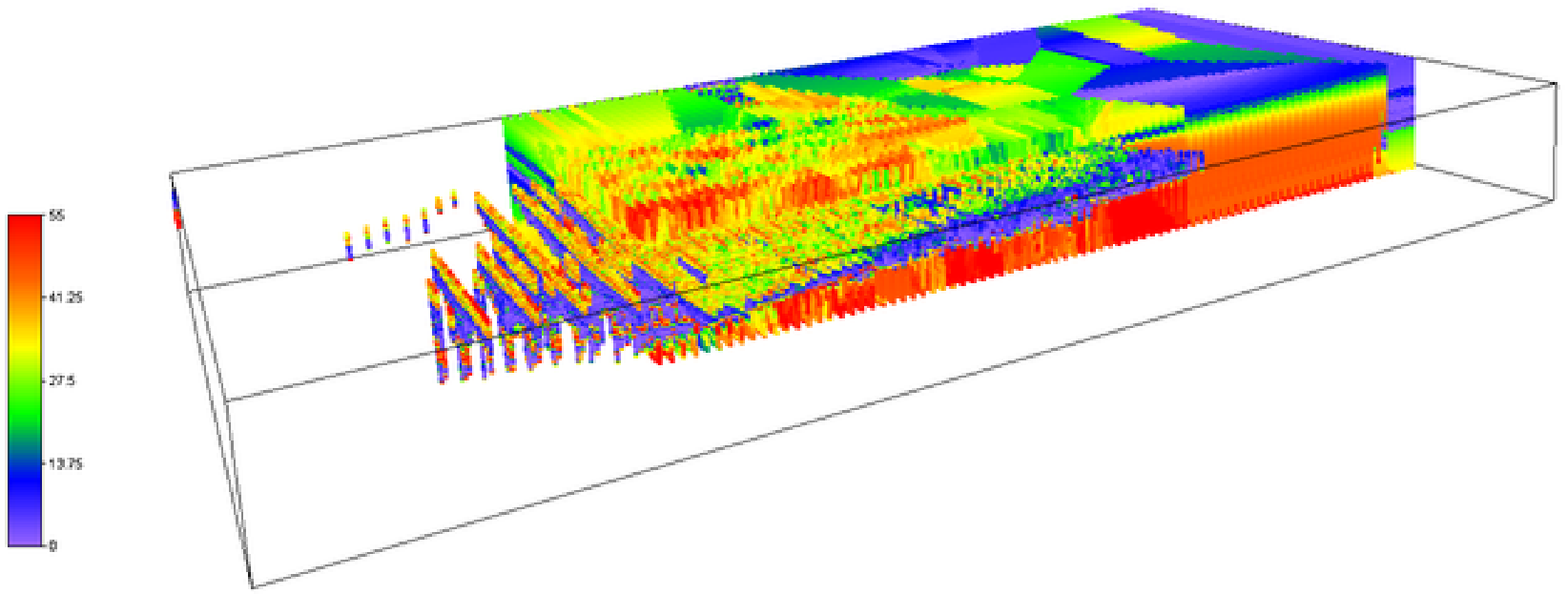}\label{fig:ope21}}
    \subfigure[2D section view of the predicted E2 concentrations]{\includegraphics[width=\columnwidth]{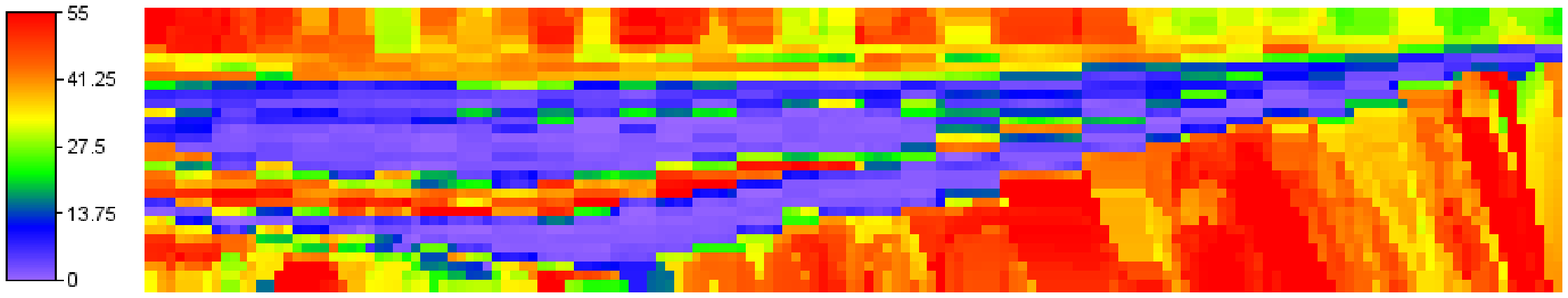}\label{fig:ope22}}
    \subfigure[Uncertainty in E2 predictions constituting the 2D section view]{\includegraphics[width=\columnwidth]{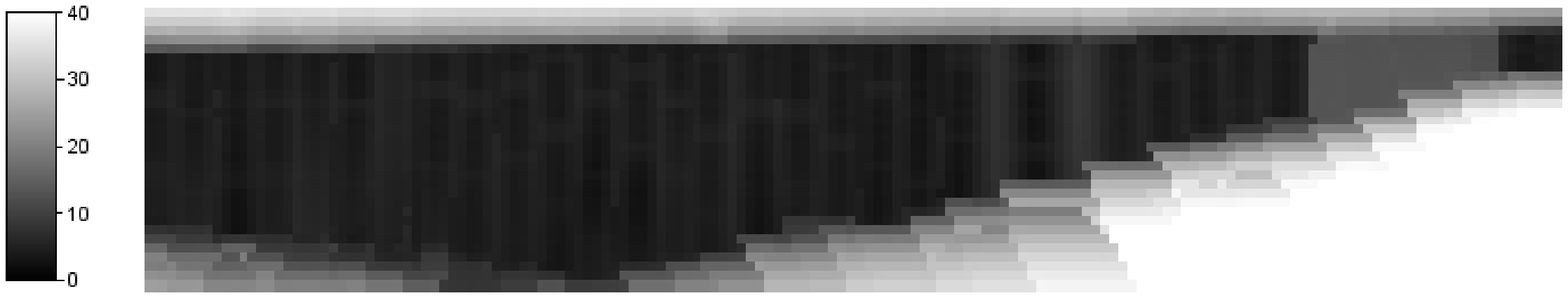}\label{fig:ope23}}
  \end{center}
  \caption{Figures \ref{fig:ope21}, \ref{fig:ope22} and \ref{fig:ope23} respectively show the predicted E2 concentrations (over the entire region) superimposed with the input data, a 2D section-view of the output data and the uncertainty in the predicted concentrations for the 2D view. Expectedly, the uncertainty is low around regions where input/given data exist and rapidly rises for predictions away from such areas - typically, the fringe areas. The 2D section view shows two violet regions corresponding two regions of low E2 concentration. The corresponding regions in Figure \ref{fig:ope1} show high E1 concentration and those from Figure \ref{fig:ope3} show low E3 concentration.}
  \label{fig:ope2}
\end{figure}

\begin{figure}[htbp]
  \begin{center}
    \subfigure[Predicted E3 concentrations over entire region superimposed with input data]{\includegraphics[width=\columnwidth]{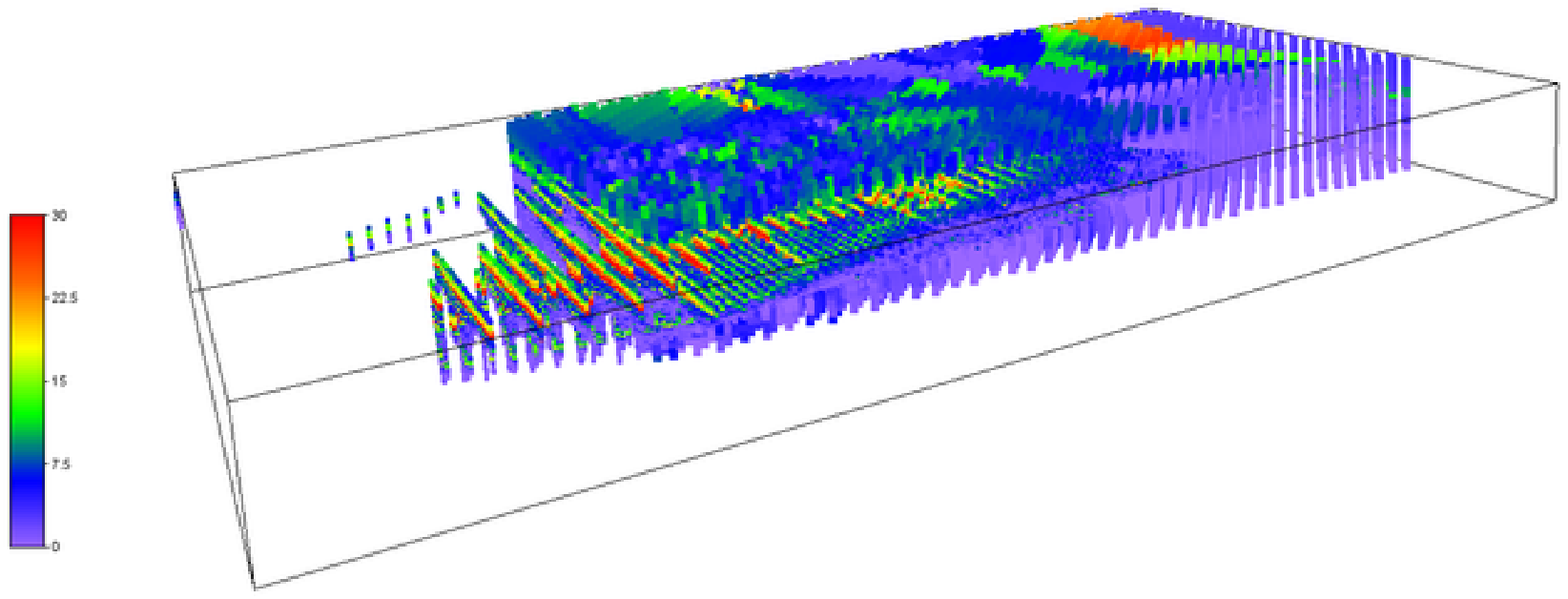}\label{fig:ope31}}
    \subfigure[2D section view of the predicted E3 concentrations]{\includegraphics[width=\columnwidth]{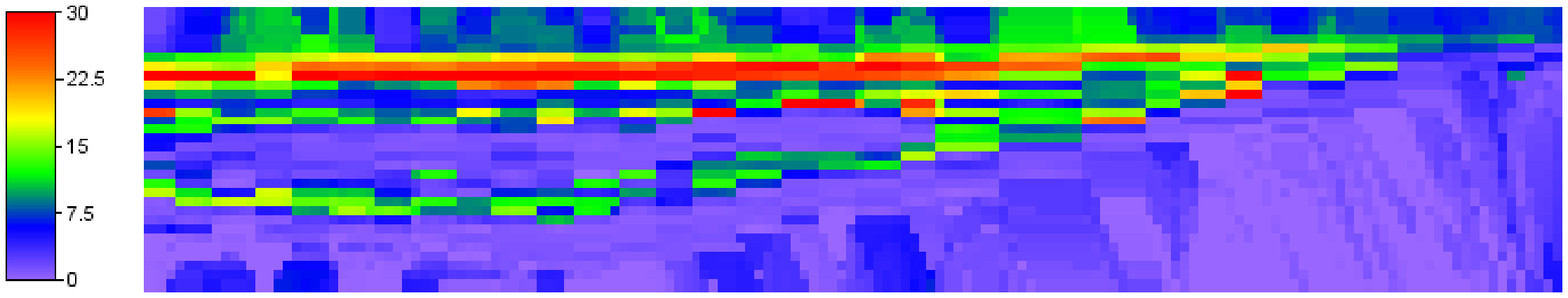}\label{fig:ope32}}
    \subfigure[Uncertainty in E3 predictions constituting the 2D section view]{\includegraphics[width=\columnwidth]{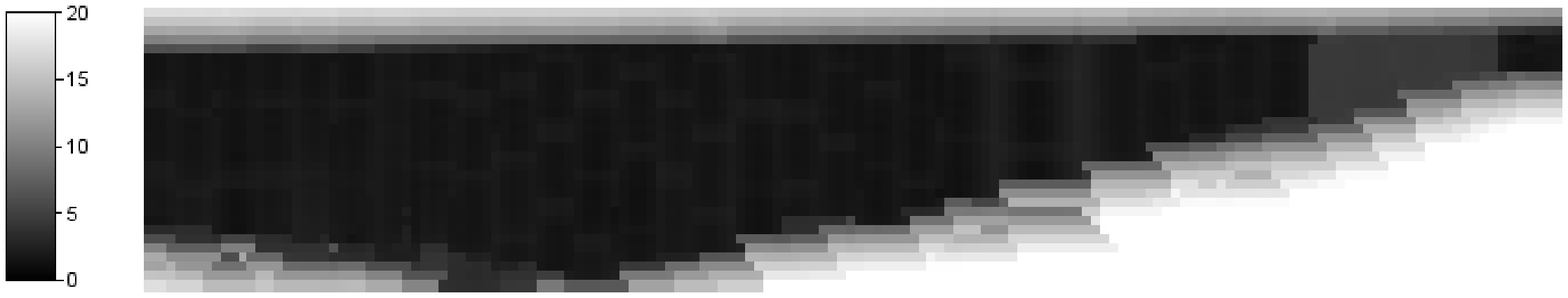}\label{fig:ope33}}
  \end{center}
  \caption{Figures \ref{fig:ope31}, \ref{fig:ope32} and \ref{fig:ope33} respectively show the predicted E3 concentrations (over the entire region) superimposed with the input data, a 2D section-view of the output data and the uncertainty in the predicted concentrations for the 2D view. Expectedly, the uncertainty is low around regions where input/given data exist and rapidly rises for predictions away from such areas - typically, the fringe areas. The 2D section view shows two violet regions corresponding two regions of low E3 concentration. The corresponding regions in Figure \ref{fig:ope1} show high E1 concentration and those from Figure \ref{fig:ope2} show low E2 concentration.}
  \label{fig:ope3}
\end{figure}

Figures \ref{fig:ope1}, \ref{fig:ope2} and \ref{fig:ope3} show the predicted concentrations of E1, E2 and E3 over the entire region of interest as well as 2D section views of this output and the uncertainty of the predictions that constitute it; these were produced using multi-task GPs using the Neural Network kernel. Tables \ref{tab:cvp_e1}, \ref{tab:cvp_e2} and \ref{tab:cvp_e3} show the results of the cross validation testing on the geological resource data set with the Neural Network (NN), Matern 3/2 (MM), Squared Exponential (SQEXP) and Matern 3/2 - Matern 3/2 - Squared exponential (MS) kernels. The three tables are visualized through numerous graphs that summarize the main trends observed; these are located in the appendix. Figures \ref{fig:e1_mtgp_se} through \ref{fig:e1_mtgp_gp_gpi_ms_nlp} depict the main results of Table \ref{tab:cvp_e1} (element E1), Figures \ref{fig:e2_mtgp_se} through \ref{fig:e2_mtgp_gp_gpi_ms_nlp} depict the main results of Table \ref{tab:cvp_e2} (element E2) and Figures \ref{fig:e3_mtgp_se} through \ref{fig:e3_mtgp_gp_gpi_ms_nlp} depict the main results of Table \ref{tab:cvp_e3} (element E3). The following observations were made from the results obtained.

\begin{enumerate}
\item Prediction error (SE) increases with increase in test block size.
  \begin{itemize}
  \item See Figures \ref{fig:e1_mtgp_se}, \ref{fig:e1_mtgp_gp_gpi_nn_se}, \ref{fig:e1_mtgp_gpi_mm_sqexp_se}, \ref{fig:e1_mtgp_gp_gpi_ms_se} for E1, Figures \ref{fig:e2_mtgp_se}, \ref{fig:e2_mtgp_gp_gpi_nn_se}, \ref{fig:e2_mtgp_gpi_mm_sqexp_se}, \ref{fig:e2_mtgp_gp_gpi_ms_se} for E2 and Figures \ref{fig:e3_mtgp_se}, \ref{fig:e3_mtgp_gp_gpi_nn_se}, \ref{fig:e3_mtgp_gpi_mm_sqexp_se}, \ref{fig:e3_mtgp_gp_gpi_ms_se} for E3.
  \item This behavior is expected. It happens because the support training data required for regressing at a test point is situated farther away. Increasing the test block size also results in reduced stratification as one fold of the cross validation may have e.g. 10,000 test points whereas another may have only 1000 points. This results in increased standard deviation of prediction error. A ten fold stratified cross validation is generally considered to be the most representative of performance measure \cite{kohavi1995study}, however testing multiple larger block sizes provides a better understanding of the model's behavior and robustness.
  \end{itemize}
\item NN kernel based MTGP/GP models trained faster than other kernels
  \begin{itemize}
  \item Further optimization of each of the MTGP/GP models could yield better results. The results shown are the result of a reasonable amount of optimization applied to each kernel and GP model. Typically, multiple attempts were performed and the best results obtained were pursued/used. One iteration consisted of a stochastic optimization step (simulated annealing) and/or a gradient based optimization step (Quasi Newton optimization with BFGS Hessian update) with 10,000 training data chosen uniformly from the data. This work uses a ``block-learning'' approximation \cite{vasudevan_iros2010} which approximates the total marginal likelihood as a sum of a sequence of marginal likelihoods computed over blocks of points comprising the training data. The size of the block is defined by the computational resources available. The stochastic optimization step was the most time consuming part; each attempt was started with completely random parameters. The code was unoptimized MATLAB code running typically on an 8-core processor based machine. Most times, not all the cores were used for the same process; multiple processes also shared the same system. Note that the experiments in this paper do not use analytical gradients for the optimization of the hyperparameters; this was a design choice made in the interest of stability and comparability of the optimization results across kernels. The use of analytical gradients can significantly reduce the total training time. Training time may also be reduced significantly by various other ways including other approximations, intelligently setting initial parameters, scaling the data etc.
    \begin{center}
      \begin{tabular}{| l | l | l |}
        \hline
        Model & Kernel & Number of training attempts, iterations \\
        &        & Total training time for successful attempt \\
        \hline
        \multirow{4}{*}{MTGP} & NN & 2 attempts, 3 iterations, total training time = 78.89 hours \\
        & MM & 3 attempts, 2 iterations, total training time = 222.15 hours \\
        & SQEXP & 4 attempts, 1 iteration, total training time = 92.52 hours \\
        & MS & 2 attempts, 3.5 iterations, 3 iterations took 113.69 hours \\
        \hline
        \multirow{4}{*}{GPI} & NN & 3 attempts, 2 iterations, total training time = 41.07 hours \\
        & MM & 2 attempts, 1 iteration, total training time = 48.91 hours \\
        & SQEXP & 2 attempts, 1 iteration, total training time = 47.67 hours \\
        \hline
      \end{tabular}
    \end{center}
    Rather than the individual training times, the relative amount of training (under similar conditions, with different kernel) required to produce a reasonable set of parameters is of more interest. Experience suggests that the NN kernel based MTGP/GP models converged faster and better as compared to other kernels.
  \end{itemize}
\item MTGP models based on the NN kernel outperform other kernels tested.
  \begin{itemize}
  \item See Figures \ref{fig:e1_mtgp_se}, \ref{fig:e1_mtgp_var}, \ref{fig:e1_mtgp_nlp} for E1, Figures \ref{fig:e2_mtgp_se}, \ref{fig:e2_mtgp_var}, \ref{fig:e2_mtgp_nlp} for E2 and Figures \ref{fig:e3_mtgp_se}, \ref{fig:e3_mtgp_var}, \ref{fig:e3_mtgp_nlp} for E3.
  \item The NN kernel is the best performing kernel of the four tested, across all block sizes tested. The MTGP based on the NN kernel produces lower SE (better estimate) and reduced NLP (better model) estimates than other kernels tested. 
  \item For small block sizes, both the NN and MM kernel are competitive; in case of E3, the MM even marginally outperforms the NN kernel for the two smallest block sizes tested. Note however that considering all test sizes and all three elements, the observation is that the MM kernel produces lower VAR for a higher SE, meaning that it is more confident of its SE values which are worse/higher than those of the NN kernel. This makes its NLP higher and the model poorer than an MTGP based on the NN kernel. Note also that as the test block size increases, the advantage in performance of the MTGP based on the NN kernel over that based on the MM kernel becomes more distinctive. Not only are the SE values smaller for the NN kernel, the NLP values remain in the same range whereas those of the MM kernel rise significantly. This proves that the MTGP-NN is better performing and more robust than the MTGP-MM. The latter property suggests that the MTGP-NN will be able to cope better with incomplete data sets.
  \item Both the MS and SQEXP kernels are not competitive with respect to the NN or MM kernels considering both the SE and NLP metrics. These kernels are discussed individually in the following paragraphs.
  \end{itemize}
\item MTGP models perform significantly better than three separate GPs (using the MTGP parameters) or three independently optimized GPs as information fusion improves estimation.
  \begin{itemize}
  \item See Figures \ref{fig:e1_mtgp_gp_gpi_nn_se} and \ref{fig:e1_mtgp_gp_gpi_nn_nlp} for E1, Figures \ref{fig:e2_mtgp_gp_gpi_nn_se} and \ref{fig:e2_mtgp_gp_gpi_nn_nlp} for E2 and Figures \ref{fig:e3_mtgp_gp_gpi_nn_se} and \ref{fig:e3_mtgp_gp_gpi_nn_nlp} for E3.
  \item For the NN kernel, the MTGP metrics are always lower than the corresponding derived GP (GP) or independent GP (GPI) metrics - lower SE (better estimate) with lower NLP (better model). This clearly demonstrates the benefits of information fusion across heterogeneous information sources so as to improve individual predictions using the MTGP model. 
  \item From Tables \ref{tab:cvp_e1}, \ref{tab:cvp_e2} and \ref{tab:cvp_e3}, the average reduction in error (i.e. improvement in performance) of MTGP models over GP/GPI models for the smallest, intermediate and largest test block sizes are - 
    \begin{itemize}
    \item E1
      \begin{itemize}
      \item 22 x 11 x 2 - 95.6\% over GP, 96.2\% over GPI
      \item 84 x 45 x 9 - 96.1\% over GP, 96.0\% over GPI
      \item 696 x 353 x 70 - 44.6\% over GP, 38.1\% over GPI
      \end{itemize}
    \item E2
      \begin{itemize}
      \item 22 x 11 x 2 - 89.6\% over GP, 92.3\% over GPI
      \item 84 x 45 x 9 - 91.6\%  over GP, 92.4\% over GPI
      \item 696 x 353 x 70 - 42.1\% over GP, 37.5\% over GPI
      \end{itemize}
    \item E3 
      \begin{itemize}
      \item 22 x 11 x 2 - 82.4\% over GP, 83.5\% over GPI
      \item 84 x 45 x 9 - 85.9\% over GP, 85.3\% over GPI
      \item 696 x 353 x 70 - 30.9\% over GP, 22.5\% over GPI
      \end{itemize}      
    \end{itemize}
  \end{itemize}
  These numbers demonstrate significant improvements in performance, even in very large test block sizes, when using the MTGP-NN model for correlated data.
\item The MS kernel was uncompetitive
  \begin{itemize}
  \item See Figures \ref{fig:e1_mtgp_gp_gpi_ms_se} and \ref{fig:e1_mtgp_gp_gpi_ms_nlp} for E1, Figures \ref{fig:e2_mtgp_gp_gpi_ms_se} and \ref{fig:e2_mtgp_gp_gpi_ms_nlp} for E2 and Figures \ref{fig:e3_mtgp_gp_gpi_ms_se} and \ref{fig:e3_mtgp_gp_gpi_ms_nlp} for E3.
  \item The MS kernel is not competitive with respect to the NN and MM kernels as discussed earlier. However, the MTGP using this kernel combination proves to be better than a derived GP and an independently optimized GP with respect to the SE metric. From the NLP perspective, the MTGP-MS model is more competitive than the other GP models for small block sizes. For larger block sizes, using an independently optimized GP proves to be a more trust worthy modeling option as the increase in error is met with a corresponding increase in uncertainty (hence low NLP) for the independent GP models. The exception to this behavior is seen in the results for E3, the MTGP model is poor in this case. This is attributed to do with inferior parameters relevant to the element E3 obtained from the optimization process. 
  \item The MS kernel performs better than the SQEXP with respect to the NLP metric and hence can be trusted more (prediction error compensated by prediction uncertainty), but in two of the three elements (E1 and E3), its SE was inferior to that of the SQEXP.
  \end{itemize}
\item The SQEXP kernel was uncompetitive and unreliable
  \begin{itemize}
  \item See Tables \ref{tab:cvp_e1}, \ref{tab:cvp_e2} and \ref{tab:cvp_e3}; see Figures \ref{fig:e1_mtgp_se}, \ref{fig:e1_mtgp_var} \ref{fig:e1_mtgp_nlp}, \ref{fig:e1_mtgp_gpi_mm_sqexp_se}, \ref{fig:e1_mtgp_gpi_mm_sqexp_nlp}, \ref{fig:e1_mtgp_gp_gpi_ms_se} and \ref{fig:e1_mtgp_gp_gpi_ms_nlp} for E1, Figures \ref{fig:e2_mtgp_se}, \ref{fig:e2_mtgp_var}, \ref{fig:e2_mtgp_nlp}, \ref{fig:e2_mtgp_gpi_mm_sqexp_se}, \ref{fig:e2_mtgp_gpi_mm_sqexp_nlp}, \ref{fig:e2_mtgp_gp_gpi_ms_se} and \ref{fig:e2_mtgp_gp_gpi_ms_nlp} for E2 and Figures \ref{fig:e3_mtgp_se}, \ref{fig:e3_mtgp_var}, \ref{fig:e3_mtgp_nlp}, \ref{fig:e3_mtgp_gpi_mm_sqexp_se}, \ref{fig:e3_mtgp_gpi_mm_sqexp_nlp}, \ref{fig:e3_mtgp_gp_gpi_ms_se} and \ref{fig:e3_mtgp_gp_gpi_ms_nlp} for E3.
  \item The MTGP-SQEXP model performs poorly in comparison with the equivalent models using the NN/MM kernels, with respect to both SE and NLP. 
  \item For elements E1 and E3, the MTGP-SQEXP has a better SE than the corresponding model based on the MS kernel; it has an SE better than the corresponding derived/independent GP models but an inferior (overconfident or low uncertainty) VAR and a fluctuating NLP trend. For element E2, the MTGP-SQEXP is worse off than both the equivalent model based on the MS kernel as well as its corresponding GP models. 
  \item Considering the results for E2, the NLP is directly proportional to the SE and inversely to the prediction variance. At the smallest block size, the MTGP-SQEXP produces relatively high SE (with respect to e.g. MTGP-NN) but very low prediction variance. This basically suggests that the model is confident of its poor estimates - a bad outcome. This results in a high NLP and poor model. As the block size increases, the prediction variance increases more relative to the prediction error resulting in the decreasing NLP trend. For elements E1 and E3, the largest block size results in a stronger increase in prediction error than the variance in the prediction resulting in an increase in NLP. Overall, the MTGP-SQEXP model is poor.
  \item  The SQEXP kernel is a limiting case of the MM kernel; both are stationary kernels. Considering the behavior of the GPI model using the SQEXP kernel and its competitive results with respect to those of the GPI-MM kernel, it is possible that the poor performance of the MTGP-SQEXP (as compared to the MTGP-MM) is due to poor optimization output (a bad local minima). Further investigation on this result is ongoing but the findings are not expected to change the conclusions of this paper.
  \end{itemize}  
\item In general, the stationary kernels tested seemed to have an inadequate increase in prediction uncertainty with increasing test block size and worsening predictions. This leads a higher NLP metric and a poor model that is overly confident of its worsening predictions. This behavior can be attributed to the correlation profile of the stationary kernels tested - they all share the ``correlation decreases with increasing distance of support data from point of interest'' trend. This results in stationary kernels not being able to cope with large test block sizes as the support data is farther away (i.e. less correlated and not of much use). In contrast, the nonstationary NN kernel has a sigmoidal profile that can handle this issue across a range of test block sizes.
\item The SE metric taken alone can be misleading. The experiments have reinforced the need for a multi-metric analysis. The SE metric only provides information on the prediction error but it does not describe the prediction uncertainty which is very important in understanding if a model is reliable or otherwise. The VAR and NLP metrics provided key insights on the difference in performance between different models and kernels. A model that is very confident of its poor predictions is unreliable (as was the case for the SQEXP kernel). Worsening predictions (due to increasing test block size) is itself not a bad outcome, provided it is met with an equivalent increase in prediction uncertainty. 
\end{enumerate}

\clearpage
\section{Discussion}
\label{sec:discussion}

On the basis of this study, an attempt is made in answering two fundamental questions - (1) How can I know if I have a good MTGP model ? and (2) Which GP model or kernel should I use ? By no means is this intended to be a ready-made prescription, universal formula or short-cut to be used as a substitute for context specific and statistically apt decisions in developing Gaussian process models. Rather, this is a reflection of the authors' experiences based on the scope of this and past work in other domains such as terrain modeling. Note that there are numerous very sophisticated GP techniques (kernels, approximation etc.) which are beyond the scope of this work and which may change some of these inferences.

\subsection{How can I know if I have a good MTGP model ?}

To effectively develop and validate MTGP models, the experiments are suggestive of the following -
\begin{enumerate}
\item The use of multiple kernel from the same family would provide a good method for validating the general behavior/trends of the model in question. For instance when developing an MTGP model based on the SQEXP kernel, developing a Matern 3/2 kernel based MTGP model could provide a means to validate the behavior of the MTGP-SQEXP model.
\item The model hyperparameter optimization performed in this paper is based on maximizing the marginal likelihood. Typically, error metrics such as the SE being sufficiently low is suggestive of the model being good. A cross validation test could also be performed to ensure that this is indeed the case. However, it is also important to check if the model in question is under/over confident (high/low uncertainty) for a given level of error. This can be done, not as a standalone test, but in comparison with alternative models or test cases.
\item When developing a MTGP model, it is a good idea to compare with an equivalent derived GP model and an independently optimized GP model. The availability of more information and the effective use of this information through the MTGP model should ideally result in significantly lower error metrics (e.g. SE) with a significant improvement in confidence (i.e. decrease in prediction variance, VAR) and a net reduction in the Negative Log Loss (NLP) metric.
\item It may be useful to design a variety of different test cases (e.g. different test block sizes) and check if the performance metrics behave as expected. Such a test would also be indicative of the robustness of the model.
\item It may be useful to optimize independent GP models for each task and use these hyperparameters as the initial parameters for the MTGP model.
\end{enumerate}

\subsection{Which GP model or kernel should I use ?}

This obviously depends on the data set at hand and the constraints of the modeling problem. The following are purely indicative, based on our experiences in multiple problem domains \cite{vasudevan_jfr2009,vasudevan2012,melkumyan2011} and may change considering alternative kernels, other novel ways of treating the modeling problem or approximation methods.
\begin{enumerate}
\item \textit{Time, complexity, computational resources are a premium. I need a method that just works}: Independently optimized GP models using the Neural Network kernel or the Matern 3/2 kernel would be a competitive solution. Note that the outcome will only be as good as the data being modeled and other information sources cannot be leveraged.
\item \textit{I need the best possible model over a range of test sizes and I do not know much about my data}: Multi-task GP models using the Neural Network kernel would be a competitive solution.
\item \textit{I need the best possible model over a range of test sizes and I know how my data changes}: Multi-task GP models with a kernel representative of the variation of the data e.g. a uniform variation (no sudden changes in trend) can be effectively modeled using the Matern 3/2 or Squared Exponential kernels.
\item \textit{I need a model that can cope with sparse data and/or incomplete data sets}: Neural network kernel based GP or MTGP models depending on the computational complexity constraints and model accuracy requirements.
\item \textit{I have ``good'' multi-attribute data. I need to model this well and fast}: Independent GP models for each of the attributes, using either a Neural Network kernel or some other kernel more suited to the data, would provide a competitive solution. The use of independent GP models will result in the ability to parallelize the modeling process and significantly reduce the possibility of poor models (poor local minima) as a consequence of a reduction in number of model parameters. Note that ``good'' here is application dependent but would certainly require being well sampled, not noisy and reasonably complete (no large gaps where other information modalities can be leveraged).
\end{enumerate}

\section{Conclusion}
\label{sec:conclusion}

This paper studied the problem of geological resource modeling using multi-task Gaussian processes (MTGPs). The concentrations of three elements were modeled and predicted over a region of interest using the MTGP as well as individual Gaussian processes (GPs) for each of these quantities. The paper demonstrates that MTGPs perform significantly better than individual GPs at the modeling problem as they effectively integrate heterogeneous sources of information (concentrations of individual elements) to improve the individual predictions of each of them. The benefits of information integration using the MTGP as against independent GPs for the task of geological resource modeling have been quantified by a multi-metric and multi-test-size cross validation study that performed both an exact and an independent comparison between MTGPs and GPs. Multi-task Gaussian process models based on the Neural Network kernel was shown to be a competitive and robust option across a range of test block sizes. 

\section*{Acknowledgements}
This work has been funded by the Rio Tinto Centre for Mine Automation.

%%%%%%%%%%%%%%%%%%%%%%%%%%%%%%%%%%%%%%%%%%%%%%%%%%%%%%%%%%%%%%%%%%%%%%%%

%%%%%%%%%%%%%%%%%%%%%%%%%%%%%%%%%%%%%%%%%%%%%%%%%%%%%%%%%%%%%%%%%%%%%%%%

\vspace{1cm}
\appendix
\section*{Appendix: Graphs of results obtained in Tables \ref{tab:cvp_e1}, \ref{tab:cvp_e2} and \ref{tab:cvp_e3}}
\clearpage

%% ge1
\begin{figure}[htbp]
  \begin{center}
    \subfigure{\includegraphics[width=0.78\columnwidth]{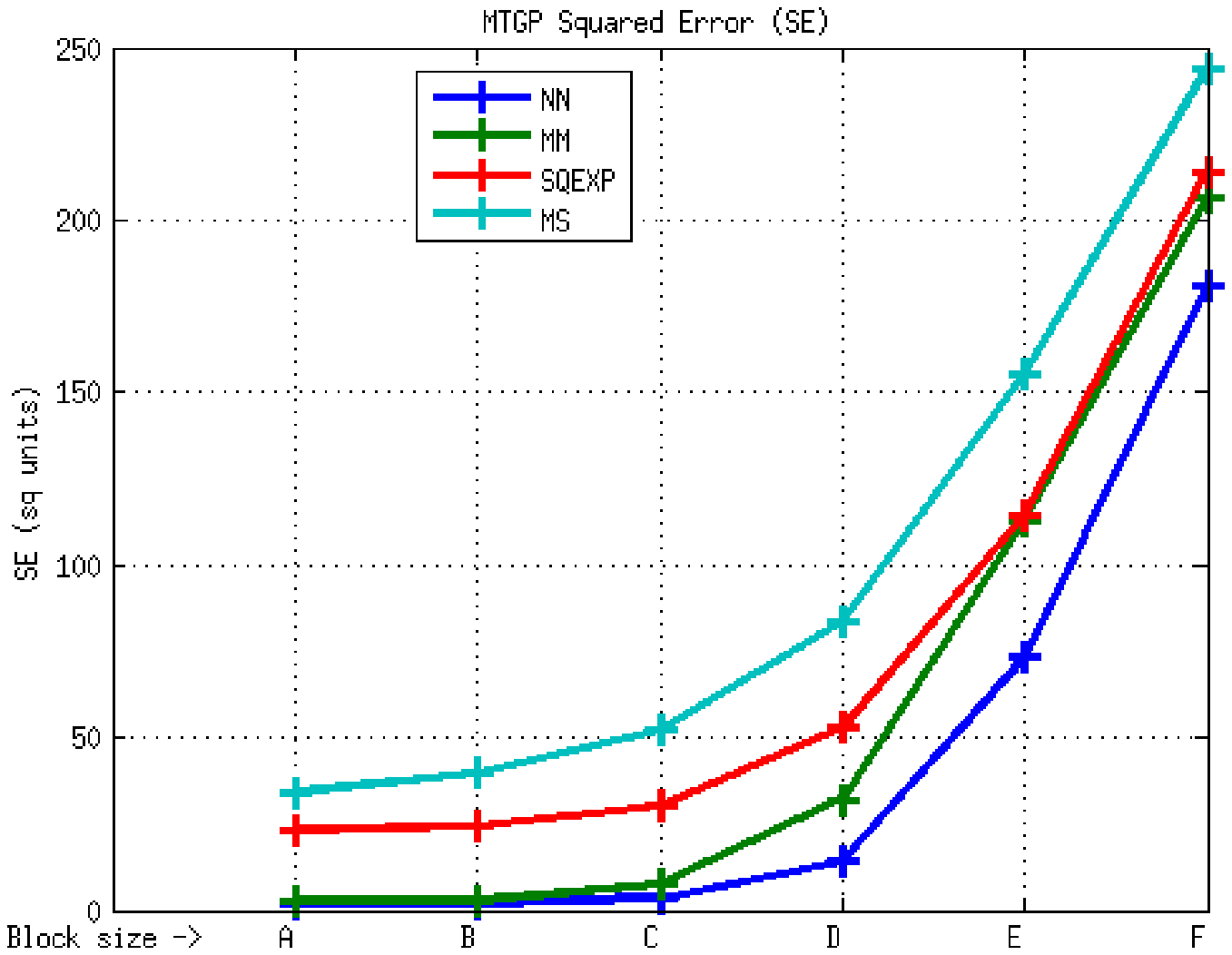}}
    \subfigure{\includegraphics[width=0.78\columnwidth]{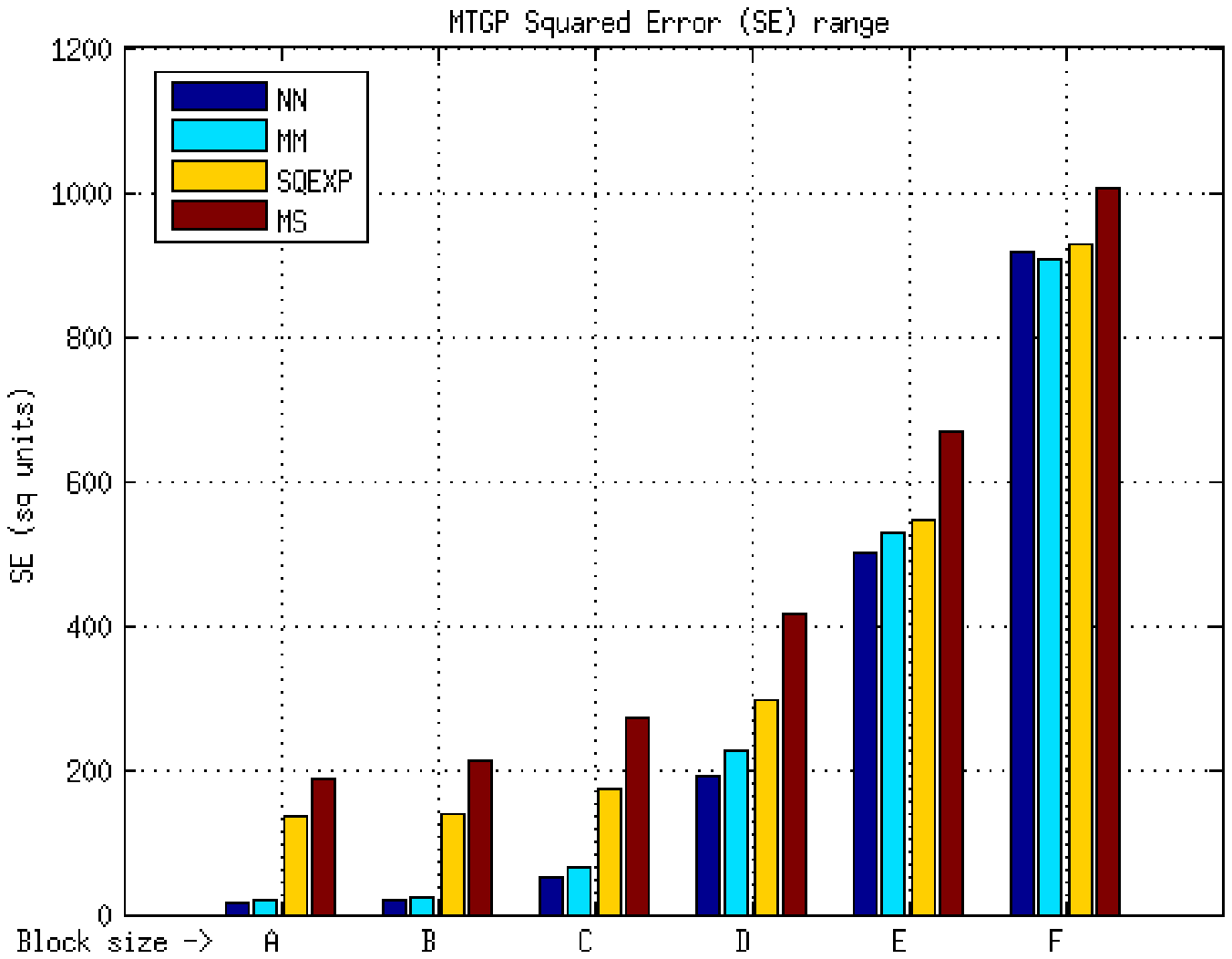}}
  \end{center}
  \caption{Element E1, MTGP approach, SE metric. The figure above shows the average values; the one below shows the range of values obtained considering two standard deviations about the mean. Test block sizes (m) - A (22 x 11 x 2), B (44 x 22 x 4), C (84 x 45 x 9), D (174 x 89 x 18), E (348 x 177 x 35) and F (696 x 353 x 70).}
  \label{fig:e1_mtgp_se}
\end{figure}

\begin{figure}[htbp]
  \begin{center}
    \subfigure{\includegraphics[width=0.78\columnwidth]{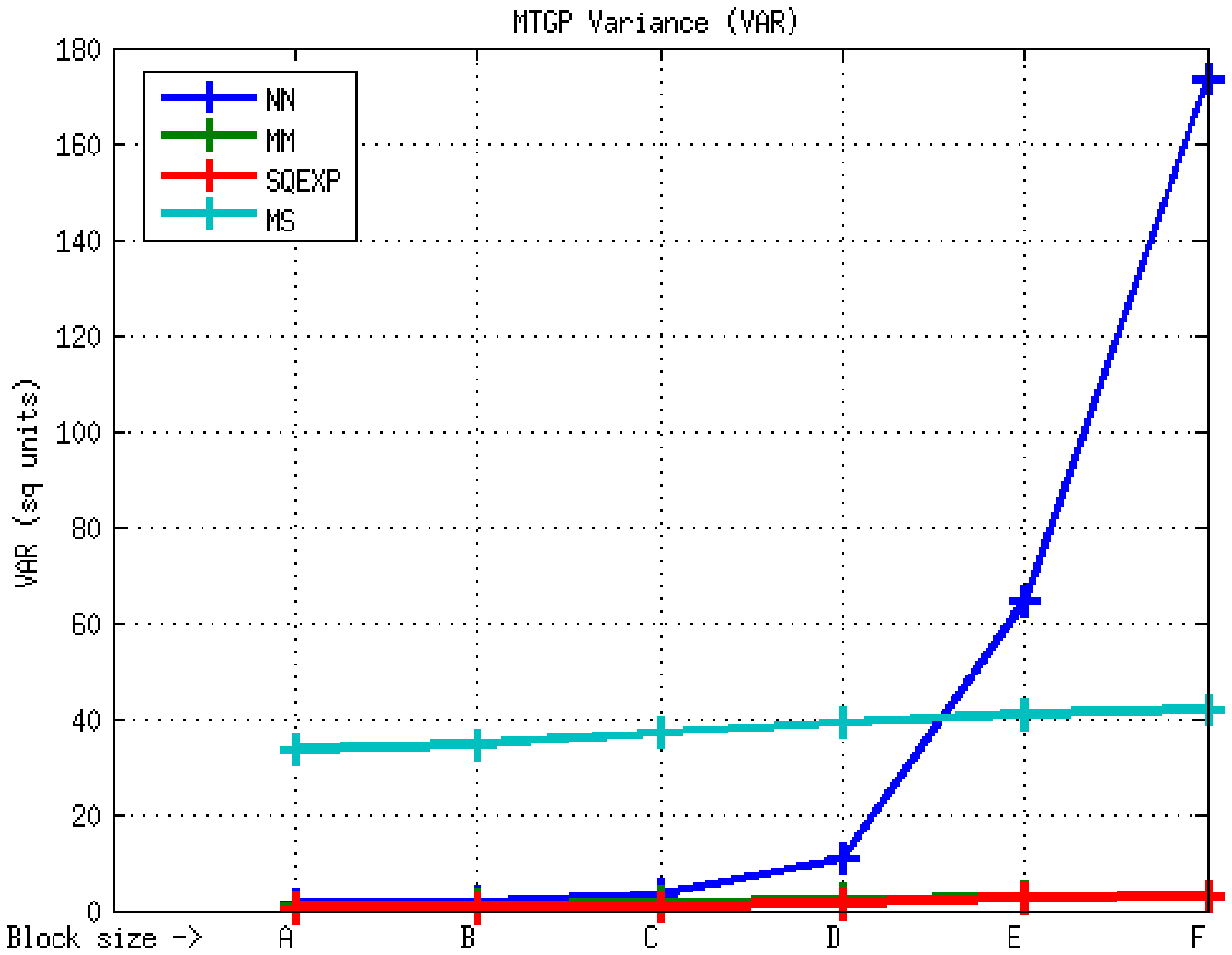}}
    \subfigure{\includegraphics[width=0.78\columnwidth]{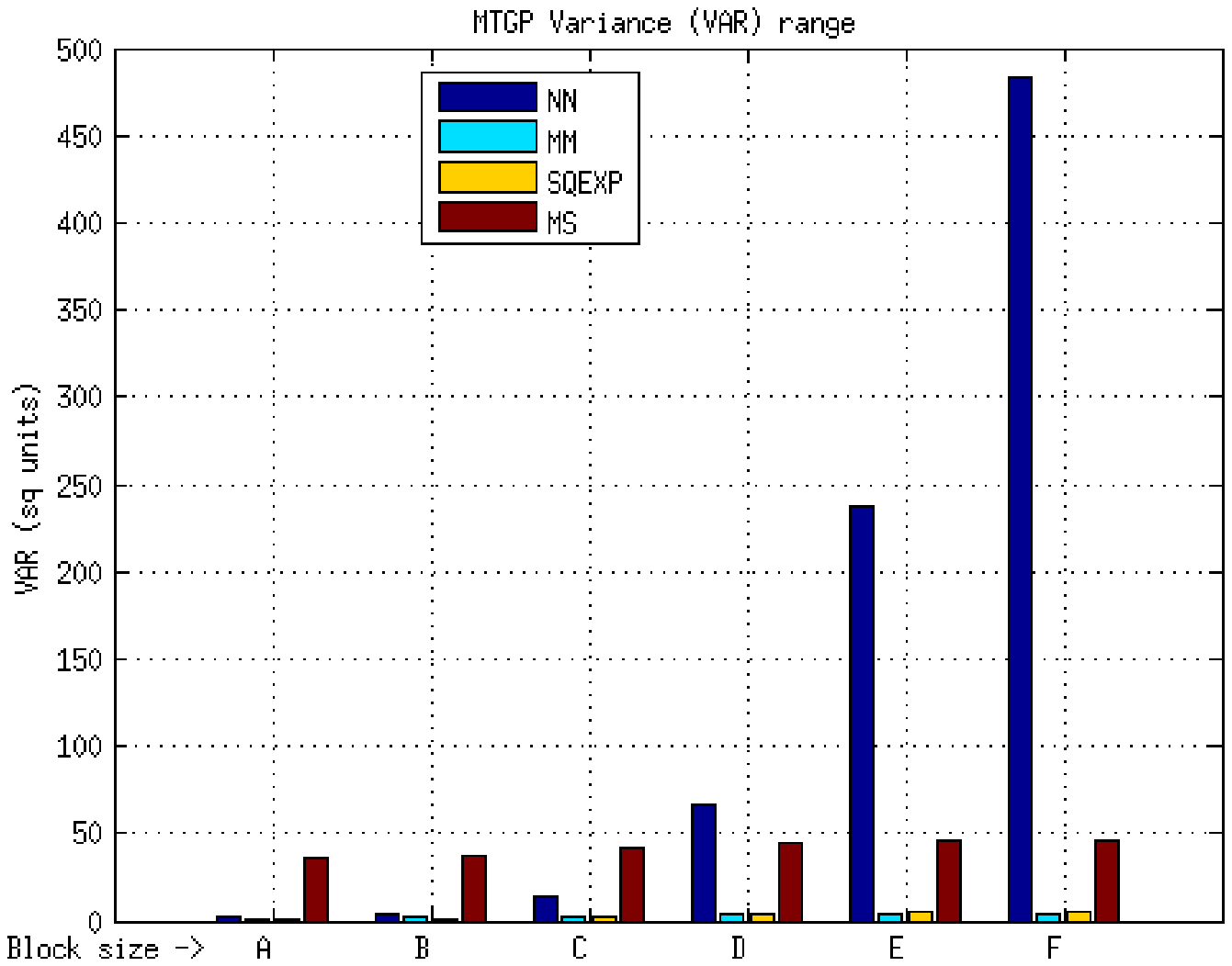}}
  \end{center}
  \caption{Element E1, MTGP approach, VAR metric. The figure above shows the average values; the one below shows the range of values obtained considering two standard deviations about the mean. Test block sizes (m) - A (22 x 11 x 2), B (44 x 22 x 4), C (84 x 45 x 9), D (174 x 89 x 18), E (348 x 177 x 35) and F (696 x 353 x 70).}
  \label{fig:e1_mtgp_var}
\end{figure}

\begin{figure}[htbp]
  \begin{center}
    \subfigure{\includegraphics[width=0.78\columnwidth]{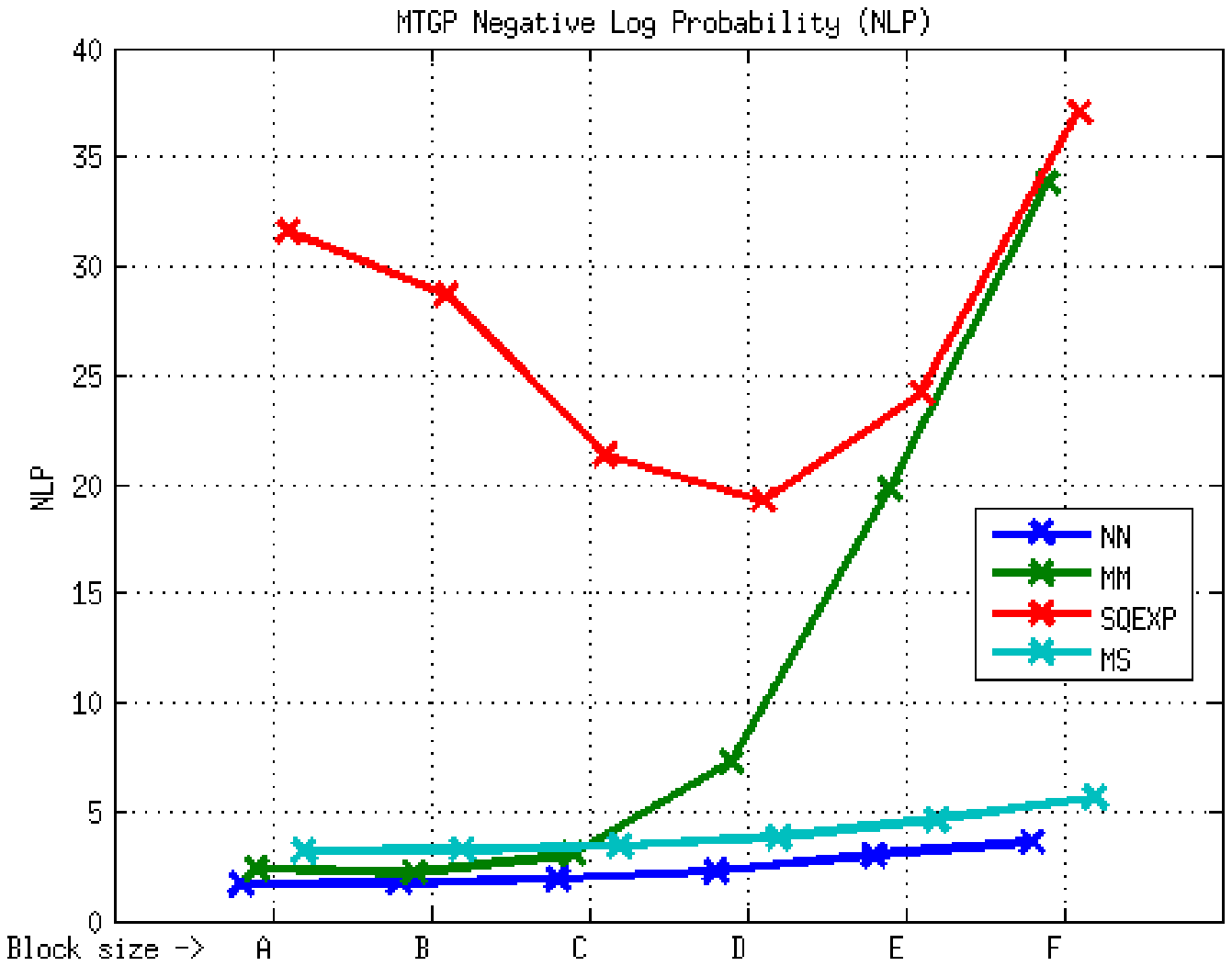}}
    \subfigure{\includegraphics[width=0.78\columnwidth]{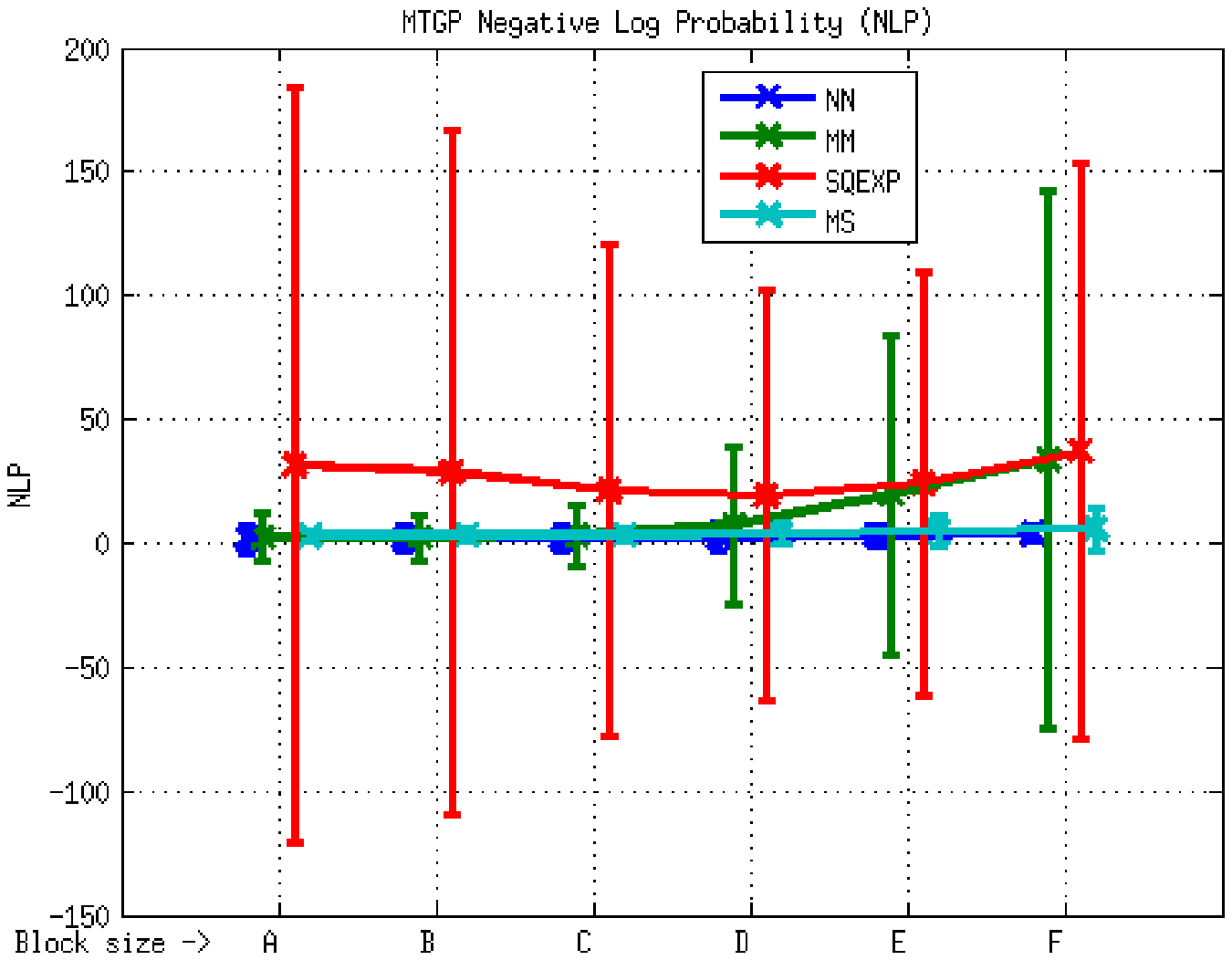}}
  \end{center}
  \caption{Element E1, MTGP approach, NLP metric. The figure above shows the average values; the one below shows the range of values obtained considering two standard deviations about the mean. Test block sizes (m) - A (22 x 11 x 2), B (44 x 22 x 4), C (84 x 45 x 9), D (174 x 89 x 18), E (348 x 177 x 35) and F (696 x 353 x 70).}
  \label{fig:e1_mtgp_nlp}
\end{figure}

\begin{figure}[htbp]
  \begin{center}
    \subfigure{\includegraphics[width=0.78\columnwidth]{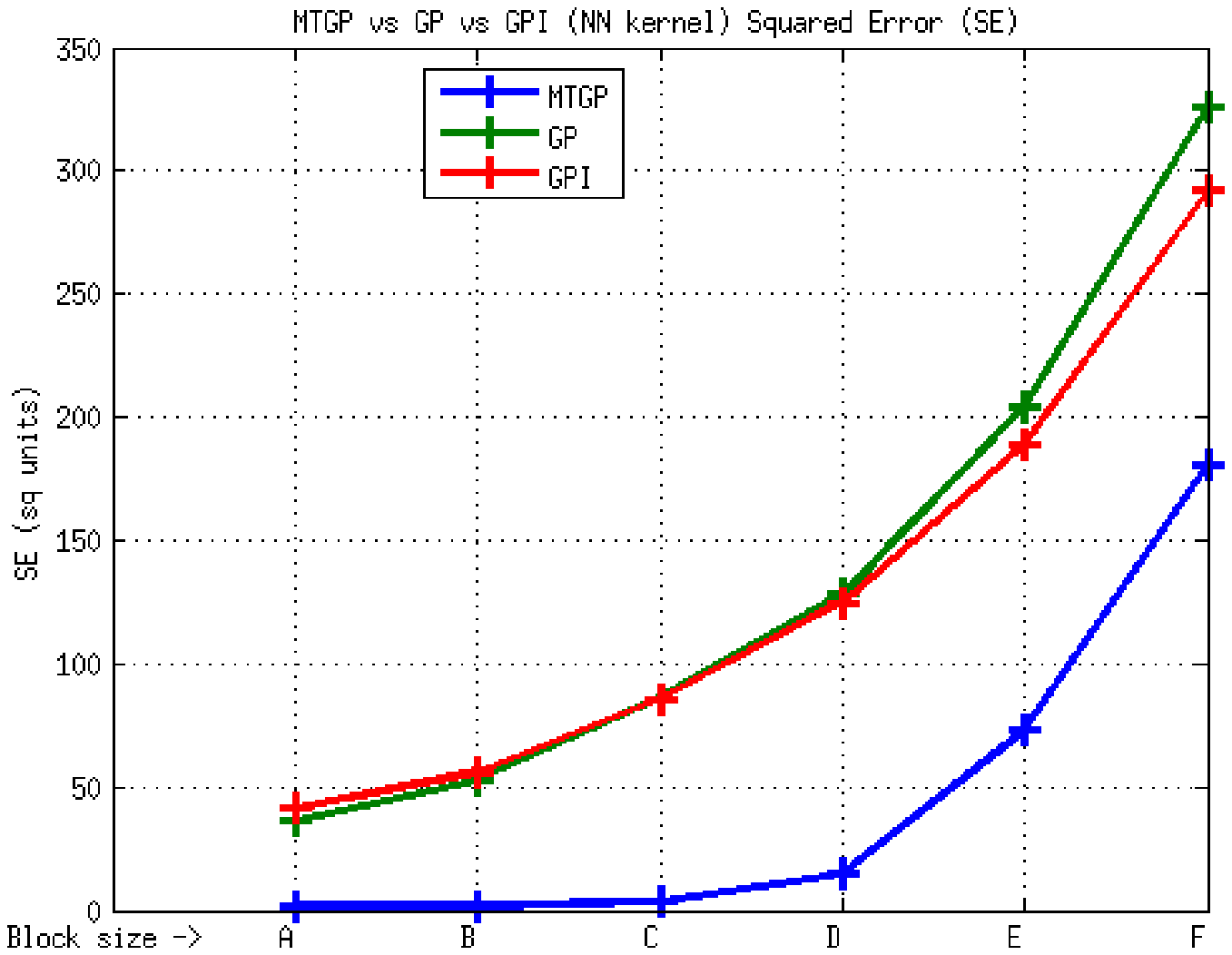}}
    \subfigure{\includegraphics[width=0.78\columnwidth]{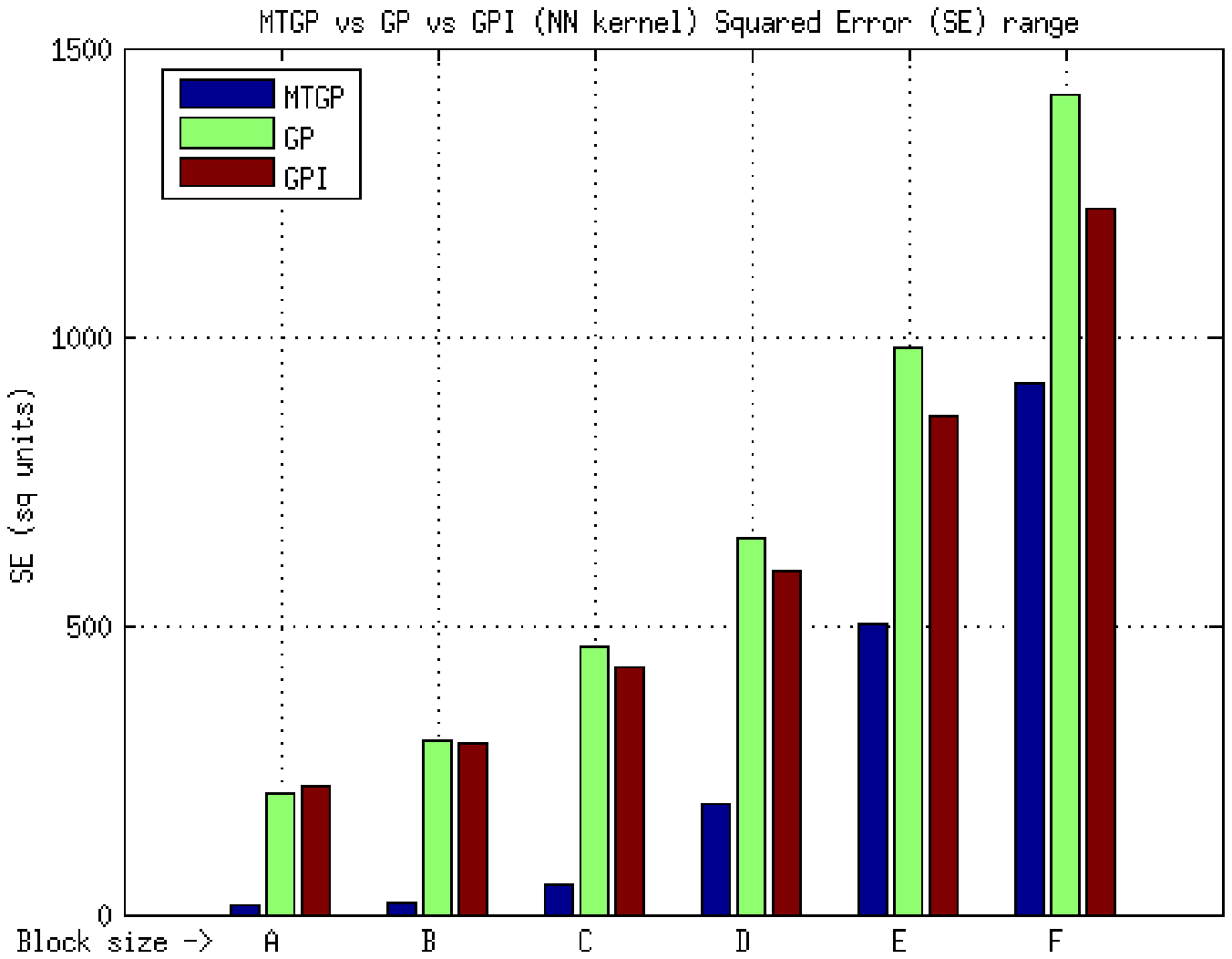}}
  \end{center}
  \caption{Element E1, MTGP vs GP vs GPI approaches, NN kernel, SE metric. The figure above shows the average values; the one below shows the range of values obtained considering two standard deviations about the mean. Test block sizes (m) - A (22 x 11 x 2), B (44 x 22 x 4), C (84 x 45 x 9), D (174 x 89 x 18), E (348 x 177 x 35) and F (696 x 353 x 70).}
  \label{fig:e1_mtgp_gp_gpi_nn_se}
\end{figure}

\begin{figure}[htbp]
  \begin{center}
    \subfigure{\includegraphics[width=0.78\columnwidth]{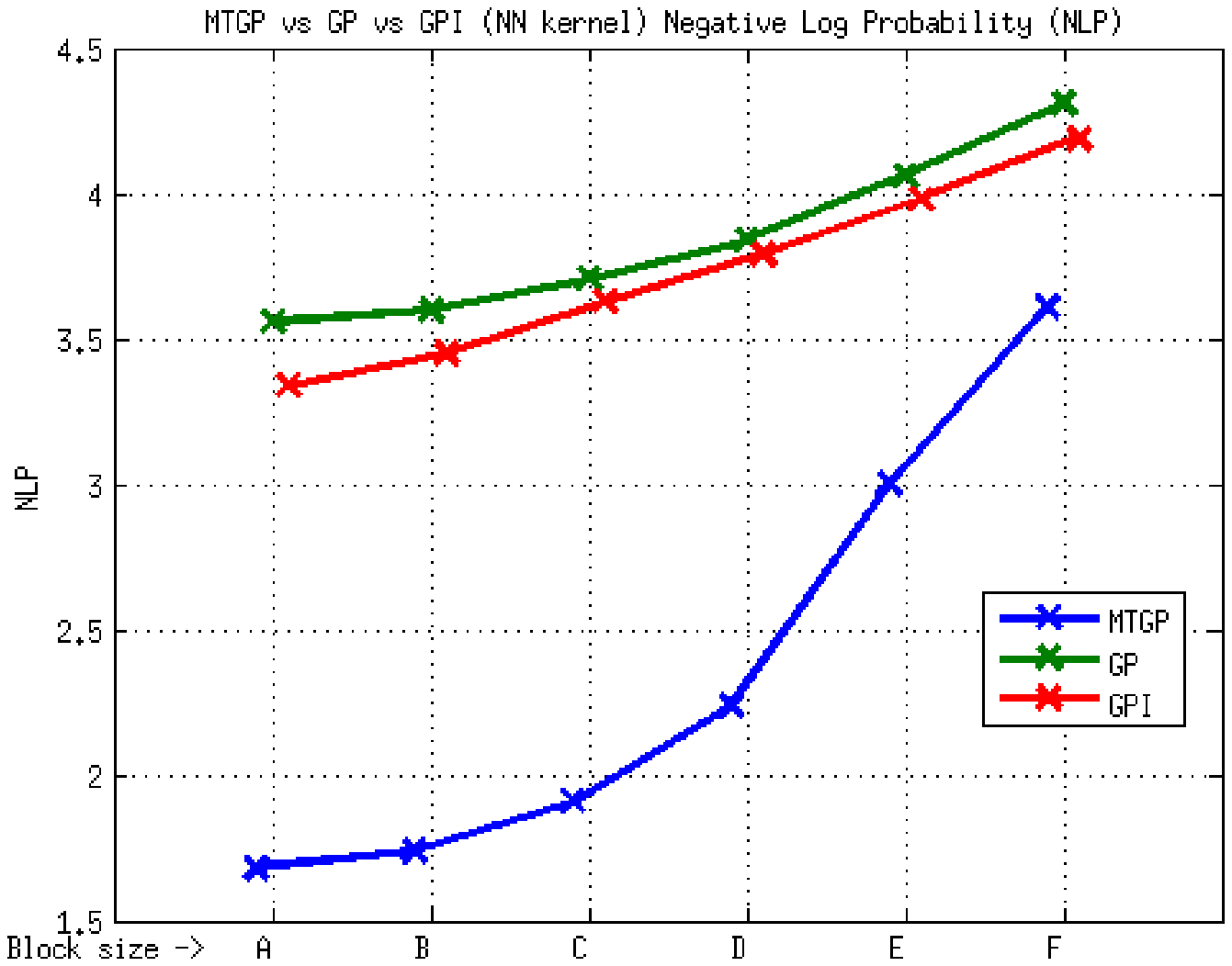}}
    \subfigure{\includegraphics[width=0.78\columnwidth]{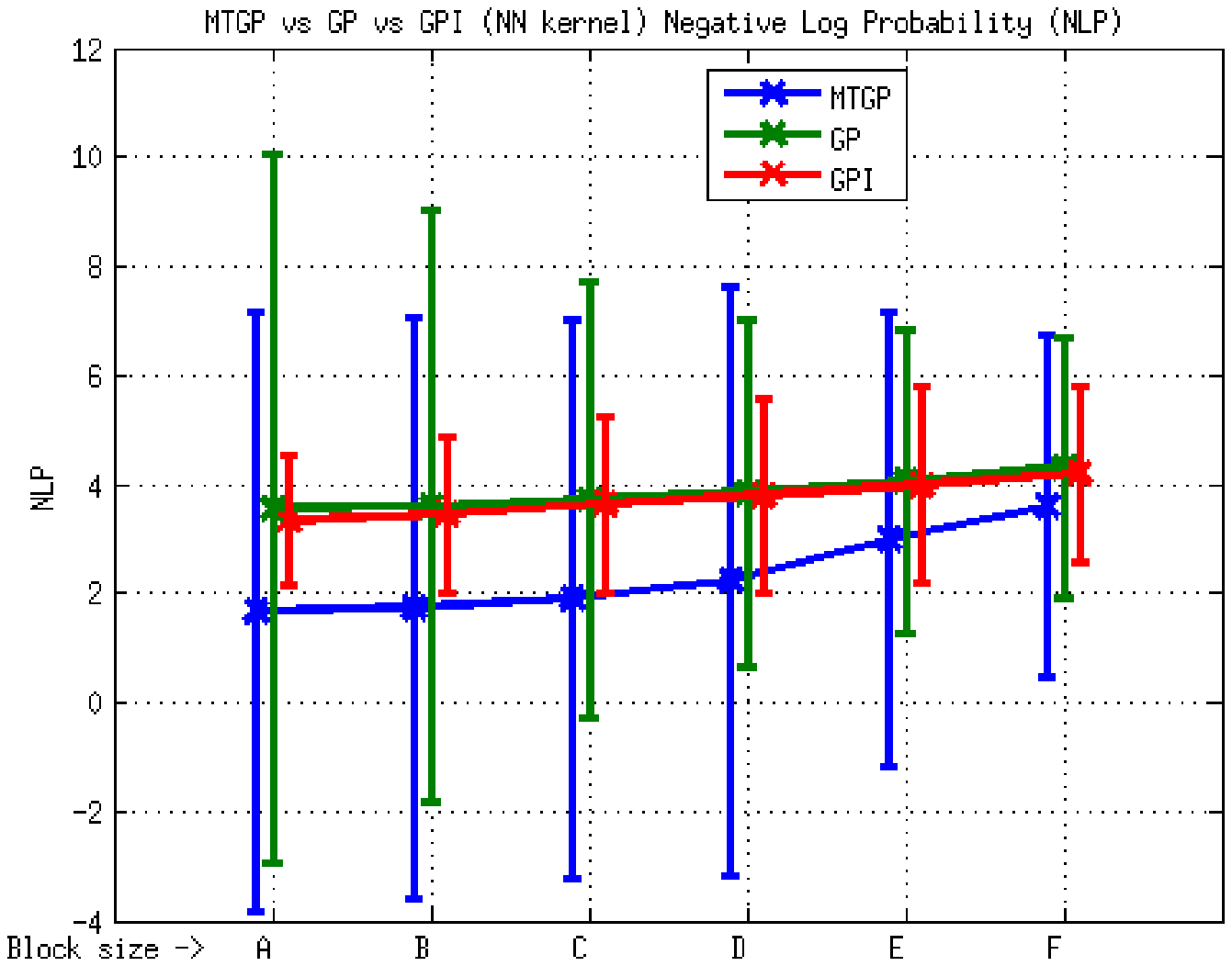}}
  \end{center}
  \caption{Element E1, MTGP vs GP vs GPI approaches, NN kernel, NLP metric. The figure above shows the average values; the one below shows the range of values obtained considering two standard deviations about the mean. Test block sizes (m) - A (22 x 11 x 2), B (44 x 22 x 4), C (84 x 45 x 9), D (174 x 89 x 18), E (348 x 177 x 35) and F (696 x 353 x 70).}
  \label{fig:e1_mtgp_gp_gpi_nn_nlp}
\end{figure}

\begin{figure}[htbp]
  \begin{center}
    \subfigure{\includegraphics[width=0.78\columnwidth]{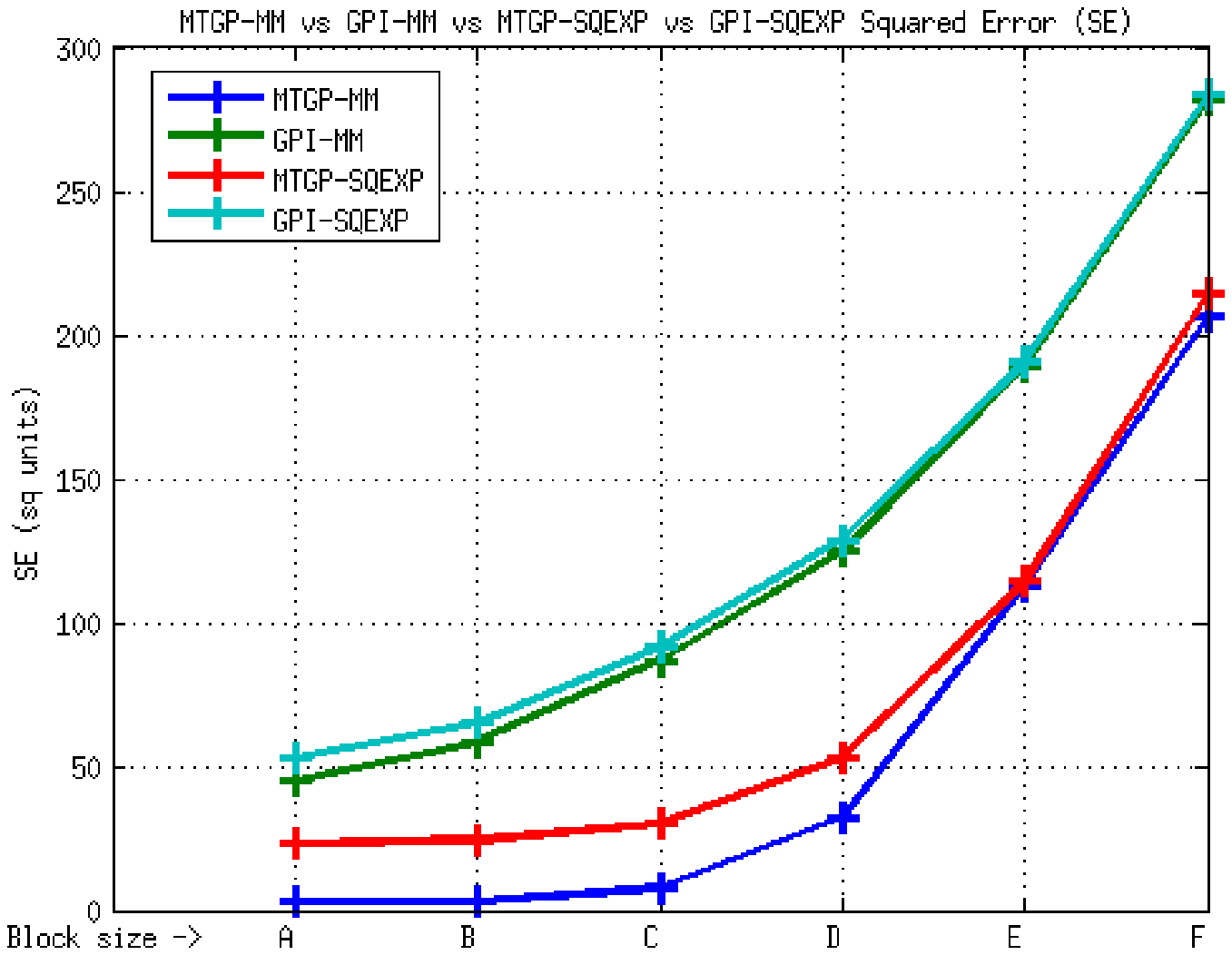}}
    \subfigure{\includegraphics[width=0.78\columnwidth]{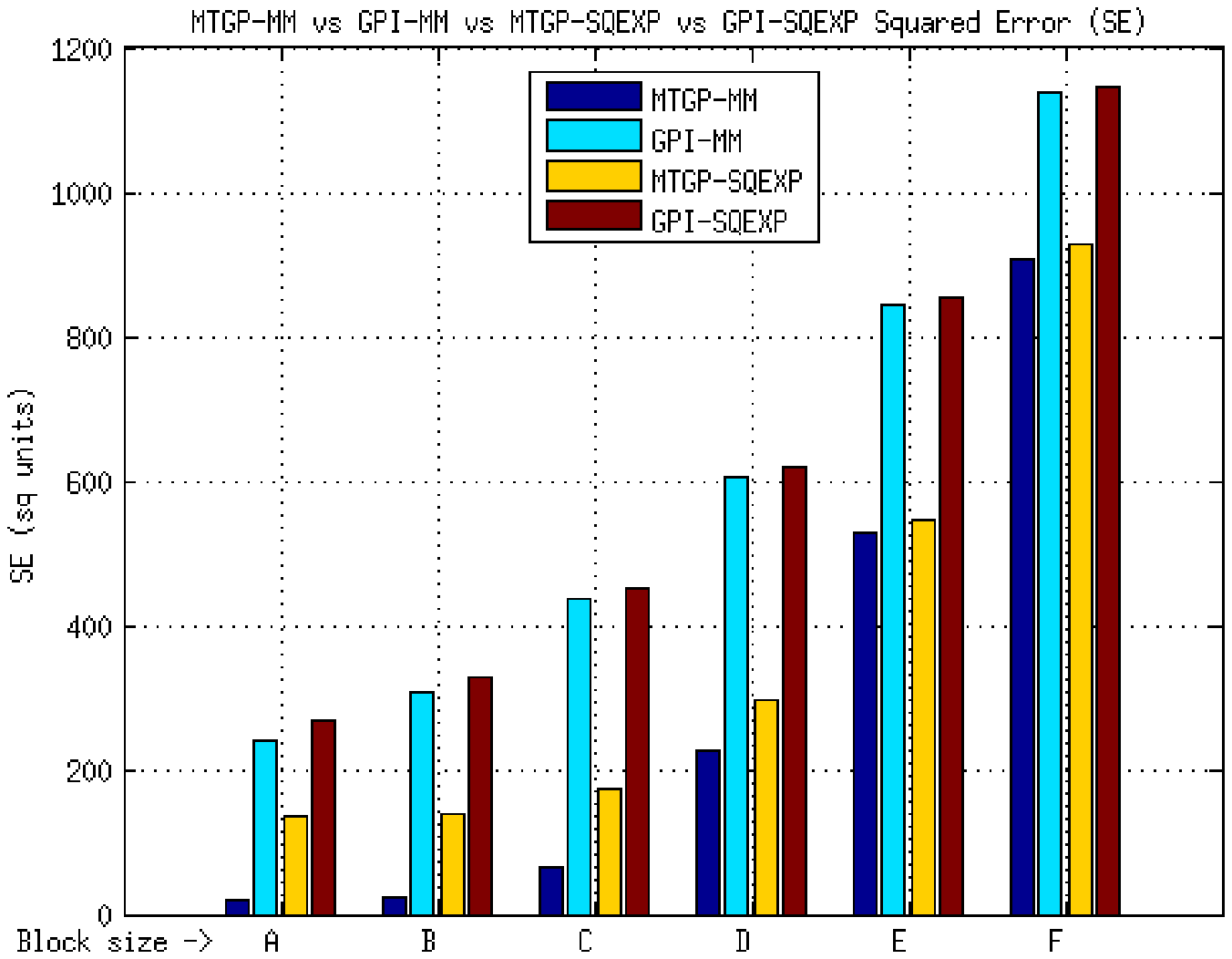}}
  \end{center}
  \caption{Element E1, MTGP vs GPI approaches, MM and SQEXP kernels, SE metric. The figure above shows the average values; the one below shows the range of values obtained considering two standard deviations about the mean. Test block sizes (m) - A (22 x 11 x 2), B (44 x 22 x 4), C (84 x 45 x 9), D (174 x 89 x 18), E (348 x 177 x 35) and F (696 x 353 x 70).}
  \label{fig:e1_mtgp_gpi_mm_sqexp_se}
\end{figure}

\begin{figure}[htbp]
  \begin{center}
    \subfigure{\includegraphics[width=0.78\columnwidth]{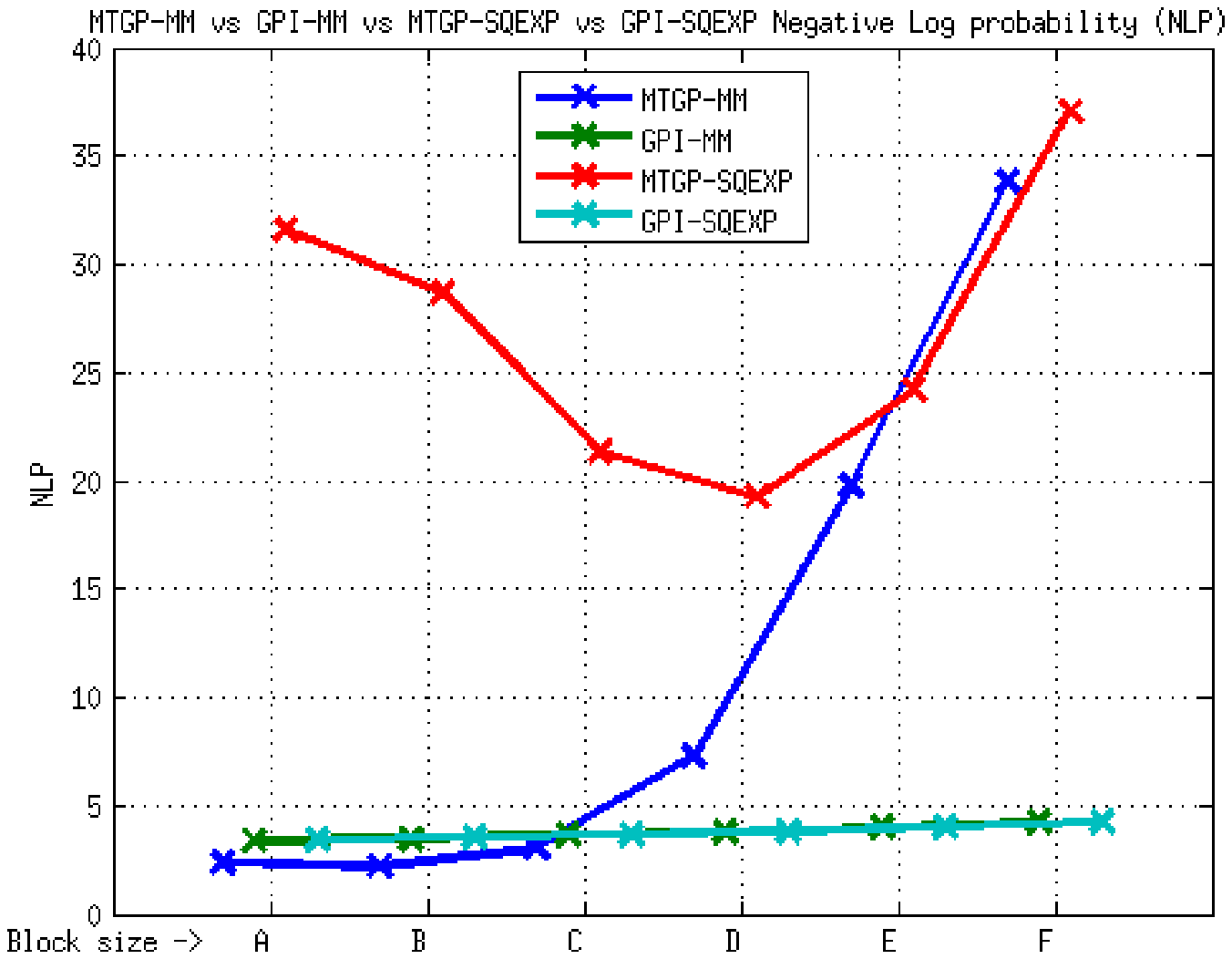}}
    \subfigure{\includegraphics[width=0.78\columnwidth]{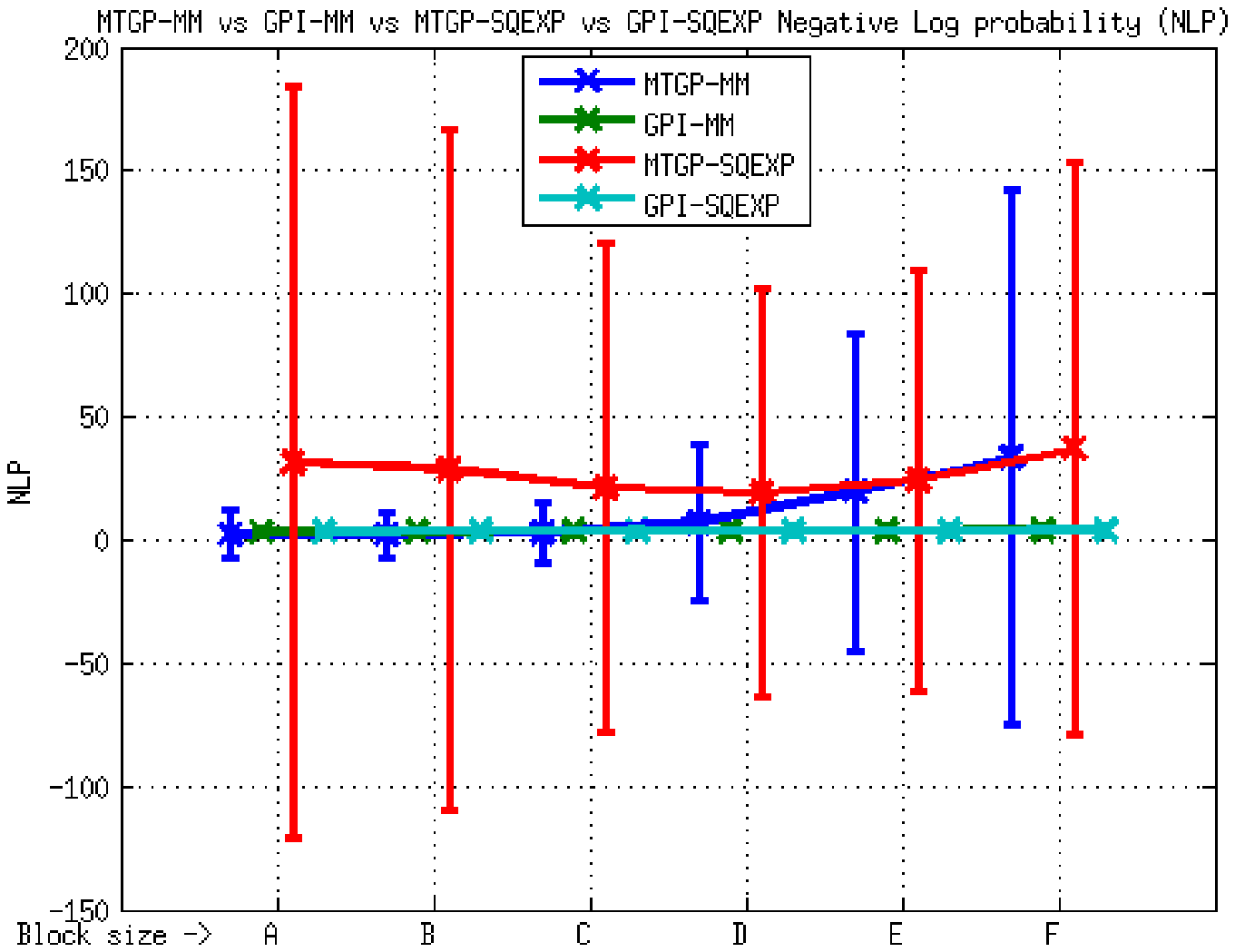}}
  \end{center}
  \caption{Element E1, MTGP vs GPI approaches, MM and SQEXP kernels, NLP metric. The figure above shows the average values; the one below shows the range of values obtained considering two standard deviations about the mean. Test block sizes (m) - A (22 x 11 x 2), B (44 x 22 x 4), C (84 x 45 x 9), D (174 x 89 x 18), E (348 x 177 x 35) and F (696 x 353 x 70).}
  \label{fig:e1_mtgp_gpi_mm_sqexp_nlp}
\end{figure}

\begin{figure}[htbp]
  \begin{center}
    \subfigure{\includegraphics[width=0.78\columnwidth]{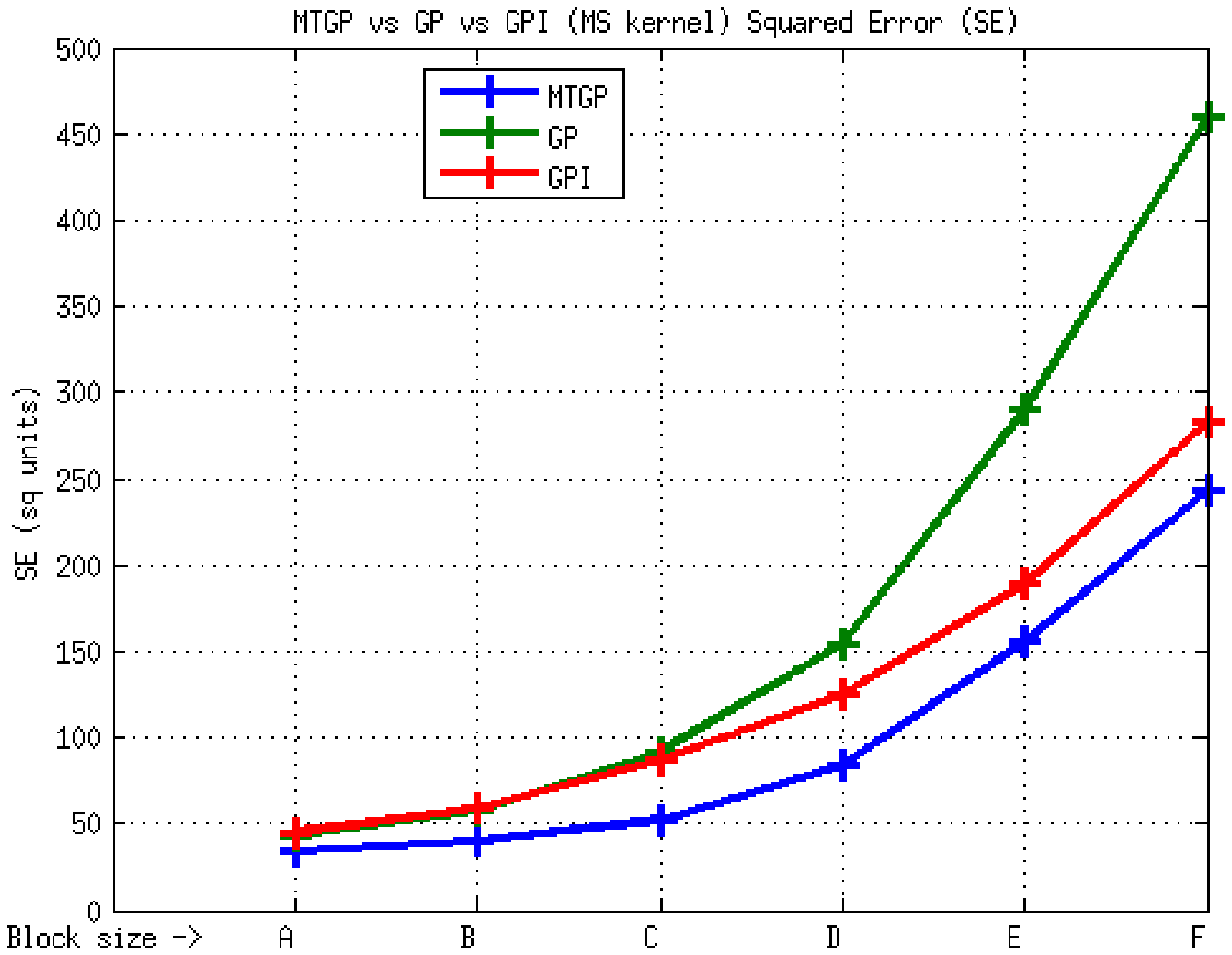}}
    \subfigure{\includegraphics[width=0.78\columnwidth]{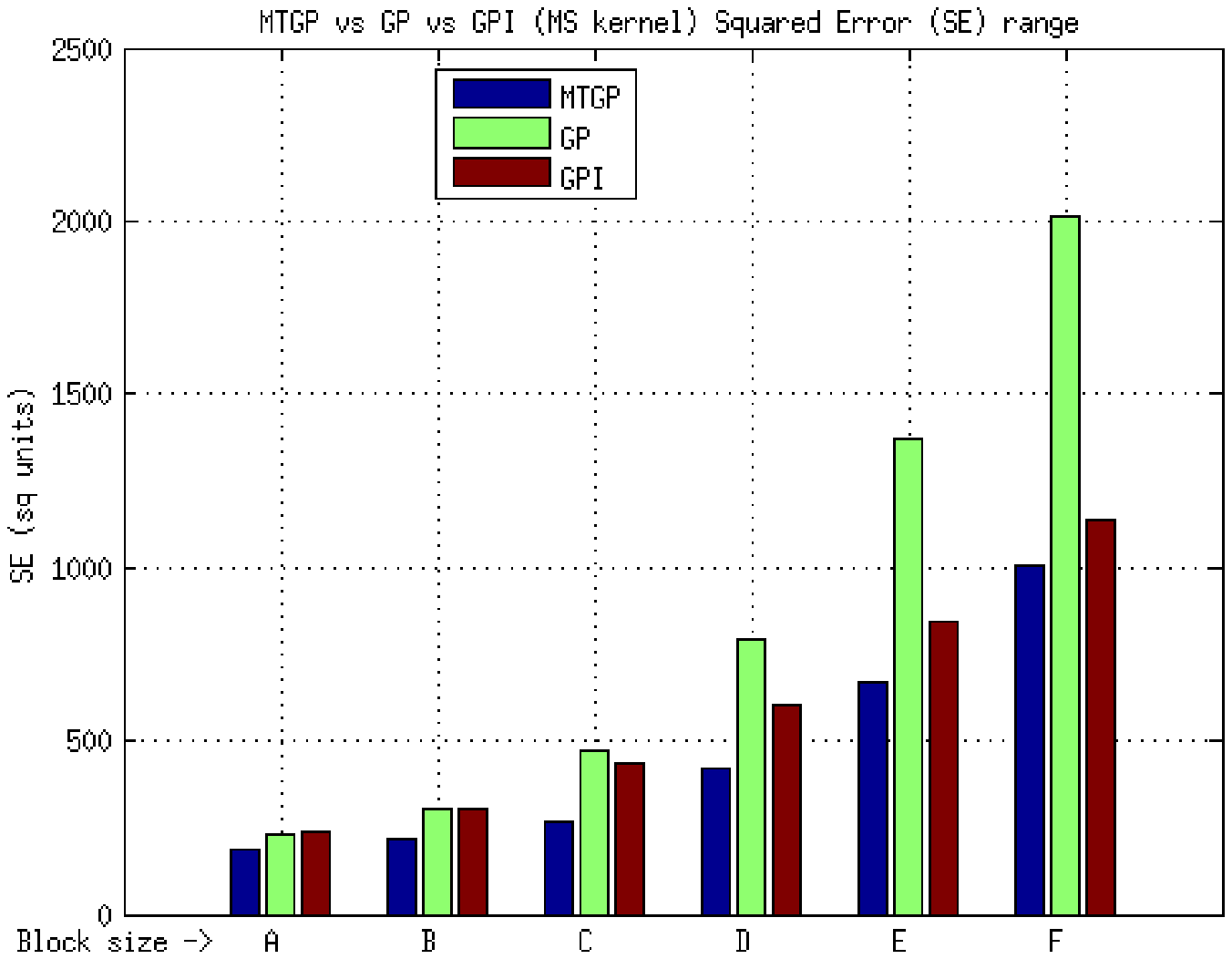}}
  \end{center}
  \caption{Element E1, MTGP vs GP vs GPI approaches, MS kernel, SE metric. The figure above shows the average values; the one below shows the range of values obtained considering two standard deviations about the mean. Test block sizes (m) - A (22 x 11 x 2), B (44 x 22 x 4), C (84 x 45 x 9), D (174 x 89 x 18), E (348 x 177 x 35) and F (696 x 353 x 70).}
  \label{fig:e1_mtgp_gp_gpi_ms_se}
\end{figure}

\begin{figure}[htbp]
  \begin{center}
    \subfigure{\includegraphics[width=0.78\columnwidth]{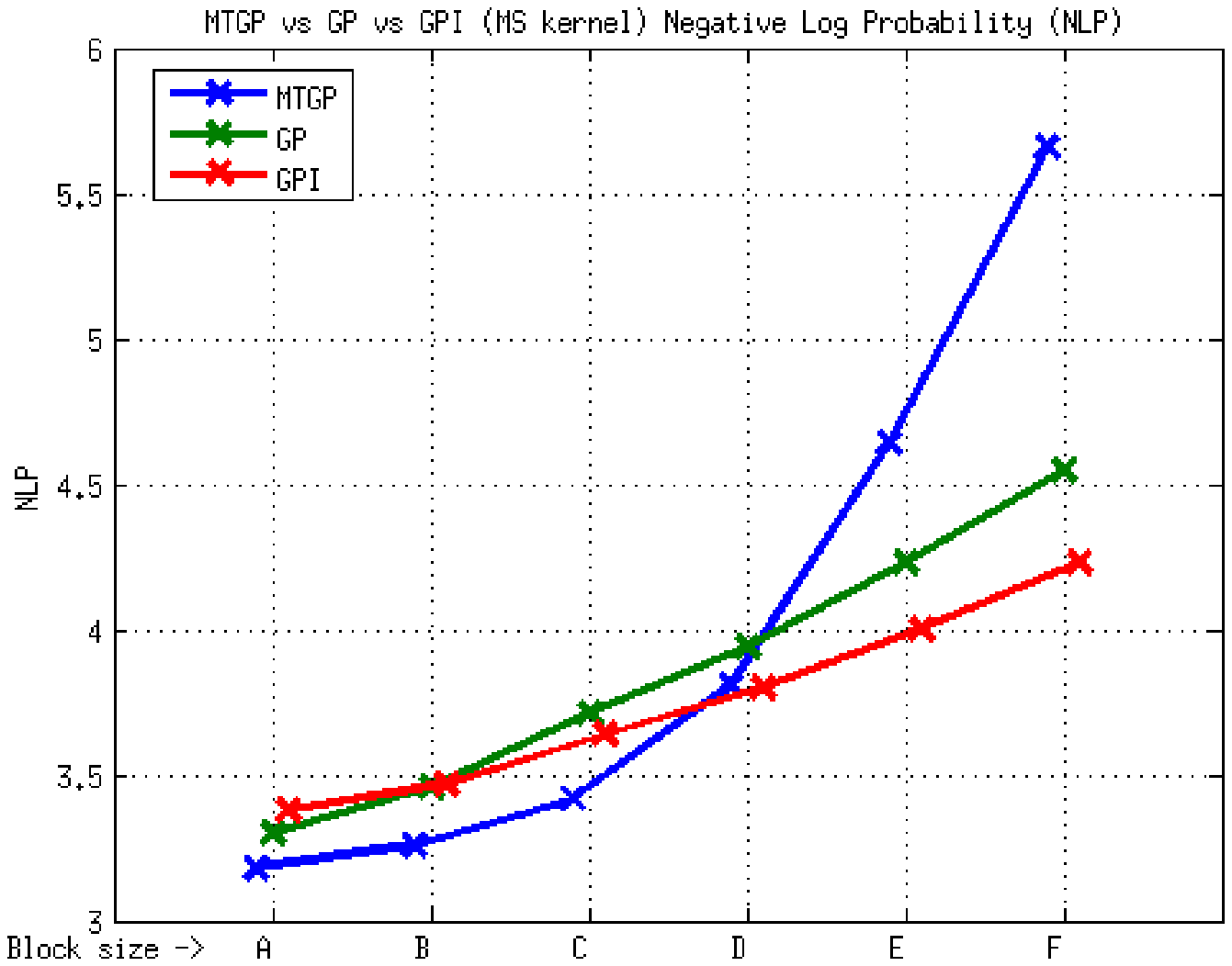}}
    \subfigure{\includegraphics[width=0.78\columnwidth]{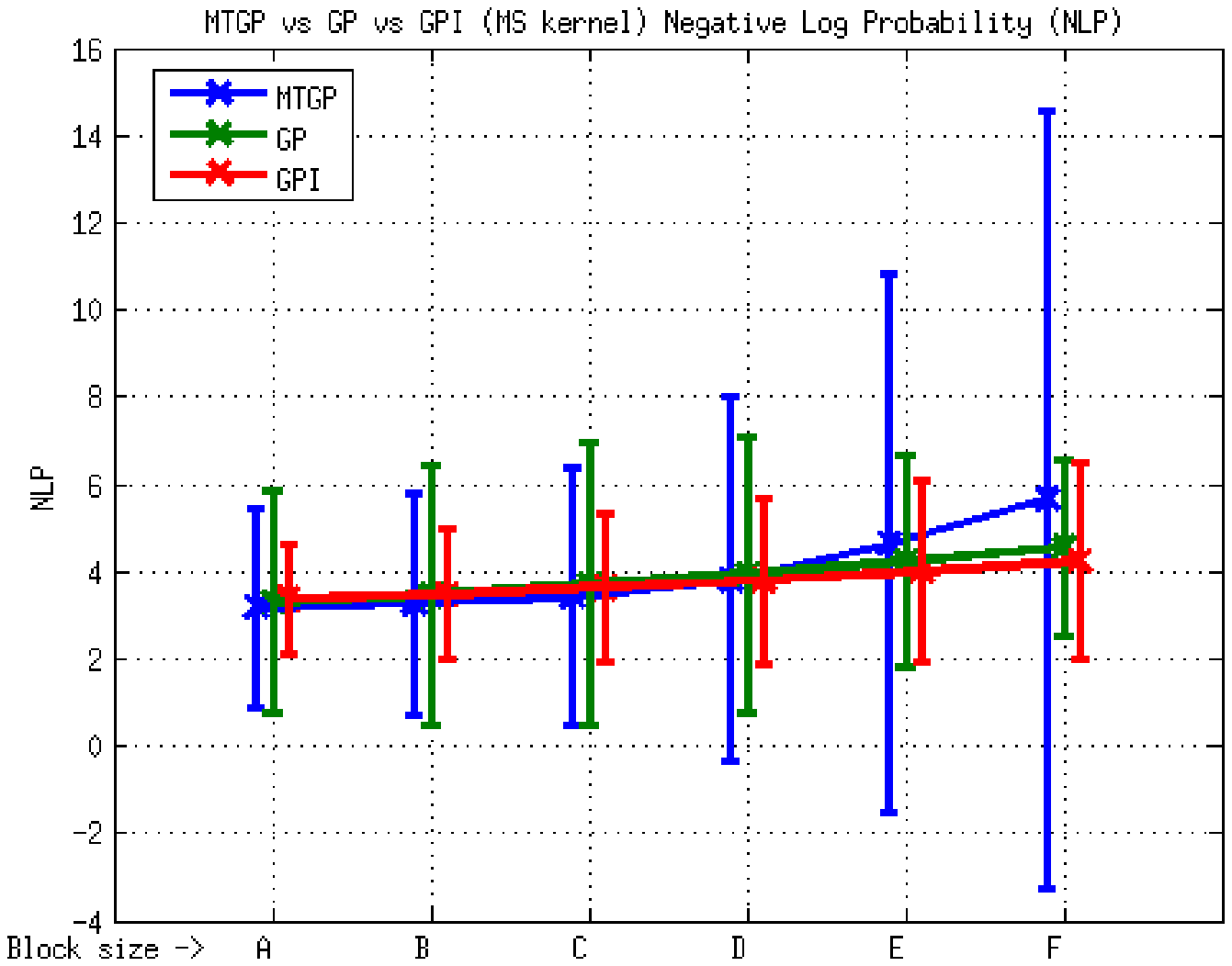}}
  \end{center}
  \caption{Element E1, MTGP vs GP vs GPI approaches, MS kernel, NLP metric. The figure above shows the average values; the one below shows the range of values obtained considering two standard deviations about the mean. Test block sizes (m) - A (22 x 11 x 2), B (44 x 22 x 4), C (84 x 45 x 9), D (174 x 89 x 18), E (348 x 177 x 35) and F (696 x 353 x 70).}
  \label{fig:e1_mtgp_gp_gpi_ms_nlp}
\end{figure}

%% ge2
\clearpage
\begin{figure}[htbp]
  \begin{center}
    \subfigure{\includegraphics[width=0.78\columnwidth]{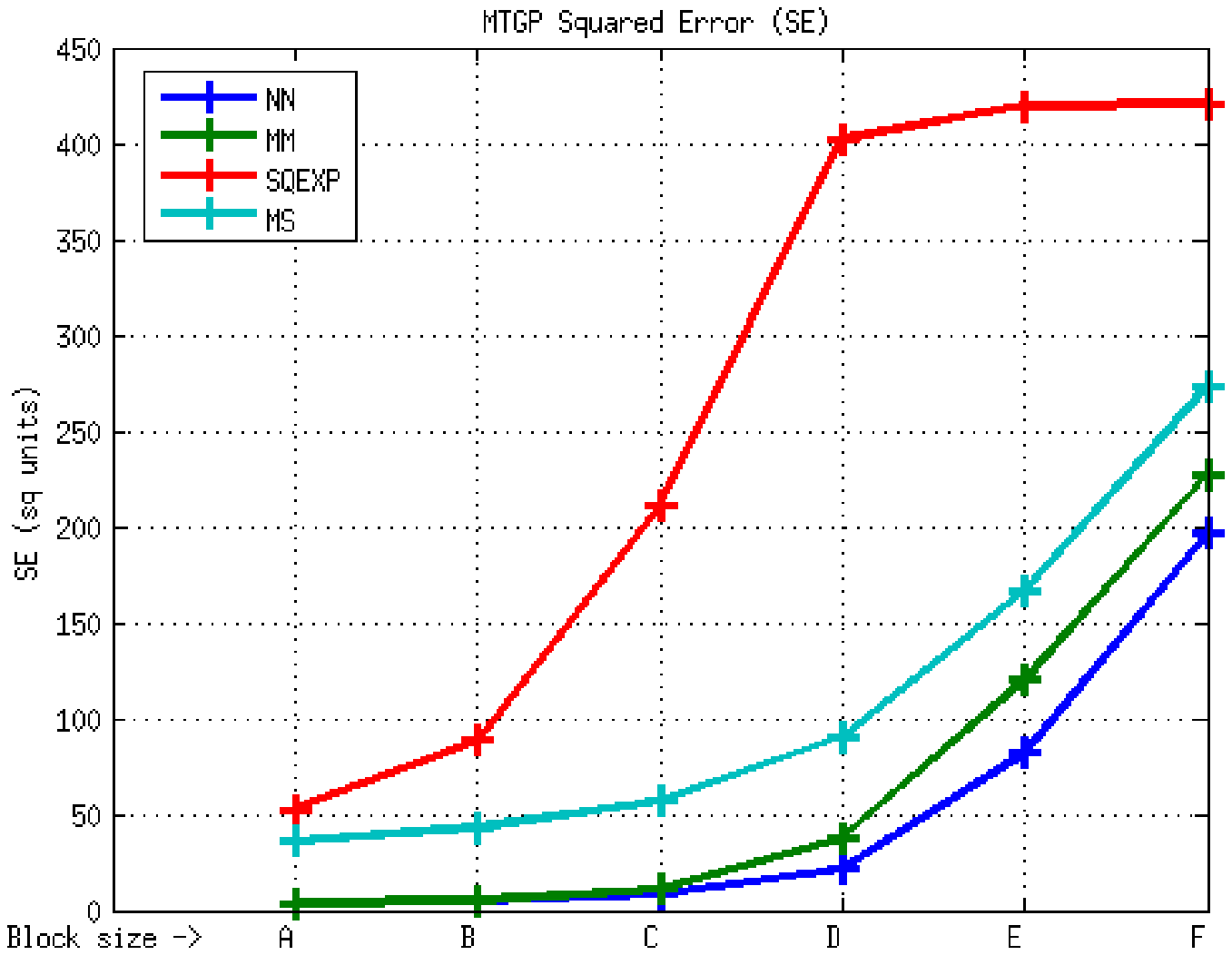}}
    \subfigure{\includegraphics[width=0.78\columnwidth]{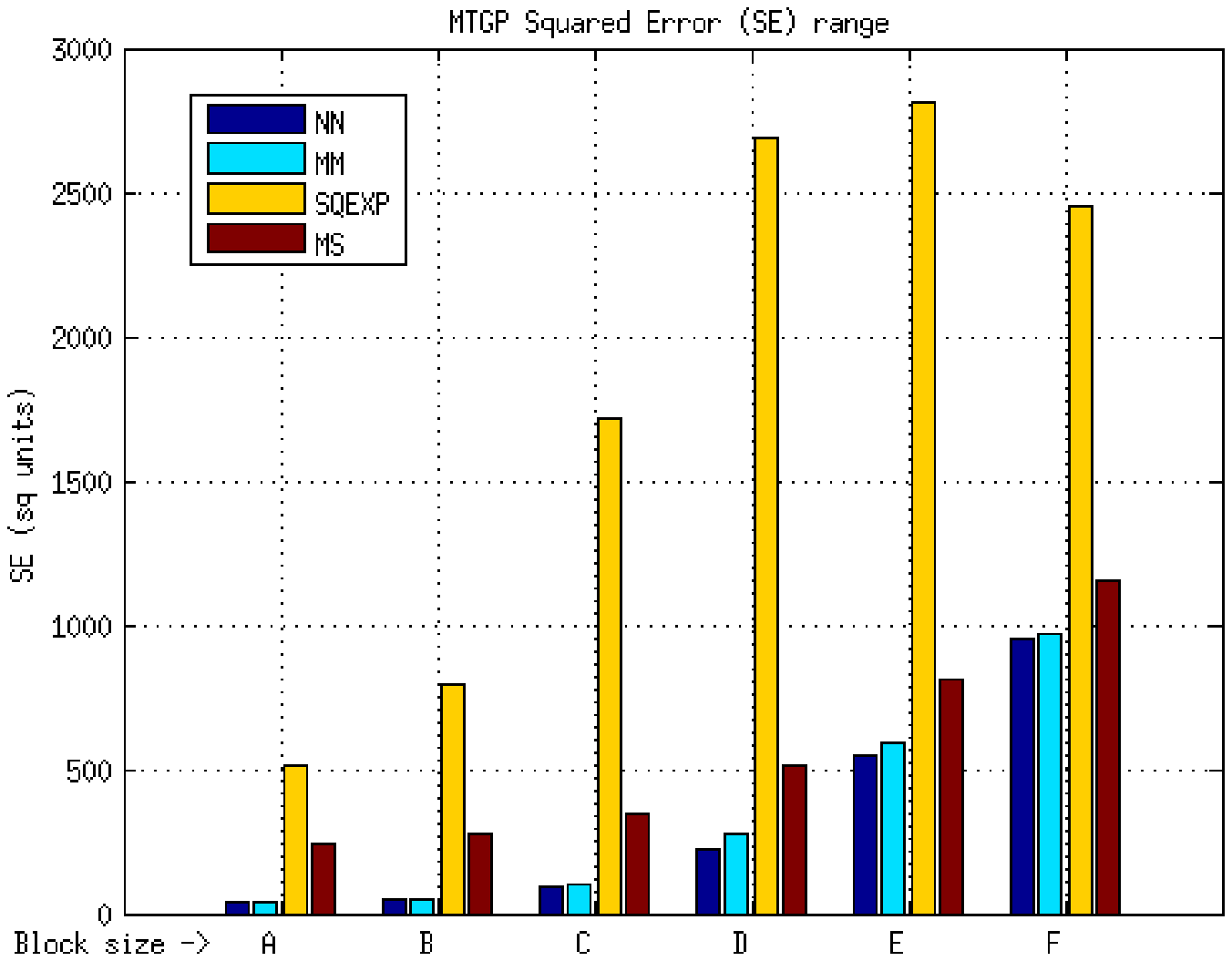}}
  \end{center}
  \caption{Element E2, MTGP approach, SE metric. The figure above shows the average values; the one below shows the range of values obtained considering two standard deviations about the mean. Test block sizes (m) - A (22 x 11 x 2), B (44 x 22 x 4), C (84 x 45 x 9), D (174 x 89 x 18), E (348 x 177 x 35) and F (696 x 353 x 70).}
  \label{fig:e2_mtgp_se}
\end{figure}

\begin{figure}[htbp]
  \begin{center}
    \subfigure{\includegraphics[width=0.78\columnwidth]{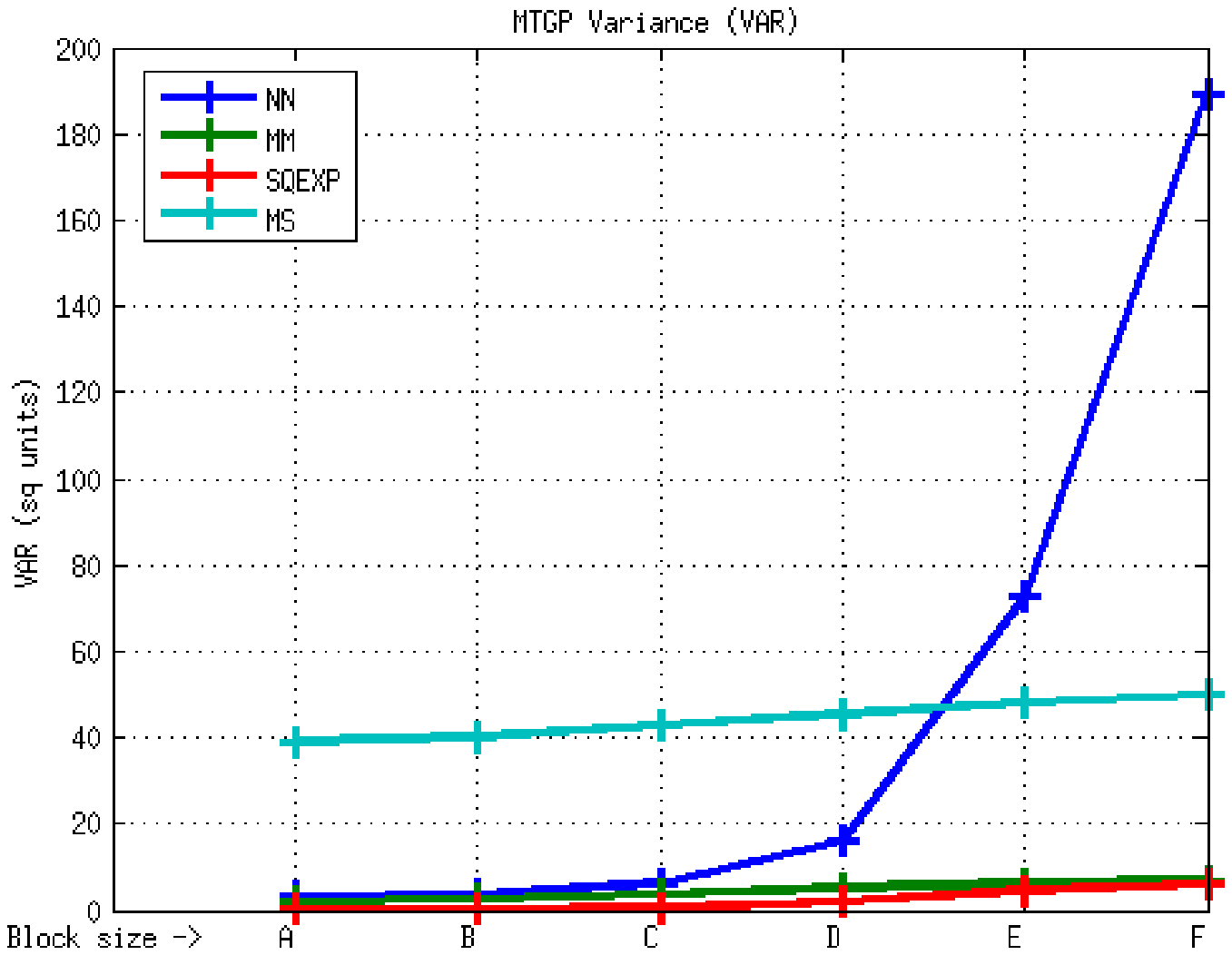}}
    \subfigure{\includegraphics[width=0.78\columnwidth]{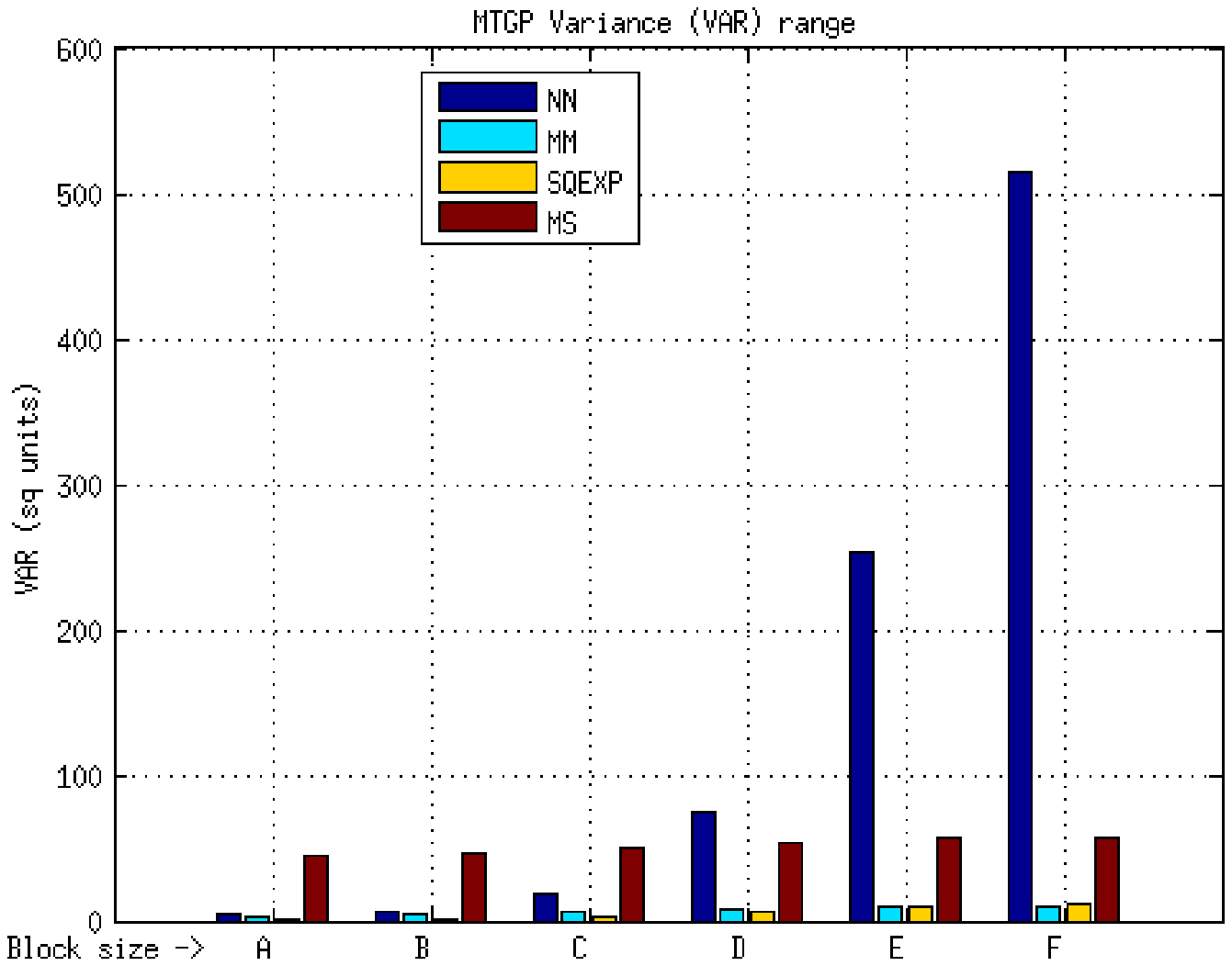}}
  \end{center}
  \caption{Element E2, MTGP approach, VAR metric. The figure above shows the average values; the one below shows the range of values obtained considering two standard deviations about the mean. Test block sizes (m) - A (22 x 11 x 2), B (44 x 22 x 4), C (84 x 45 x 9), D (174 x 89 x 18), E (348 x 177 x 35) and F (696 x 353 x 70).}
  \label{fig:e2_mtgp_var}
\end{figure}

\begin{figure}[htbp]
  \begin{center}
    \subfigure{\includegraphics[width=0.49\columnwidth]{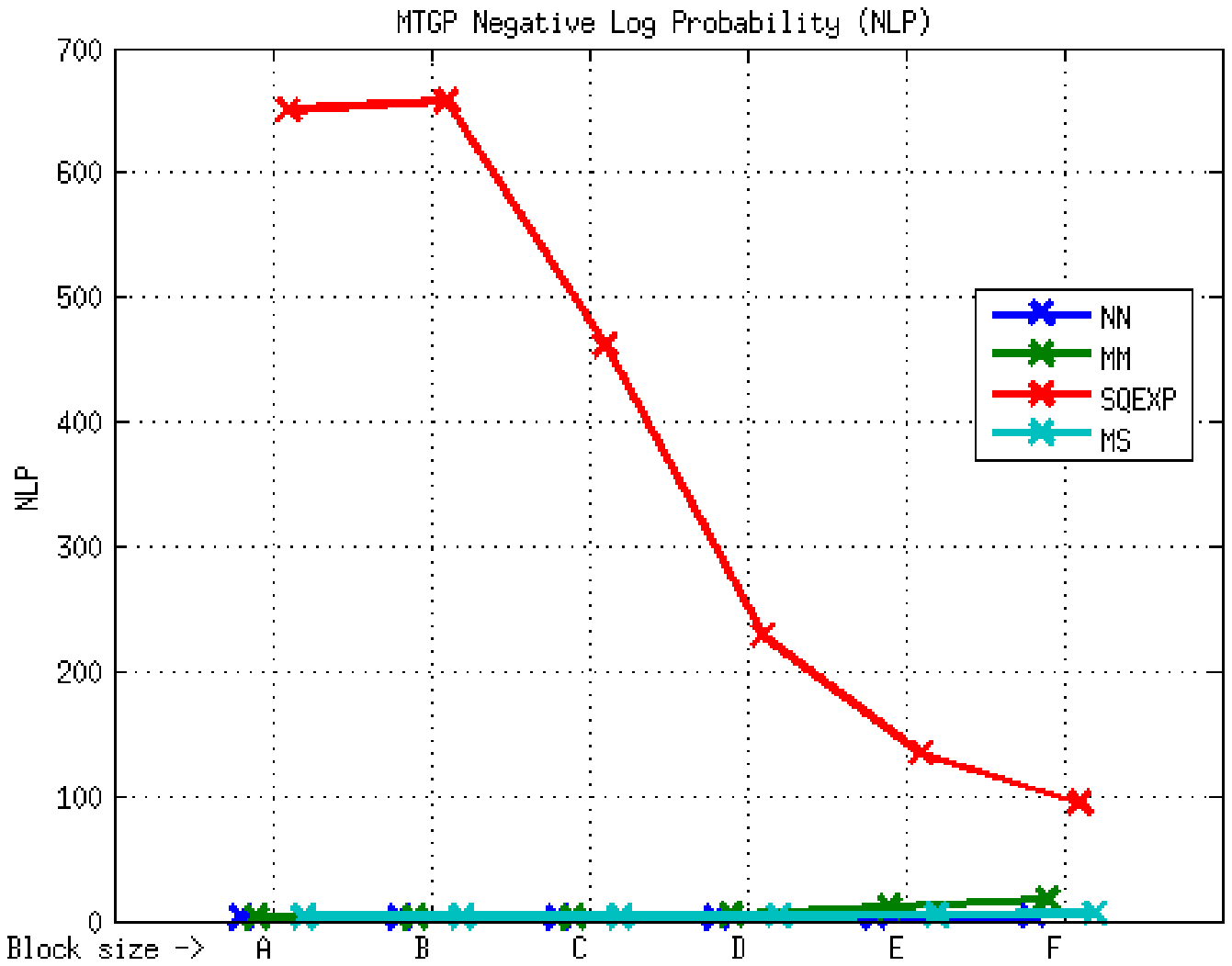}\includegraphics[width=0.49\columnwidth]{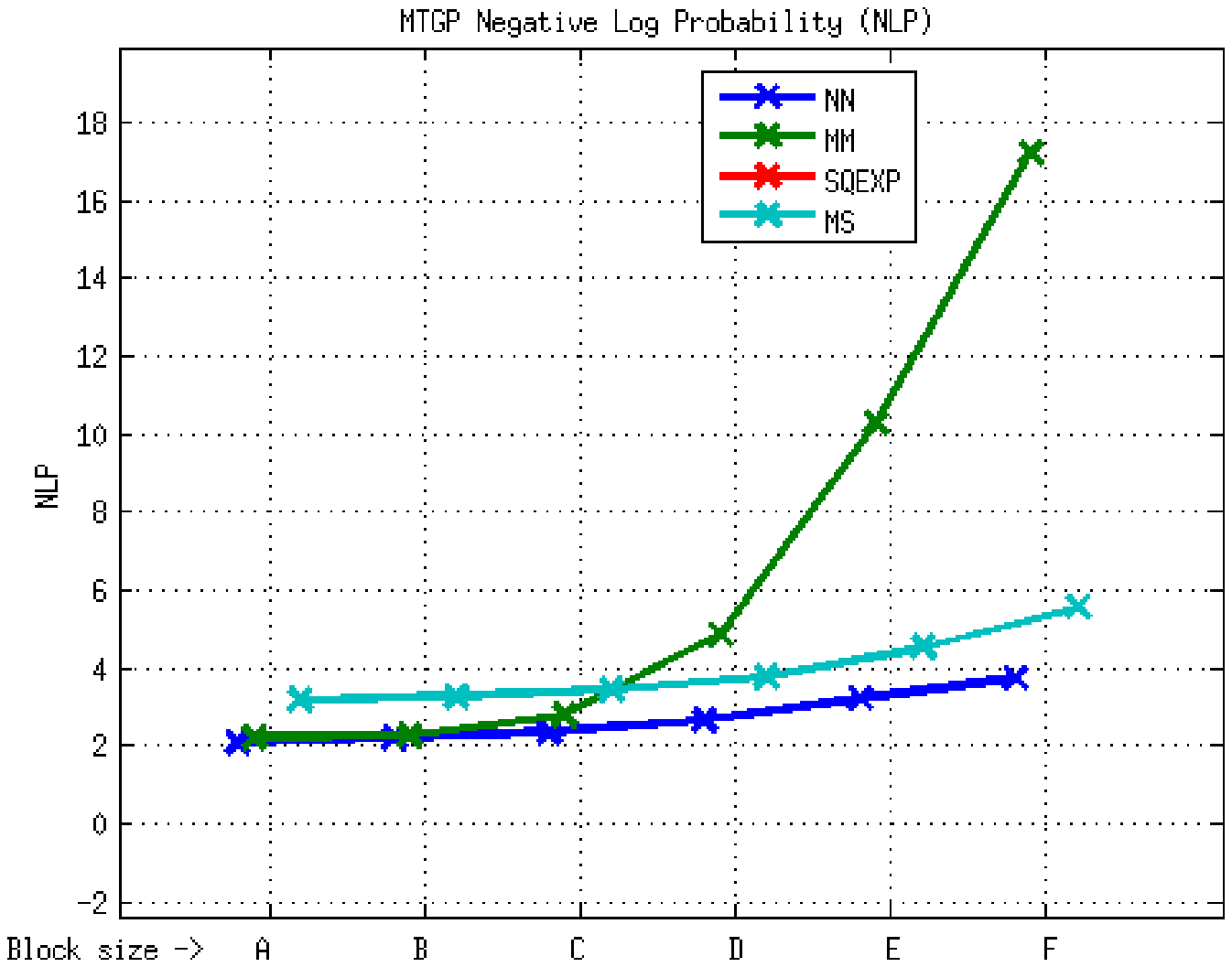}}\vspace{5mm}
    \subfigure{\includegraphics[width=0.8\columnwidth]{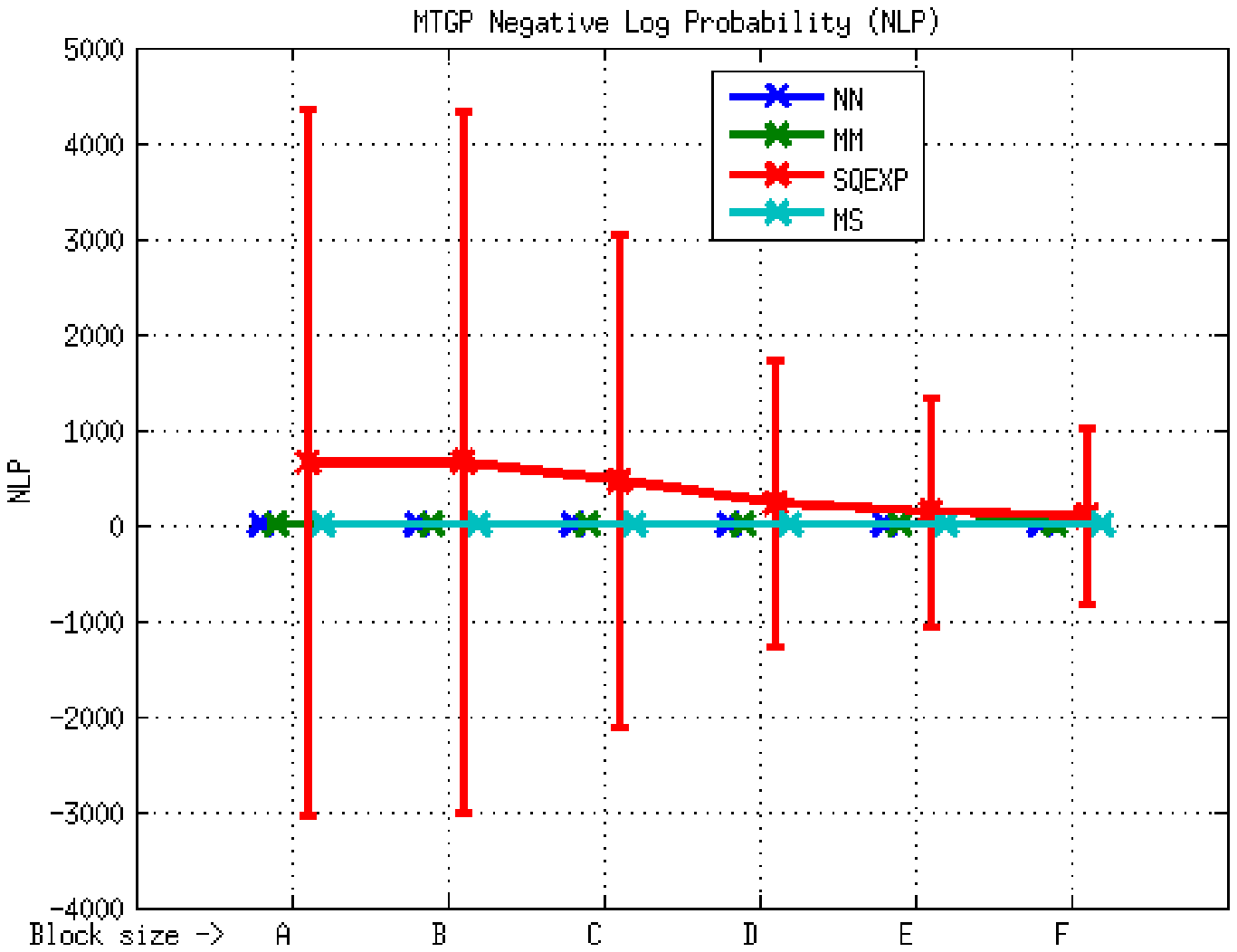}}
  \end{center}
  \caption{Element E2, MTGP approach, NLP metric. The figure above shows the average values, the right image being a zoomed in view of the left; the one below shows the range of values obtained considering two standard deviations about the mean. Test block sizes (m) - A (22 x 11 x 2), B (44 x 22 x 4), C (84 x 45 x 9), D (174 x 89 x 18), E (348 x 177 x 35) and F (696 x 353 x 70).}
  \label{fig:e2_mtgp_nlp}
\end{figure}

\begin{figure}[htbp]
  \begin{center}
    \subfigure{\includegraphics[width=0.78\columnwidth]{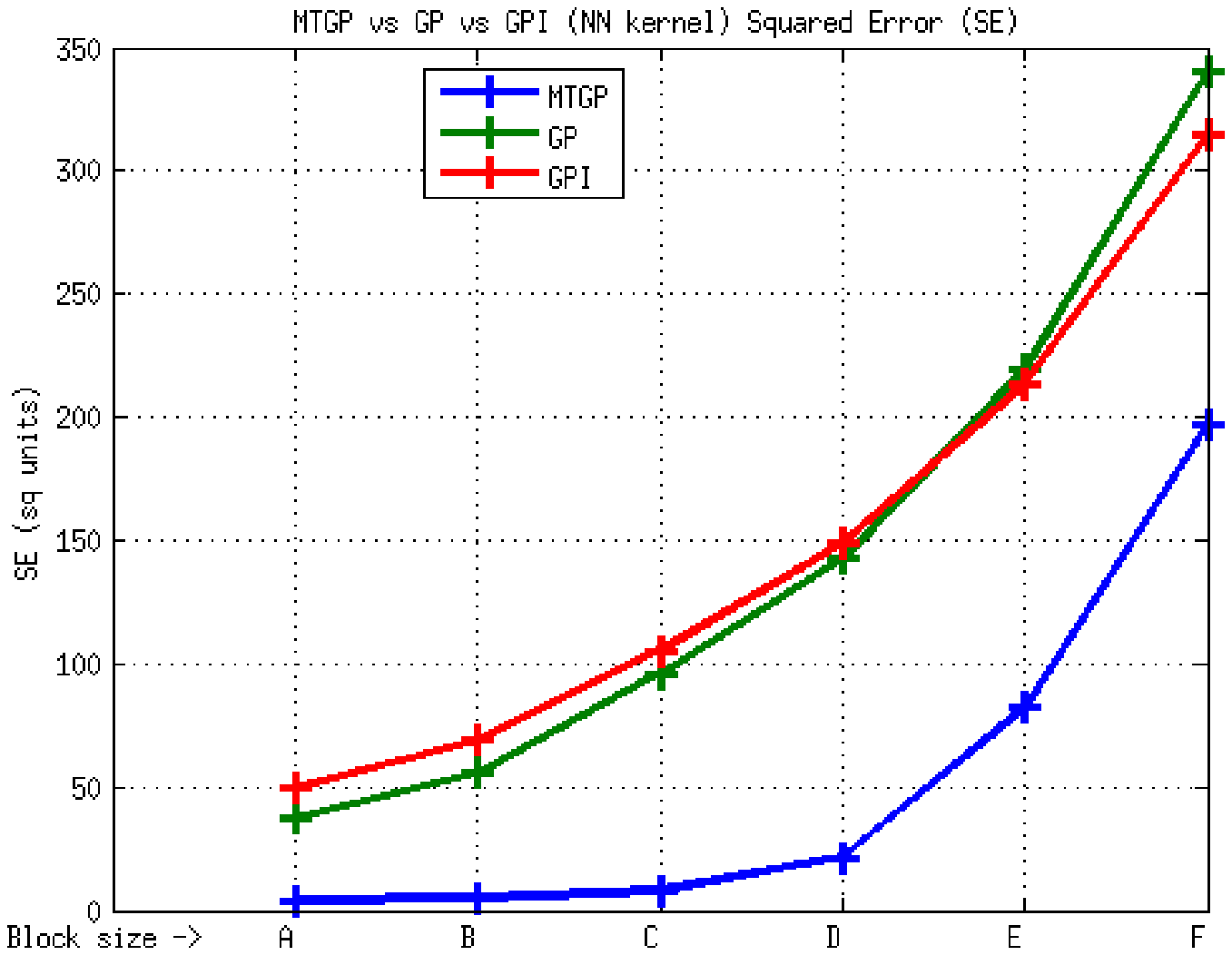}}
    \subfigure{\includegraphics[width=0.78\columnwidth]{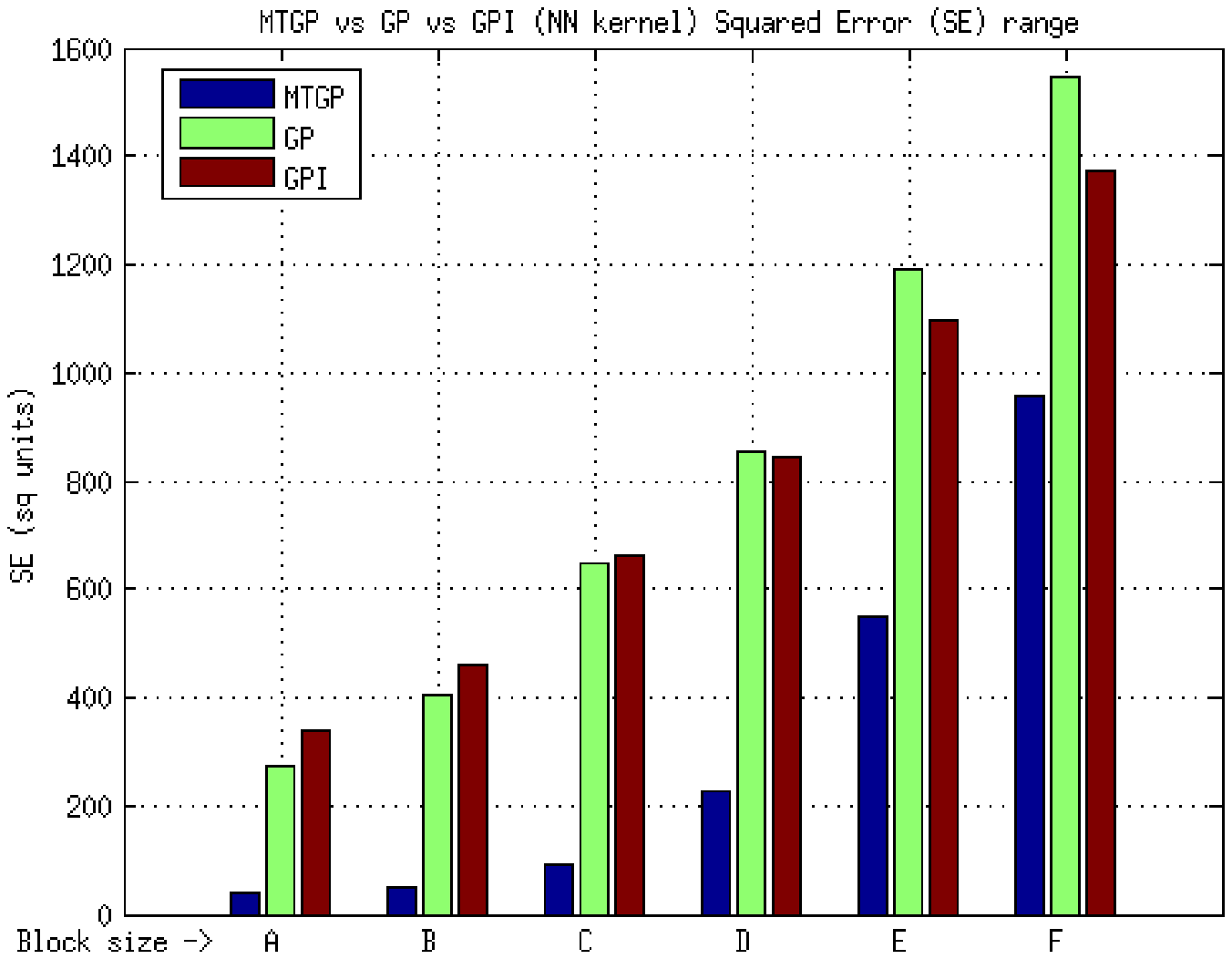}}
  \end{center}
  \caption{Element E2, MTGP vs GP vs GPI approaches, NN kernel, SE metric. The figure above shows the average values; the one below shows the range of values obtained considering two standard deviations about the mean. Test block sizes (m) - A (22 x 11 x 2), B (44 x 22 x 4), C (84 x 45 x 9), D (174 x 89 x 18), E (348 x 177 x 35) and F (696 x 353 x 70).}
  \label{fig:e2_mtgp_gp_gpi_nn_se}
\end{figure}

\begin{figure}[htbp]
  \begin{center}
    \subfigure{\includegraphics[width=0.78\columnwidth]{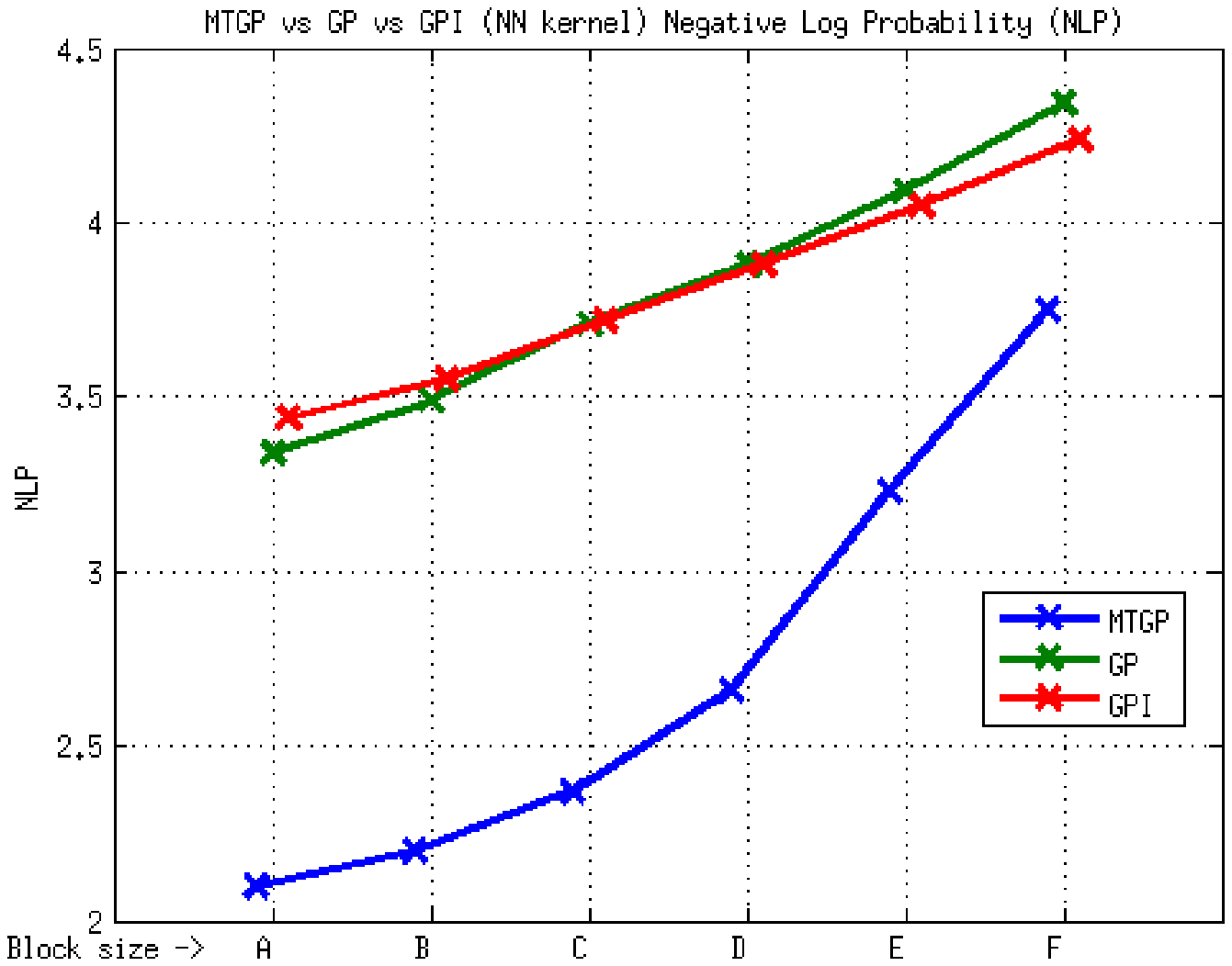}}
    \subfigure{\includegraphics[width=0.78\columnwidth]{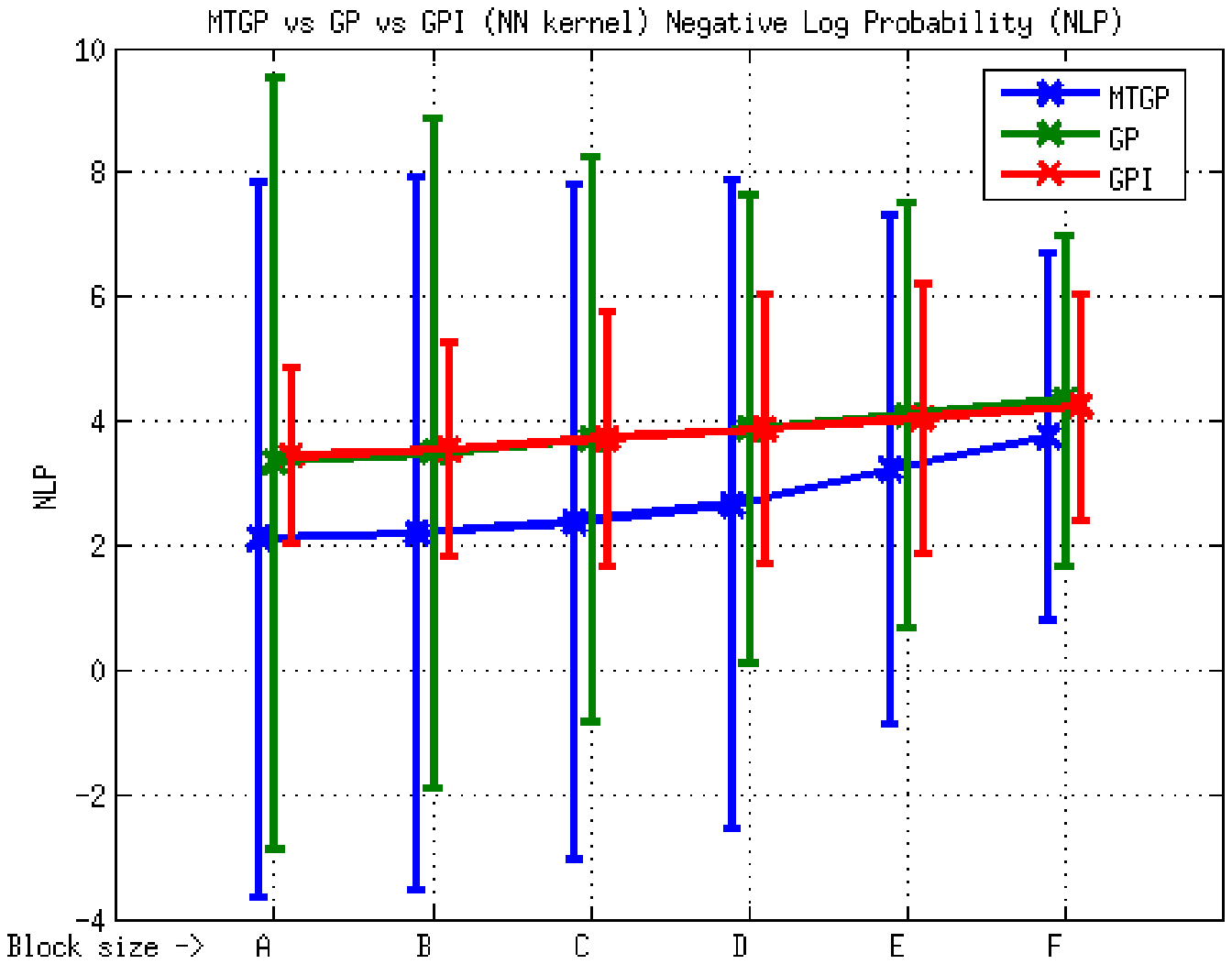}}
  \end{center}
  \caption{Element E2, MTGP vs GP vs GPI approaches, NN kernel, NLP metric. The figure above shows the average values; the one below shows the range of values obtained considering two standard deviations about the mean. Test block sizes (m) - A (22 x 11 x 2), B (44 x 22 x 4), C (84 x 45 x 9), D (174 x 89 x 18), E (348 x 177 x 35) and F (696 x 353 x 70).}
  \label{fig:e2_mtgp_gp_gpi_nn_nlp}
\end{figure}

\begin{figure}[htbp]
  \begin{center}
    \subfigure{\includegraphics[width=0.78\columnwidth]{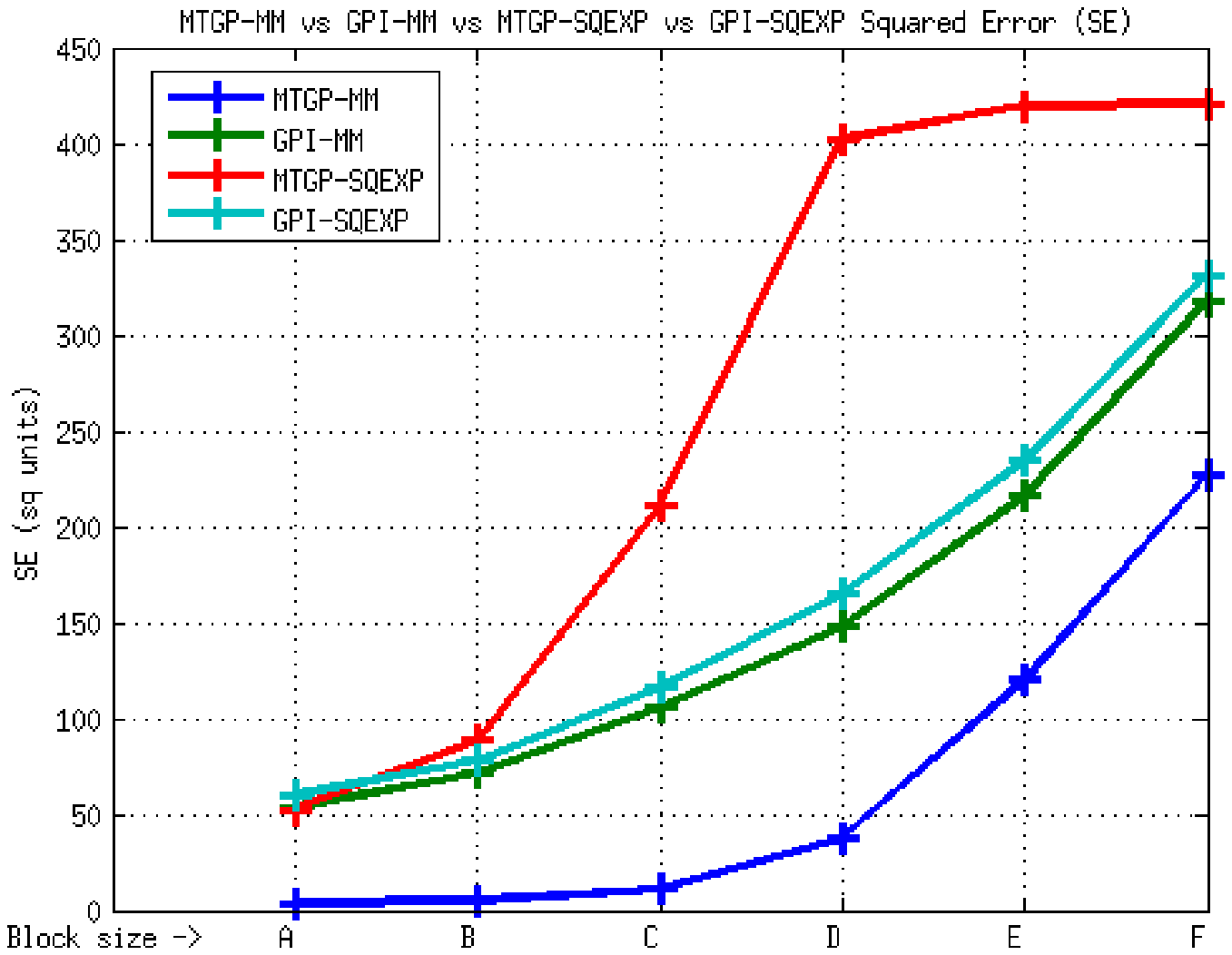}}
    \subfigure{\includegraphics[width=0.78\columnwidth]{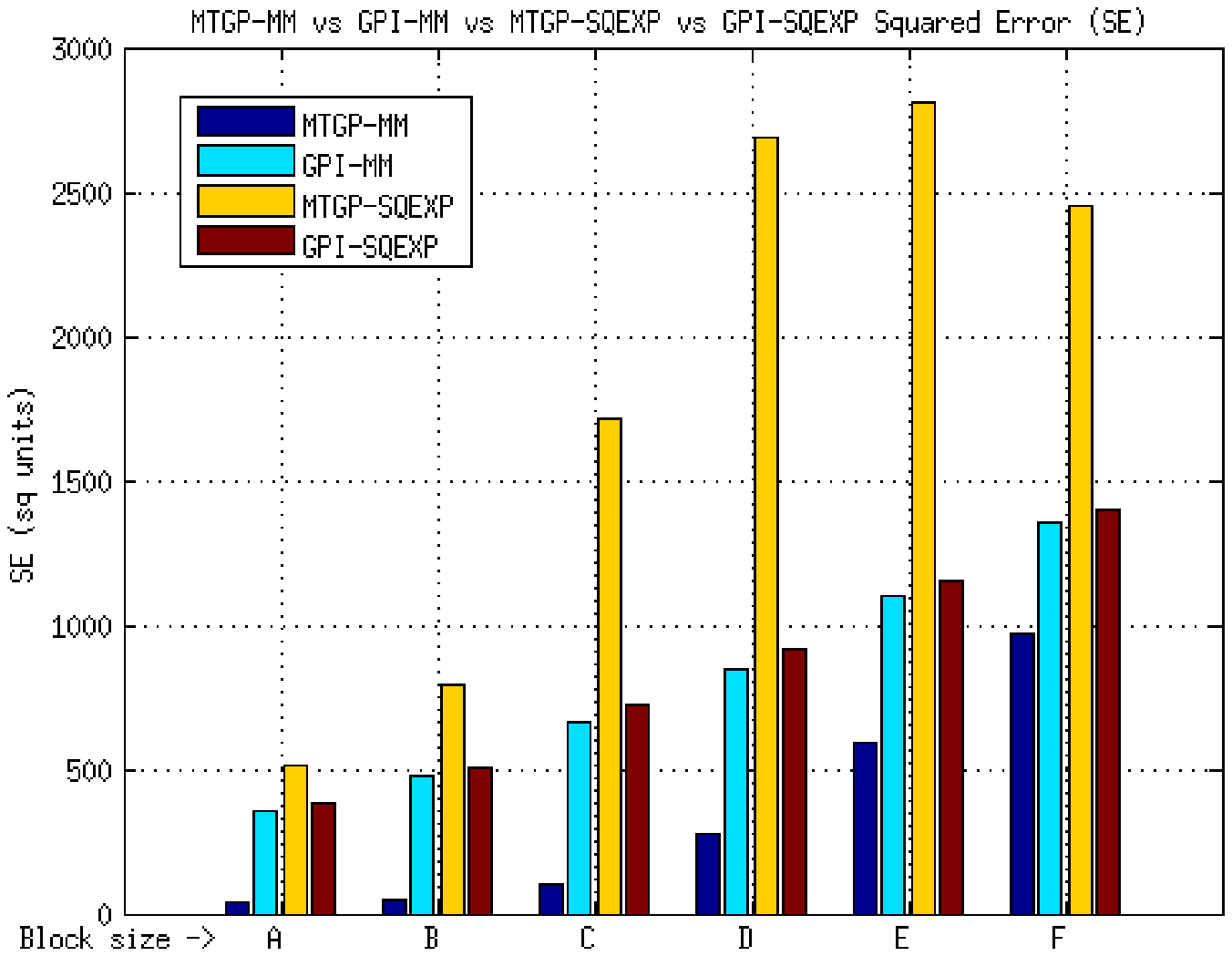}}
  \end{center}
  \caption{Element E2, MTGP vs GPI approaches, MM and SQEXP kernels, SE metric. The figure above shows the average values; the one below shows the range of values obtained considering two standard deviations about the mean. Test block sizes (m) - A (22 x 11 x 2), B (44 x 22 x 4), C (84 x 45 x 9), D (174 x 89 x 18), E (348 x 177 x 35) and F (696 x 353 x 70).}
  \label{fig:e2_mtgp_gpi_mm_sqexp_se}
\end{figure}

\begin{figure}[htbp]
  \begin{center}
    \subfigure{\includegraphics[width=0.49\columnwidth]{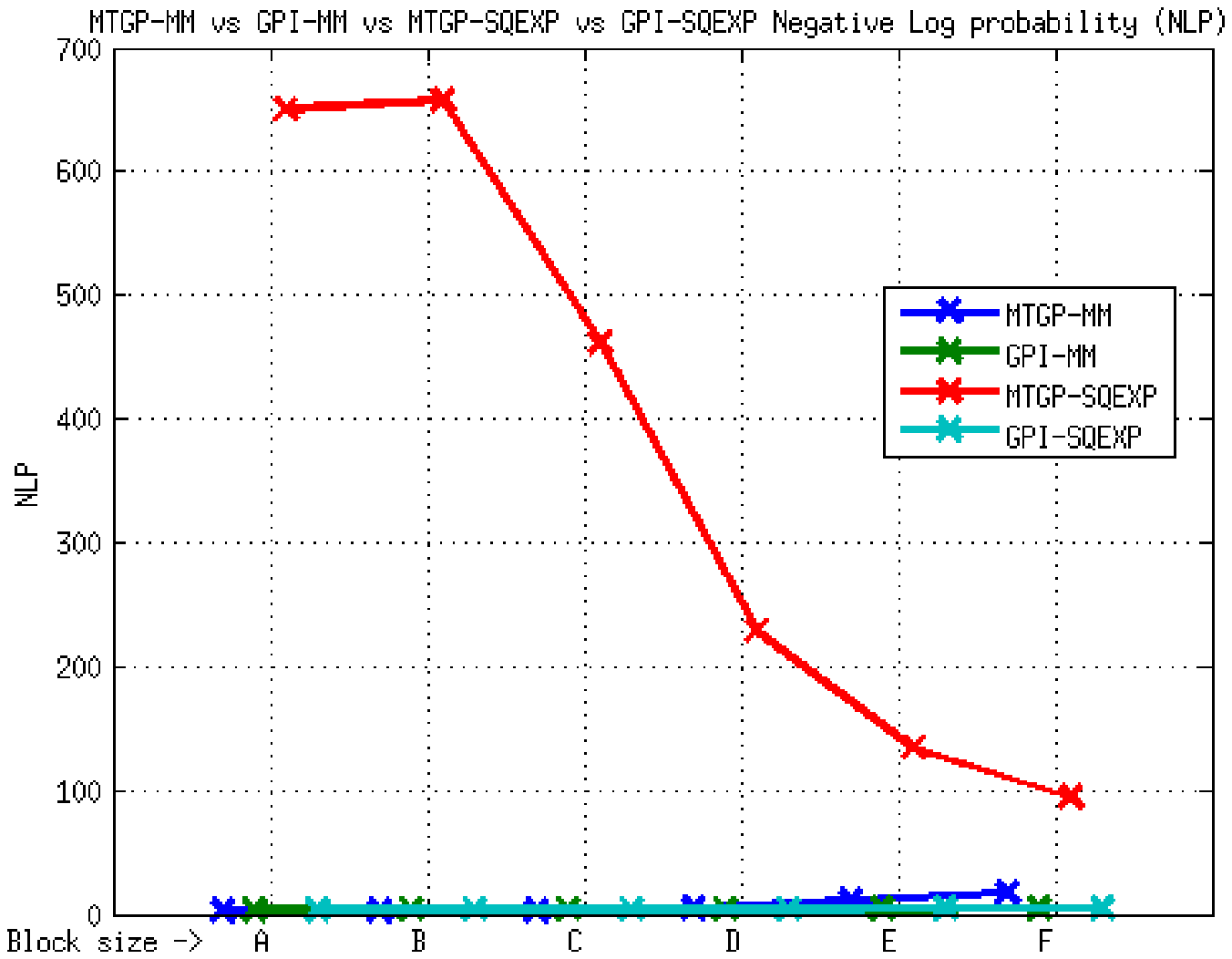}\includegraphics[width=0.49\columnwidth]{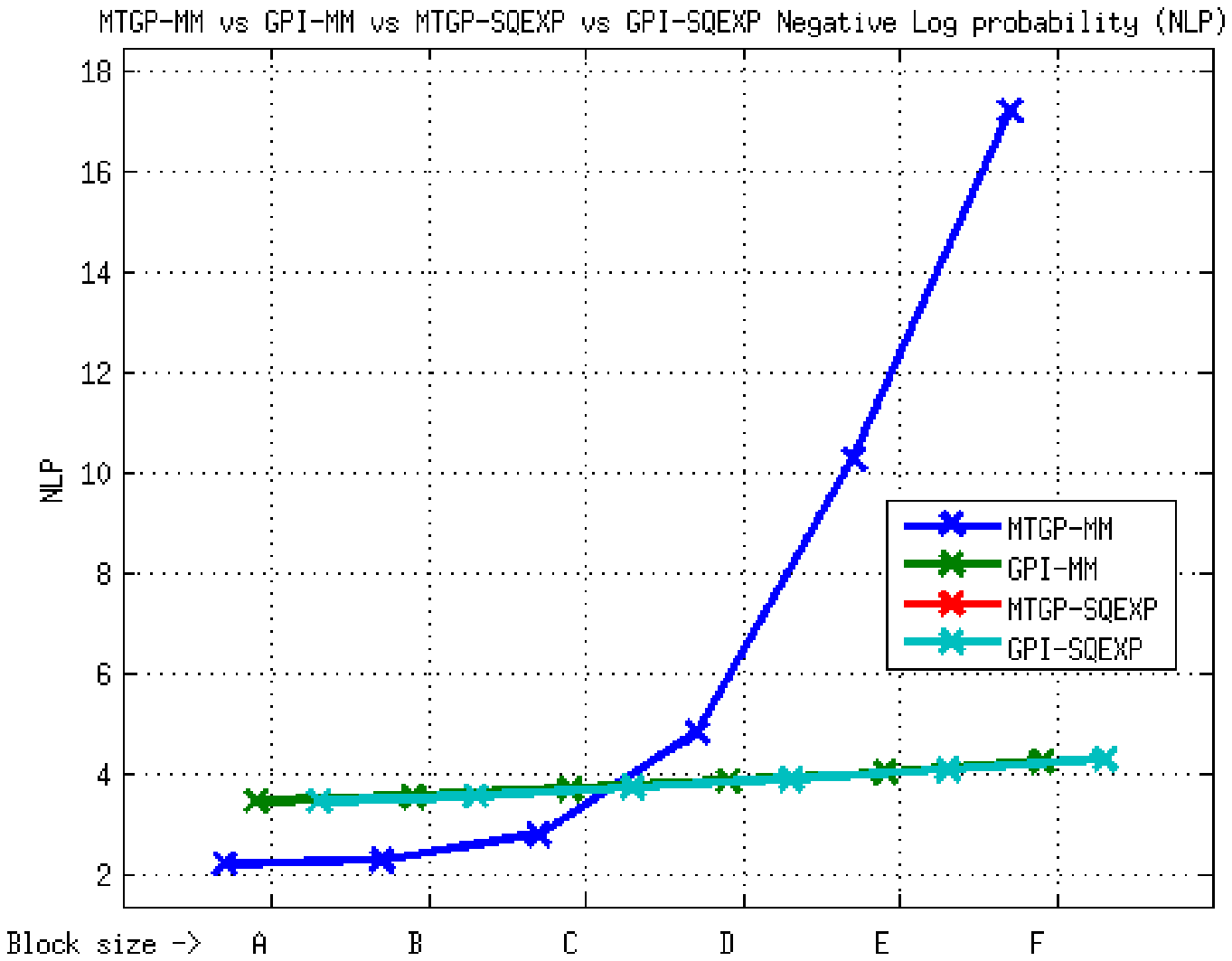}}\vspace{5mm}
    \subfigure{\includegraphics[width=0.8\columnwidth]{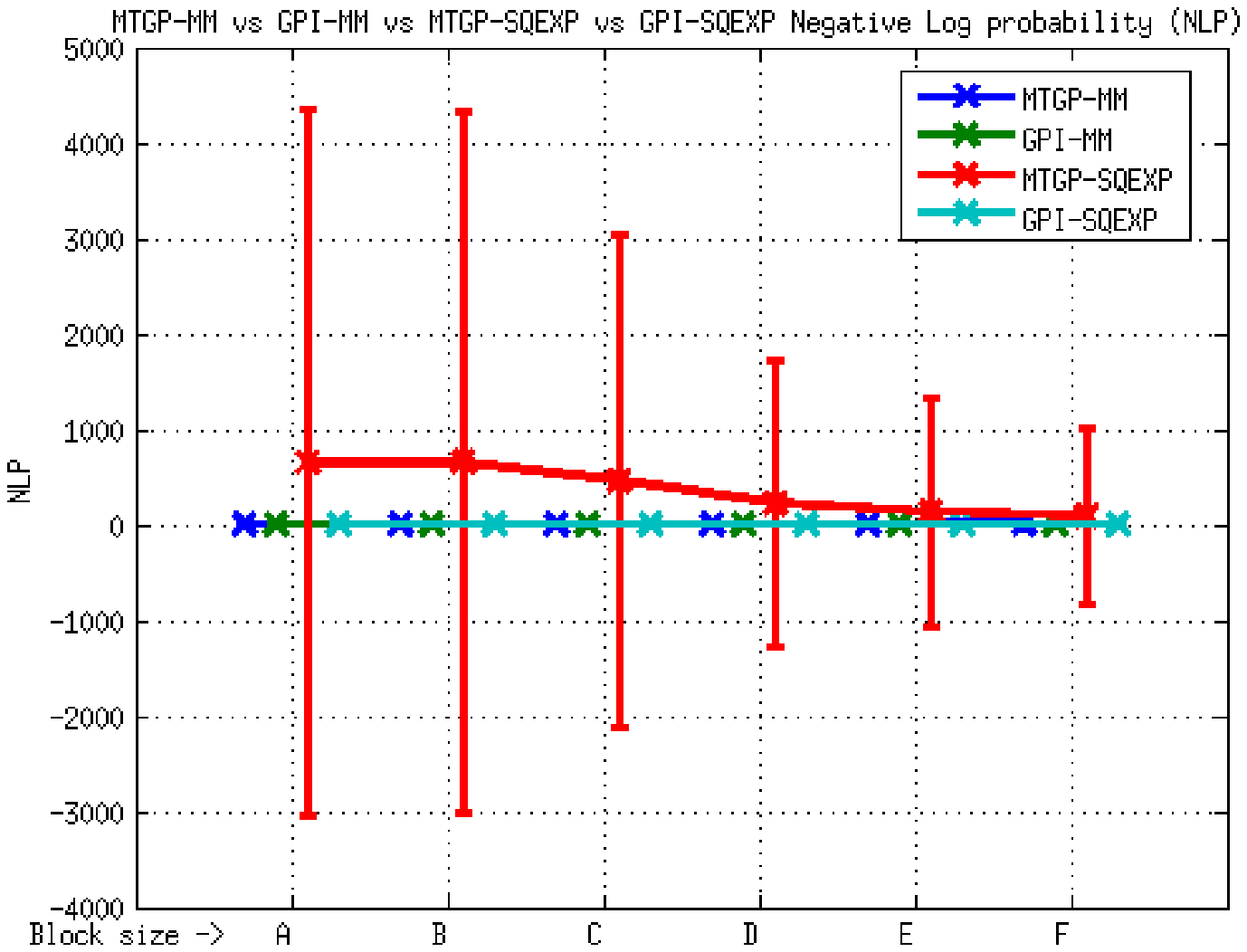}}
  \end{center}
  \caption{Element E2, MTGP vs GPI approaches, MM and SQEXP kernels, NLP metric. The figure above shows the average values, the right image being a zoomed in view of the left; the one below shows the range of values obtained considering two standard deviations about the mean. Test block sizes (m) - A (22 x 11 x 2), B (44 x 22 x 4), C (84 x 45 x 9), D (174 x 89 x 18), E (348 x 177 x 35) and F (696 x 353 x 70).}
  \label{fig:e2_mtgp_gpi_mm_sqexp_nlp}
\end{figure}

\begin{figure}[htbp]
  \begin{center}
    \subfigure{\includegraphics[width=0.78\columnwidth]{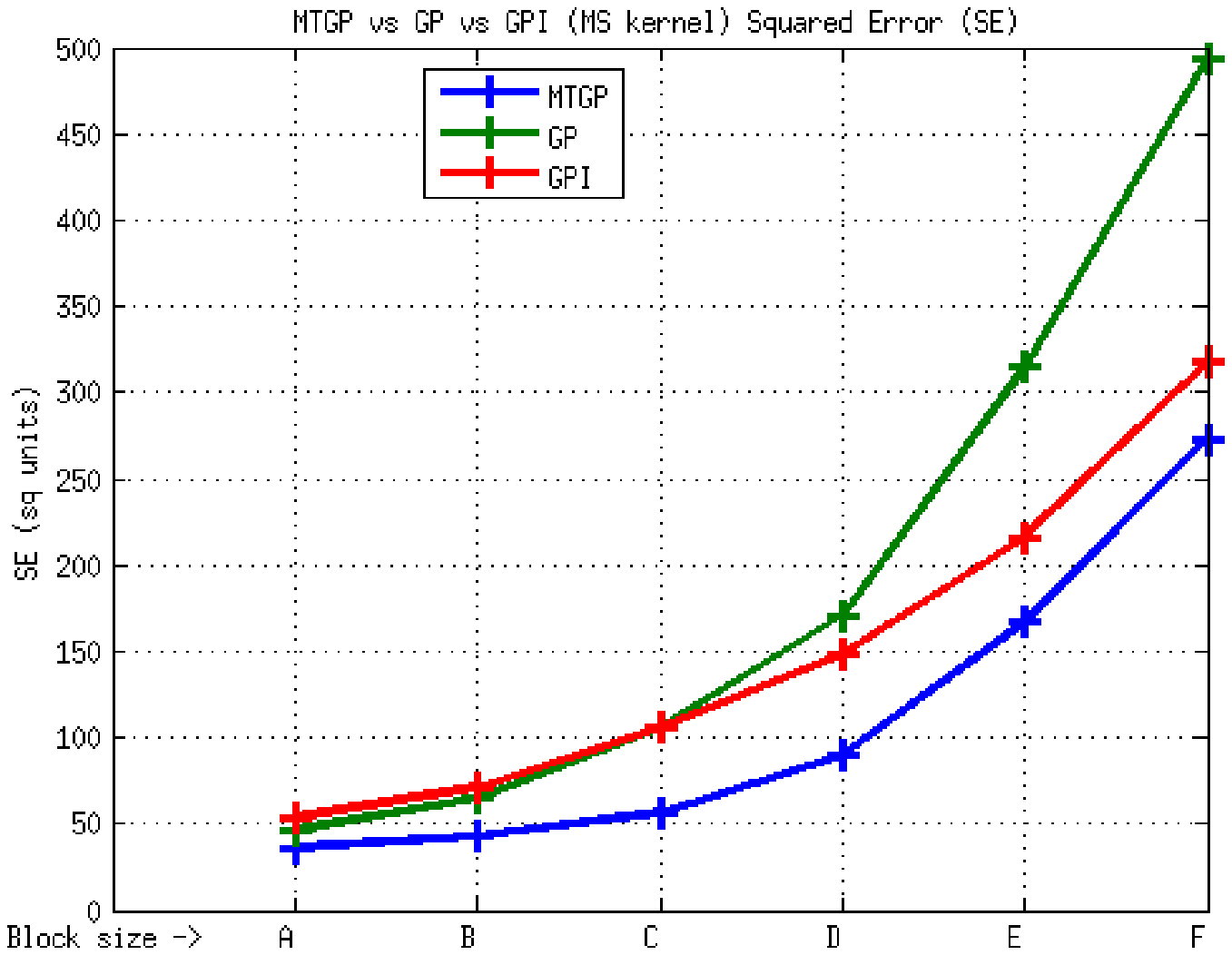}}
    \subfigure{\includegraphics[width=0.78\columnwidth]{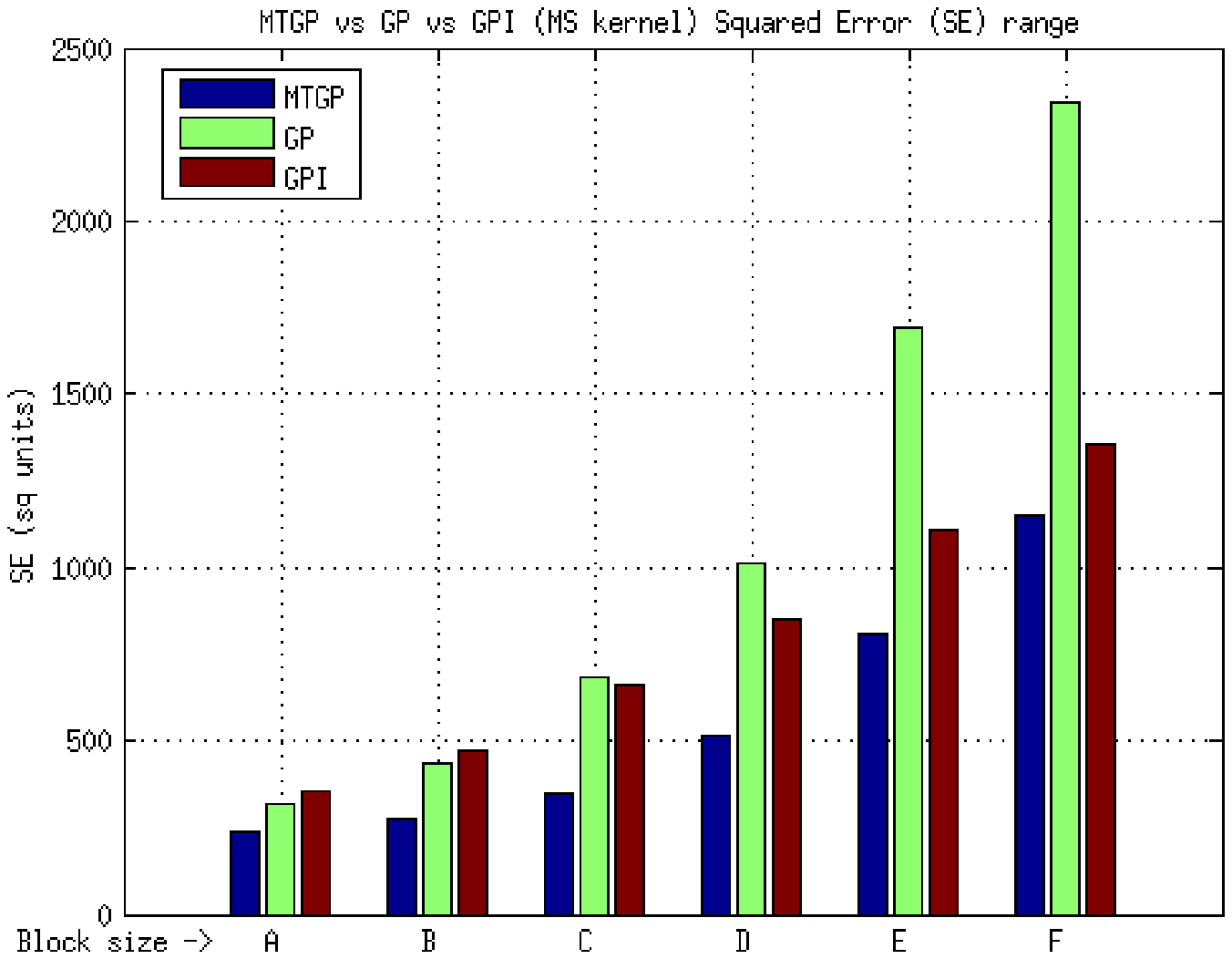}}
  \end{center}
  \caption{Element E2, MTGP vs GP vs GPI approaches, MS kernel, SE metric. The figure above shows the average values; the one below shows the range of values obtained considering two standard deviations about the mean. Test block sizes (m) - A (22 x 11 x 2), B (44 x 22 x 4), C (84 x 45 x 9), D (174 x 89 x 18), E (348 x 177 x 35) and F (696 x 353 x 70).}
  \label{fig:e2_mtgp_gp_gpi_ms_se}
\end{figure}

\begin{figure}[htbp]
  \begin{center}
    \subfigure{\includegraphics[width=0.78\columnwidth]{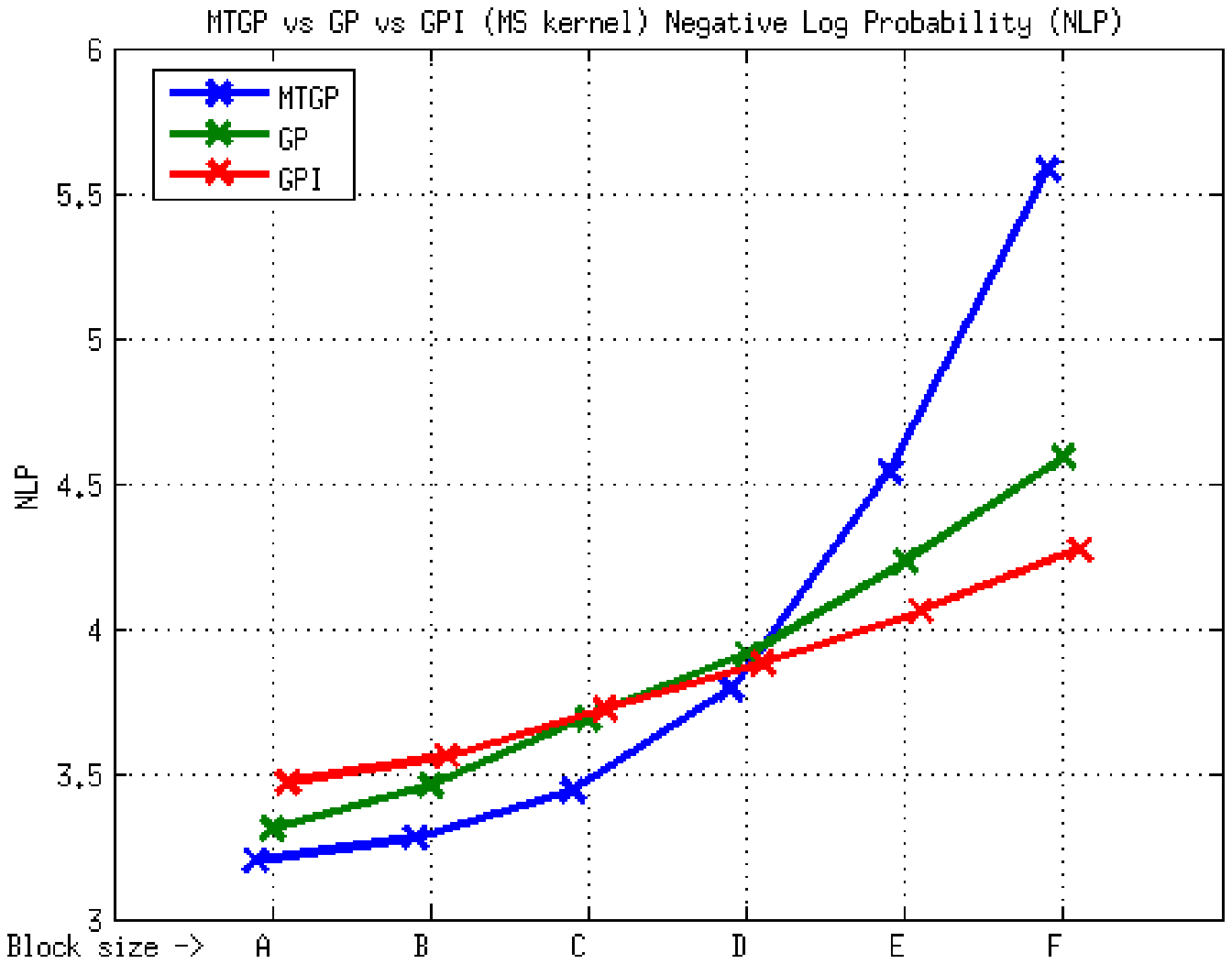}}
    \subfigure{\includegraphics[width=0.78\columnwidth]{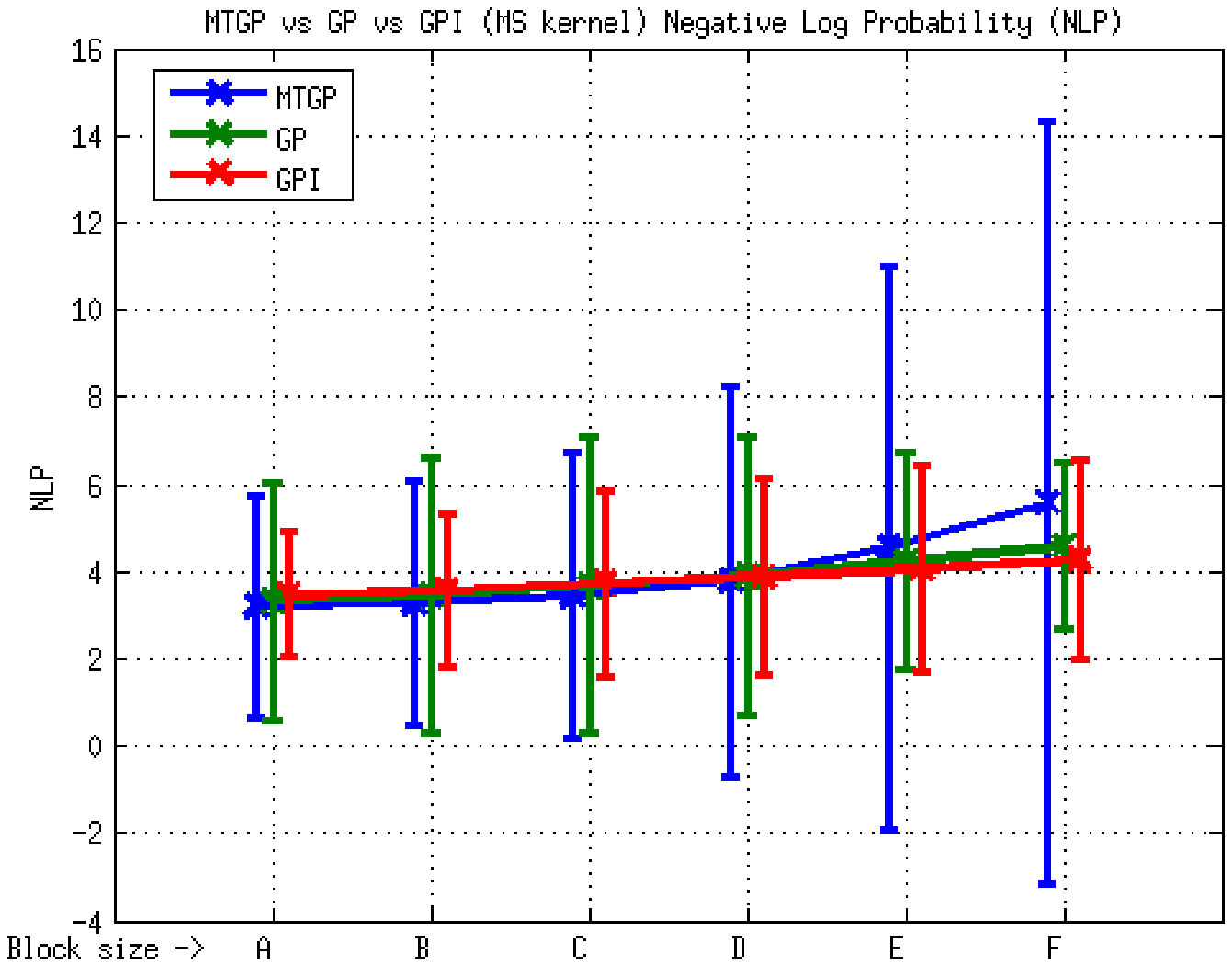}}
  \end{center}
  \caption{Element E2, MTGP vs GP vs GPI approaches, MS kernel, NLP metric. The figure above shows the average values; the one below shows the range of values obtained considering two standard deviations about the mean. Test block sizes (m) - A (22 x 11 x 2), B (44 x 22 x 4), C (84 x 45 x 9), D (174 x 89 x 18), E (348 x 177 x 35) and F (696 x 353 x 70).}
  \label{fig:e2_mtgp_gp_gpi_ms_nlp}
\end{figure}

%% ge3
\clearpage
\begin{figure}[htbp]
  \begin{center}
    \subfigure{\includegraphics[width=0.78\columnwidth]{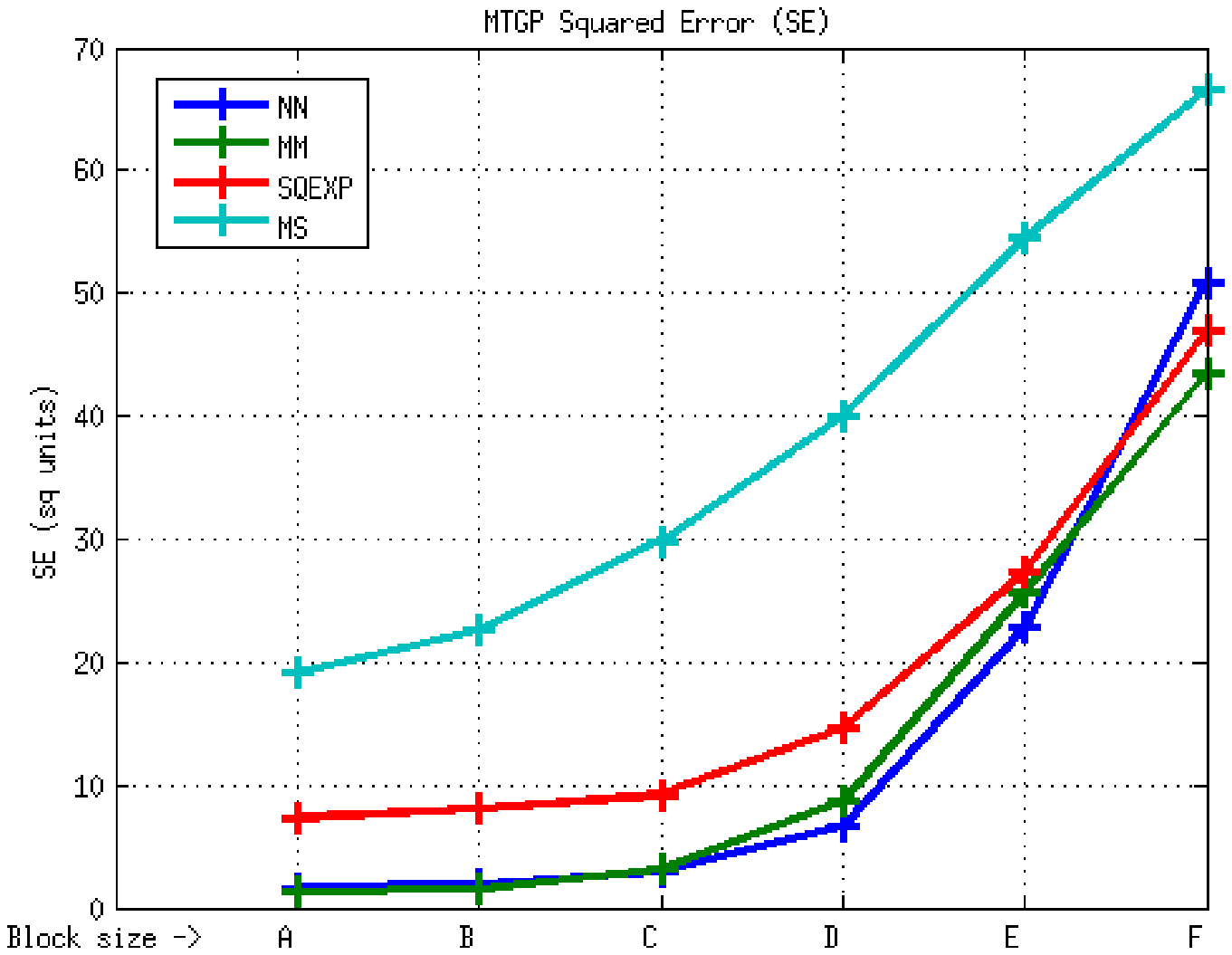}}
    \subfigure{\includegraphics[width=0.78\columnwidth]{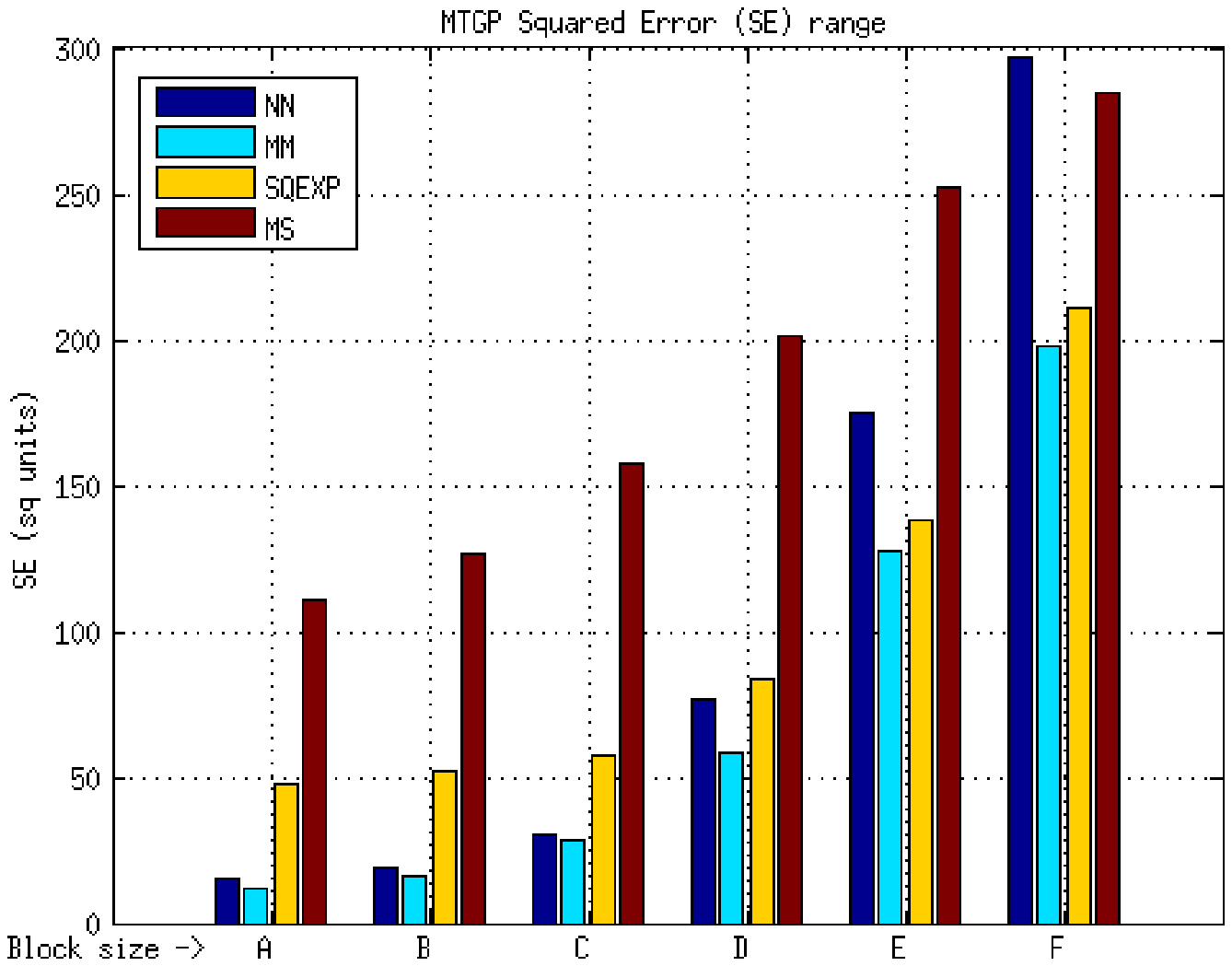}}
  \end{center}
  \caption{Element E3, MTGP approach, SE metric. The figure above shows the average values; the one below shows the range of values obtained considering two standard deviations about the mean. Test block sizes (m) - A (22 x 11 x 2), B (44 x 22 x 4), C (84 x 45 x 9), D (174 x 89 x 18), E (348 x 177 x 35) and F (696 x 353 x 70).}
  \label{fig:e3_mtgp_se}
\end{figure}

\begin{figure}[htbp]
  \begin{center}
    \subfigure{\includegraphics[width=0.78\columnwidth]{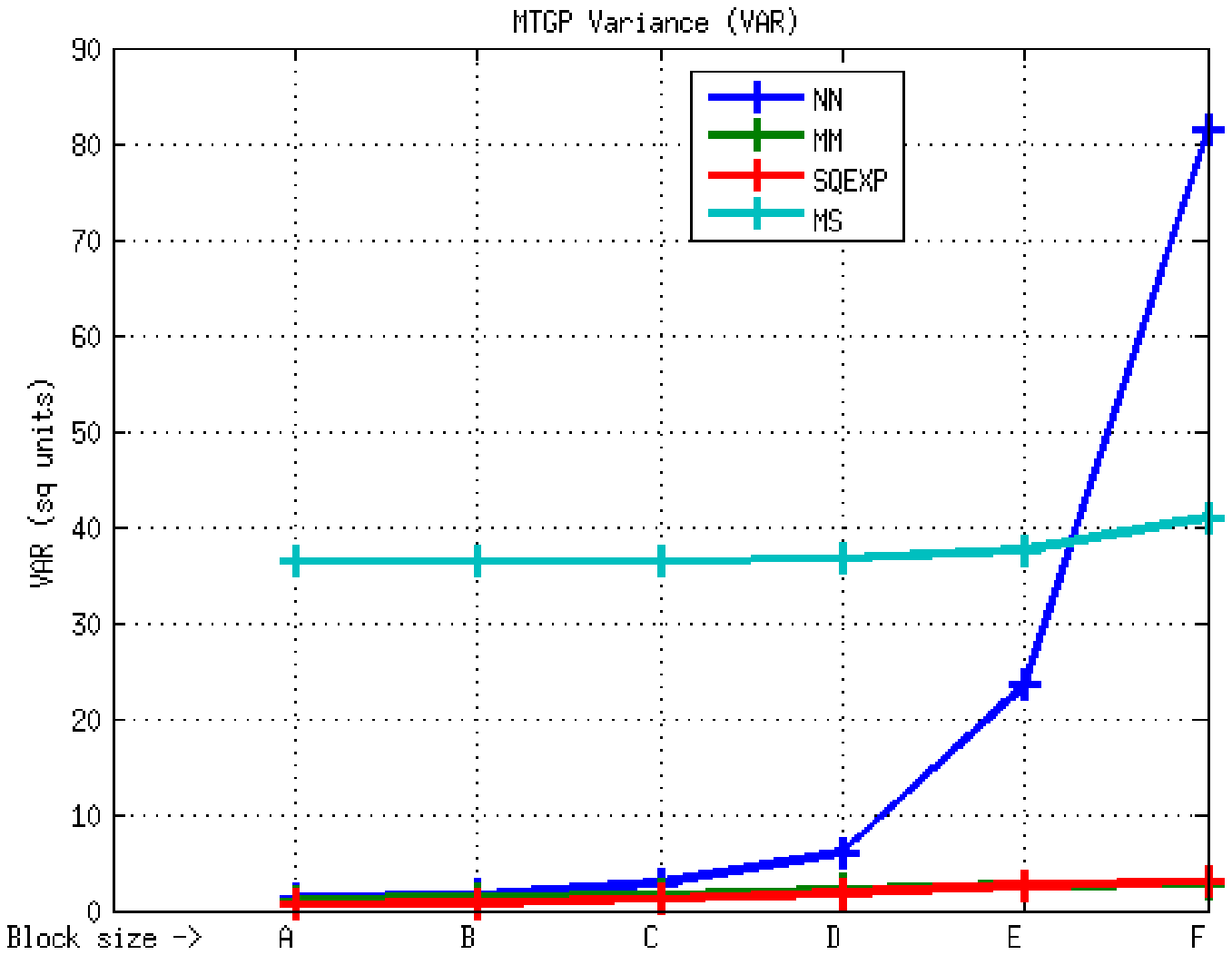}}
    \subfigure{\includegraphics[width=0.78\columnwidth]{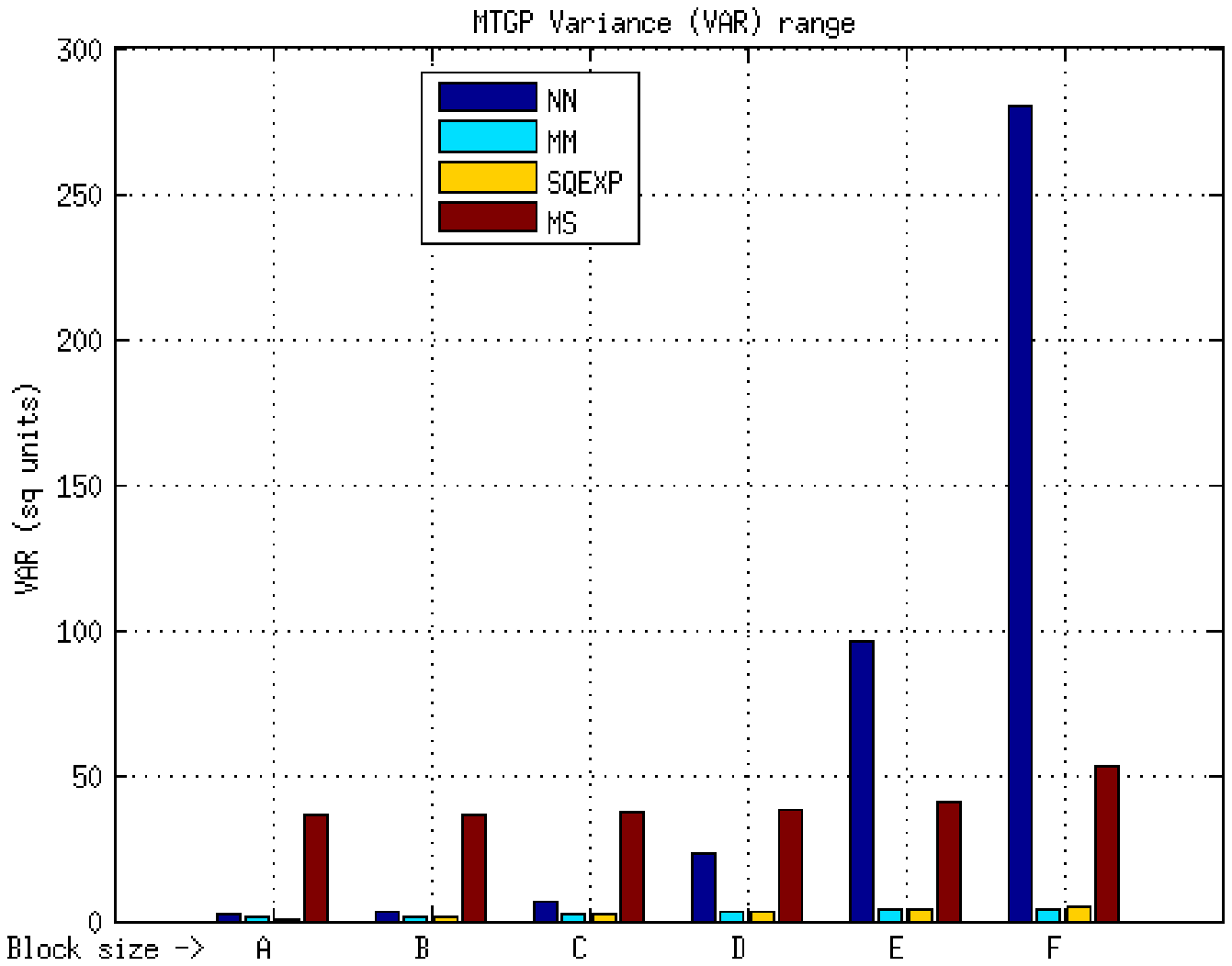}}
  \end{center}
  \caption{Element E3, MTGP approach, VAR metric. The figure above shows the average values; the one below shows the range of values obtained considering two standard deviations about the mean. Test block sizes (m) - A (22 x 11 x 2), B (44 x 22 x 4), C (84 x 45 x 9), D (174 x 89 x 18), E (348 x 177 x 35) and F (696 x 353 x 70).}
  \label{fig:e3_mtgp_var}
\end{figure}

\begin{figure}[htbp]
  \begin{center}
    \subfigure{\includegraphics[width=0.78\columnwidth]{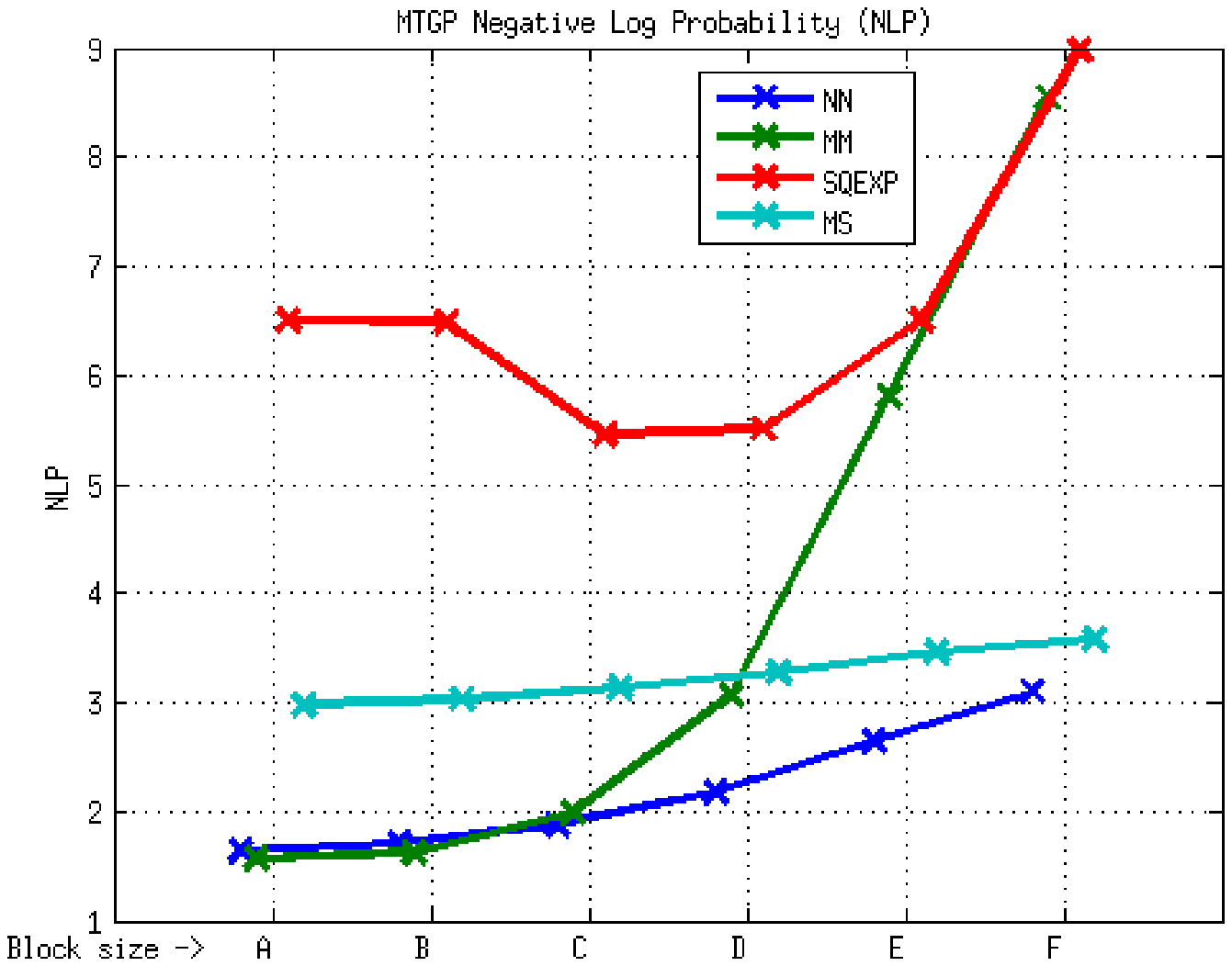}}
    \subfigure{\includegraphics[width=0.78\columnwidth]{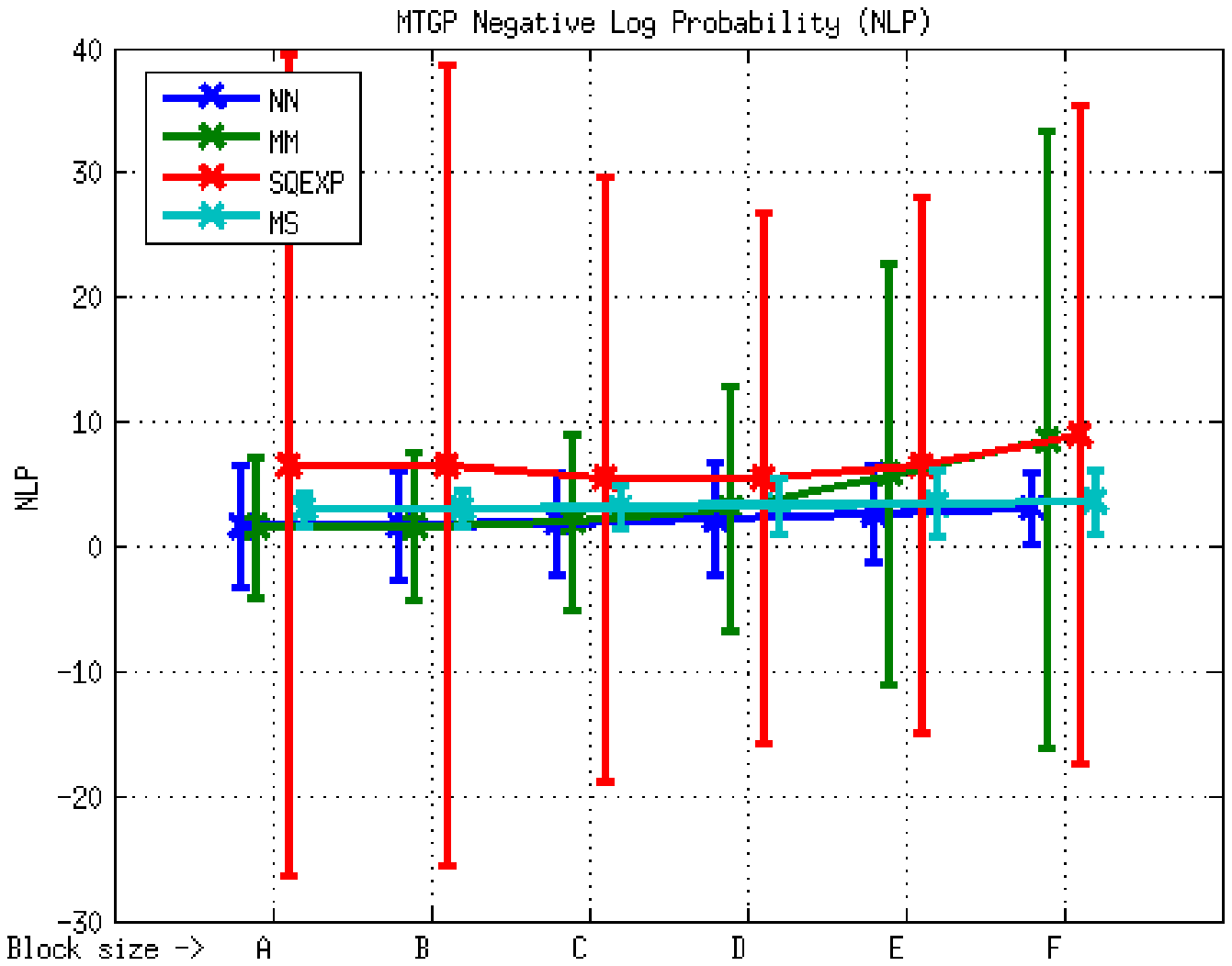}}
  \end{center}
  \caption{Element E3, MTGP approach, NLP metric. The figure above shows the average values; the one below shows the range of values obtained considering two standard deviations about the mean. Test block sizes (m) - A (22 x 11 x 2), B (44 x 22 x 4), C (84 x 45 x 9), D (174 x 89 x 18), E (348 x 177 x 35) and F (696 x 353 x 70).}
  \label{fig:e3_mtgp_nlp}
\end{figure}

\begin{figure}[htbp]
  \begin{center}
    \subfigure{\includegraphics[width=0.78\columnwidth]{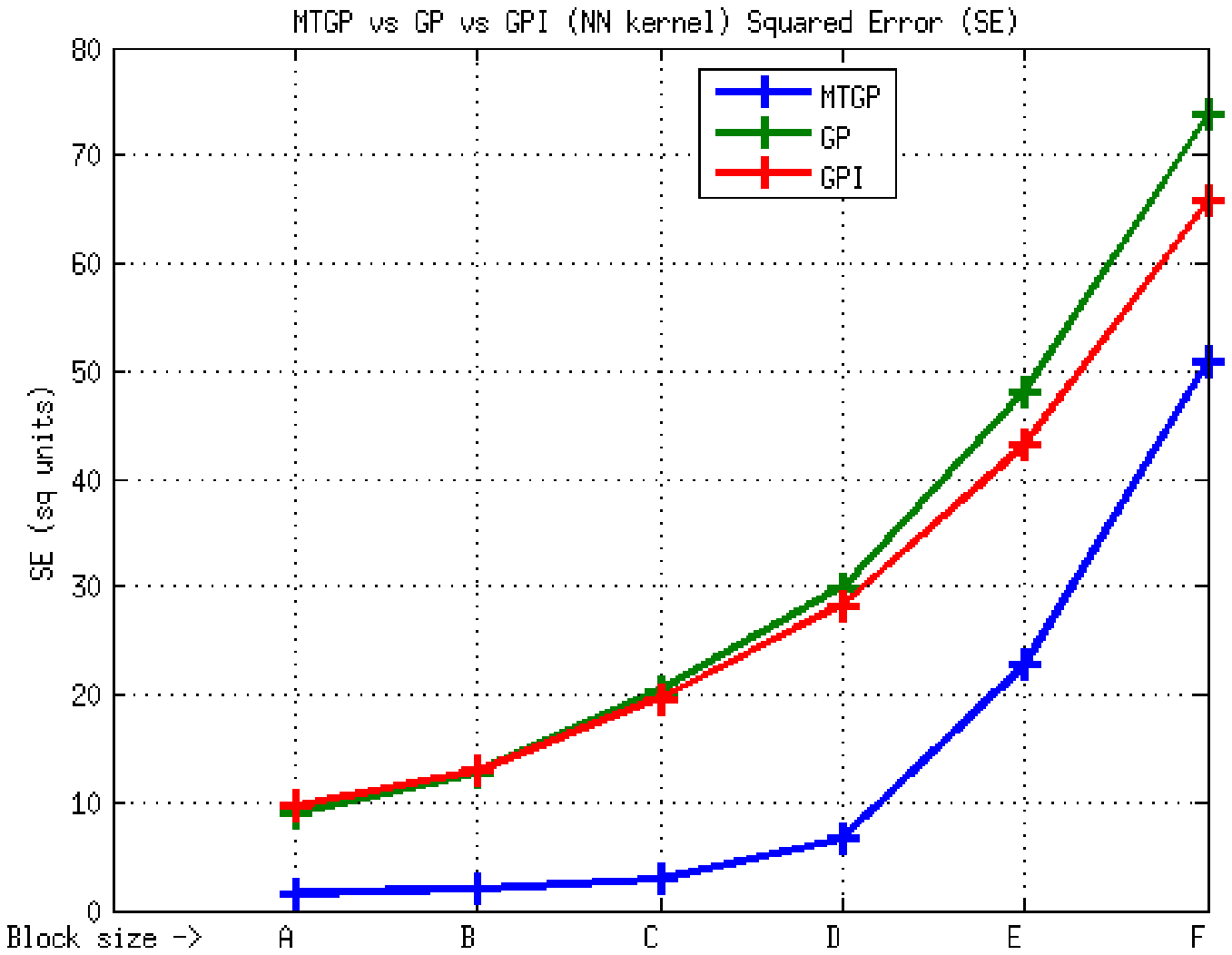}}
    \subfigure{\includegraphics[width=0.78\columnwidth]{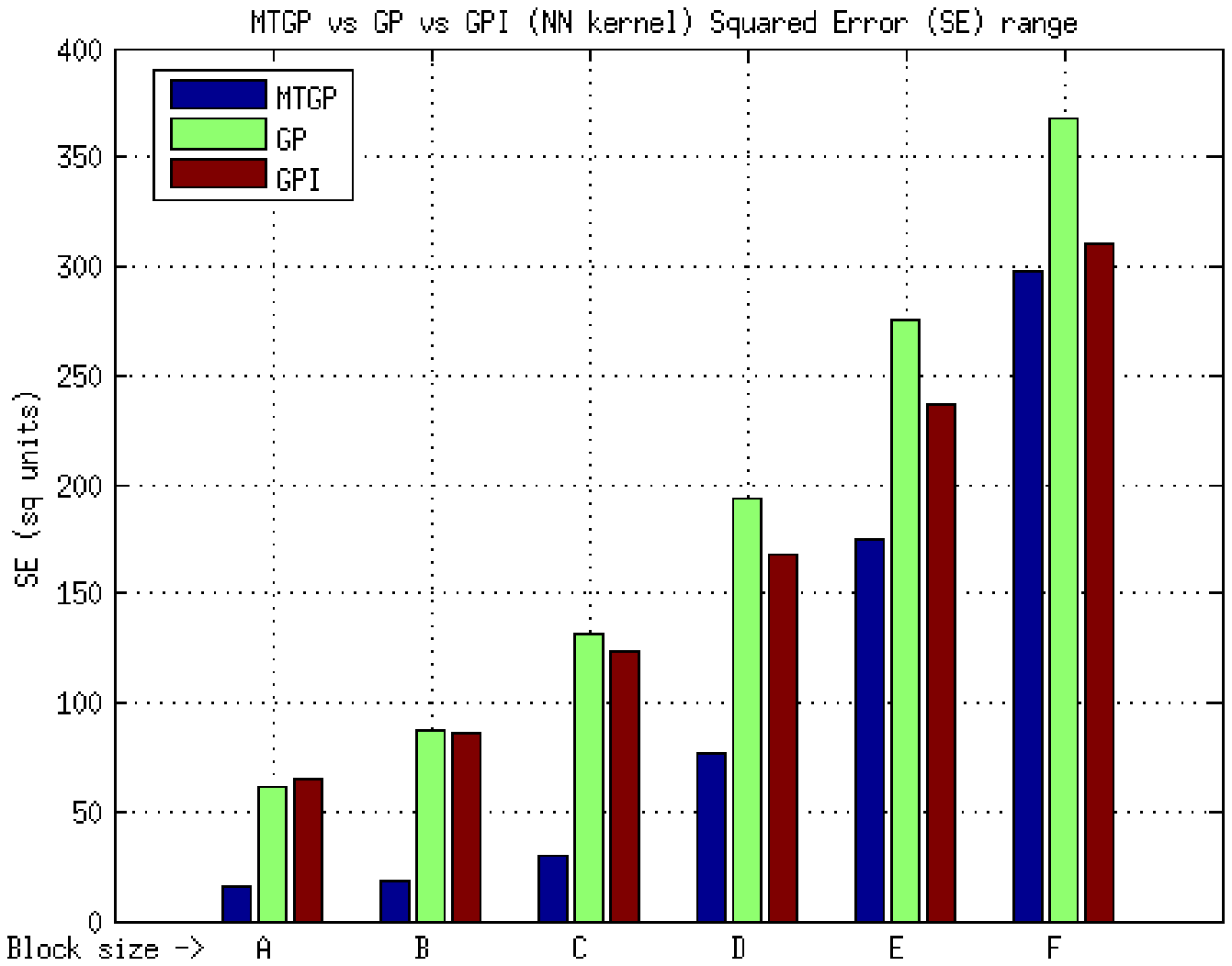}}
  \end{center}
  \caption{Element E3, MTGP vs GP vs GPI approaches, NN kernel, SE metric. The figure above shows the average values; the one below shows the range of values obtained considering two standard deviations about the mean. Test block sizes (m) - A (22 x 11 x 2), B (44 x 22 x 4), C (84 x 45 x 9), D (174 x 89 x 18), E (348 x 177 x 35) and F (696 x 353 x 70).}
  \label{fig:e3_mtgp_gp_gpi_nn_se}
\end{figure}

\begin{figure}[htbp]
  \begin{center}
    \subfigure{\includegraphics[width=0.78\columnwidth]{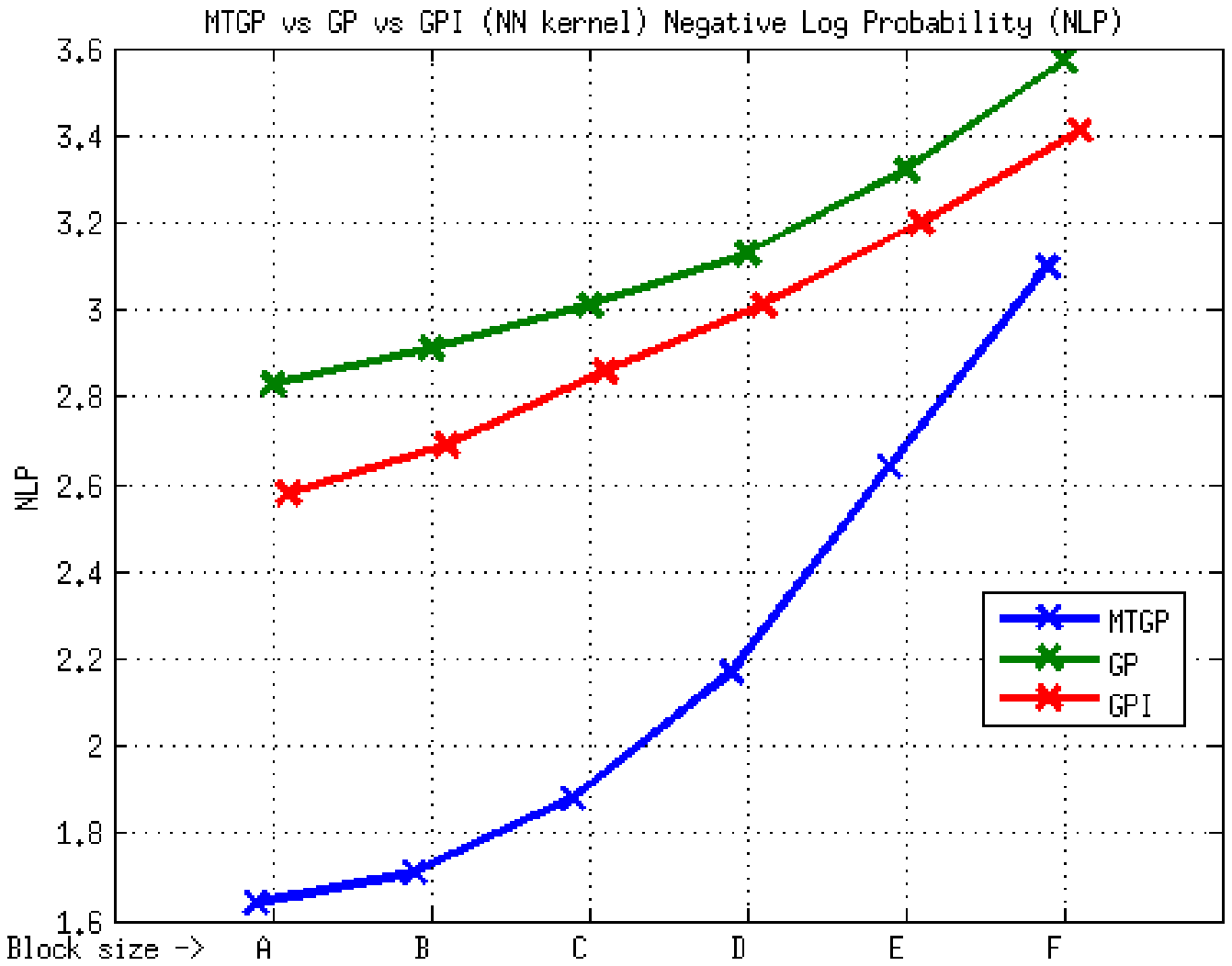}}
    \subfigure{\includegraphics[width=0.78\columnwidth]{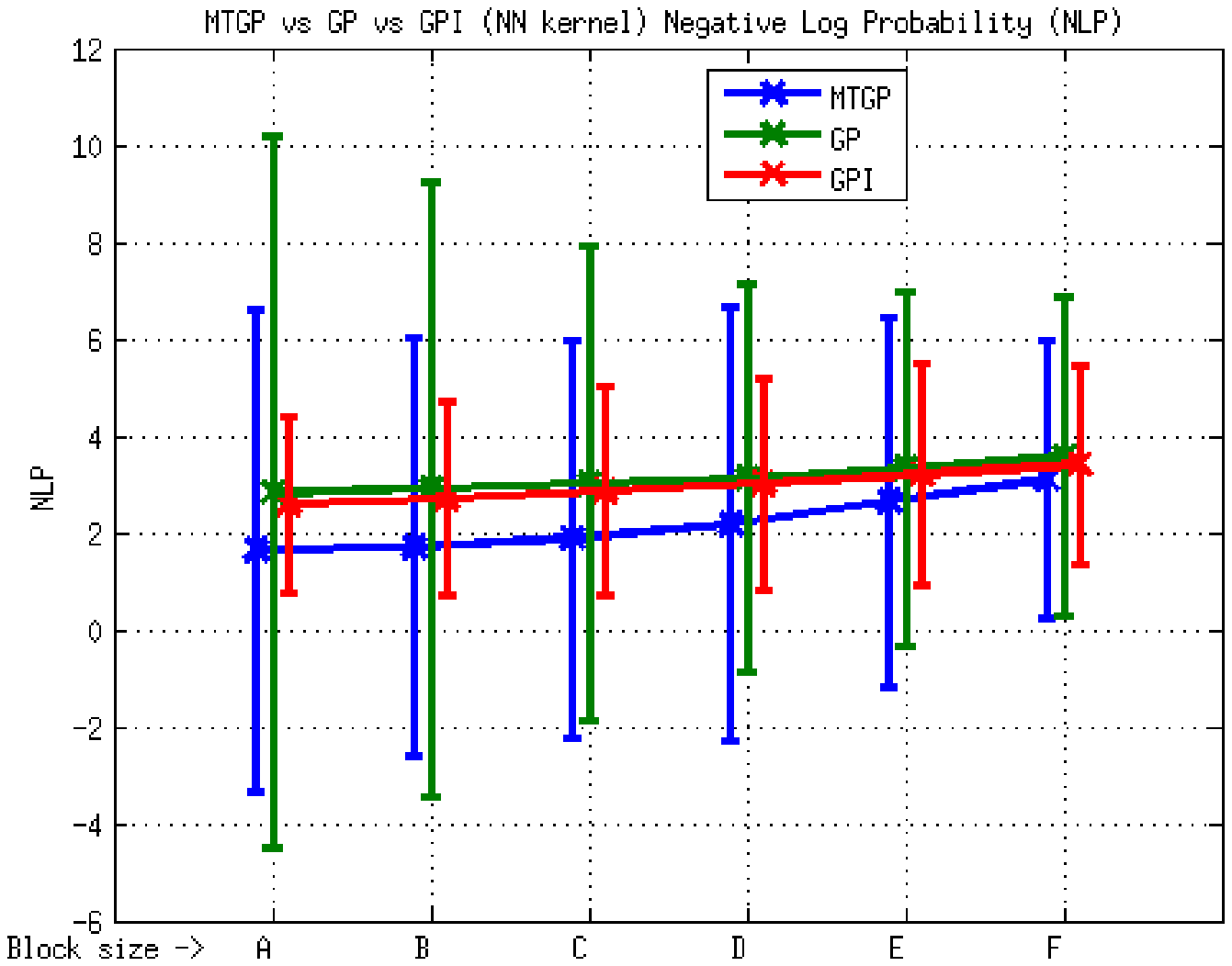}}
  \end{center}
  \caption{Element E3, MTGP vs GP vs GPI approaches, NN kernel, NLP metric. The figure above shows the average values; the one below shows the range of values obtained considering two standard deviations about the mean. Test block sizes (m) - A (22 x 11 x 2), B (44 x 22 x 4), C (84 x 45 x 9), D (174 x 89 x 18), E (348 x 177 x 35) and F (696 x 353 x 70).}
  \label{fig:e3_mtgp_gp_gpi_nn_nlp}
\end{figure}

\begin{figure}[htbp]
  \begin{center}
    \subfigure{\includegraphics[width=0.78\columnwidth]{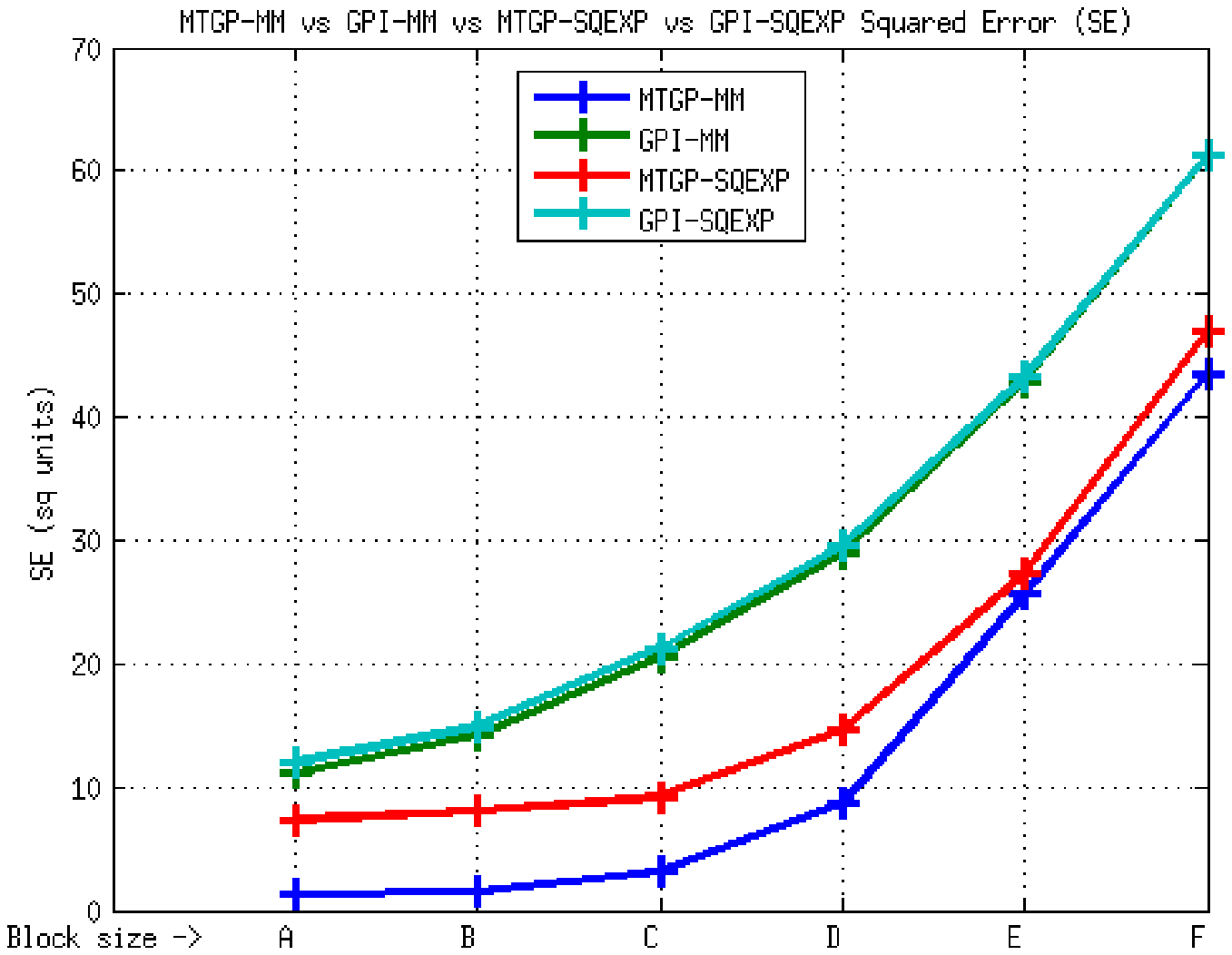}}
    \subfigure{\includegraphics[width=0.78\columnwidth]{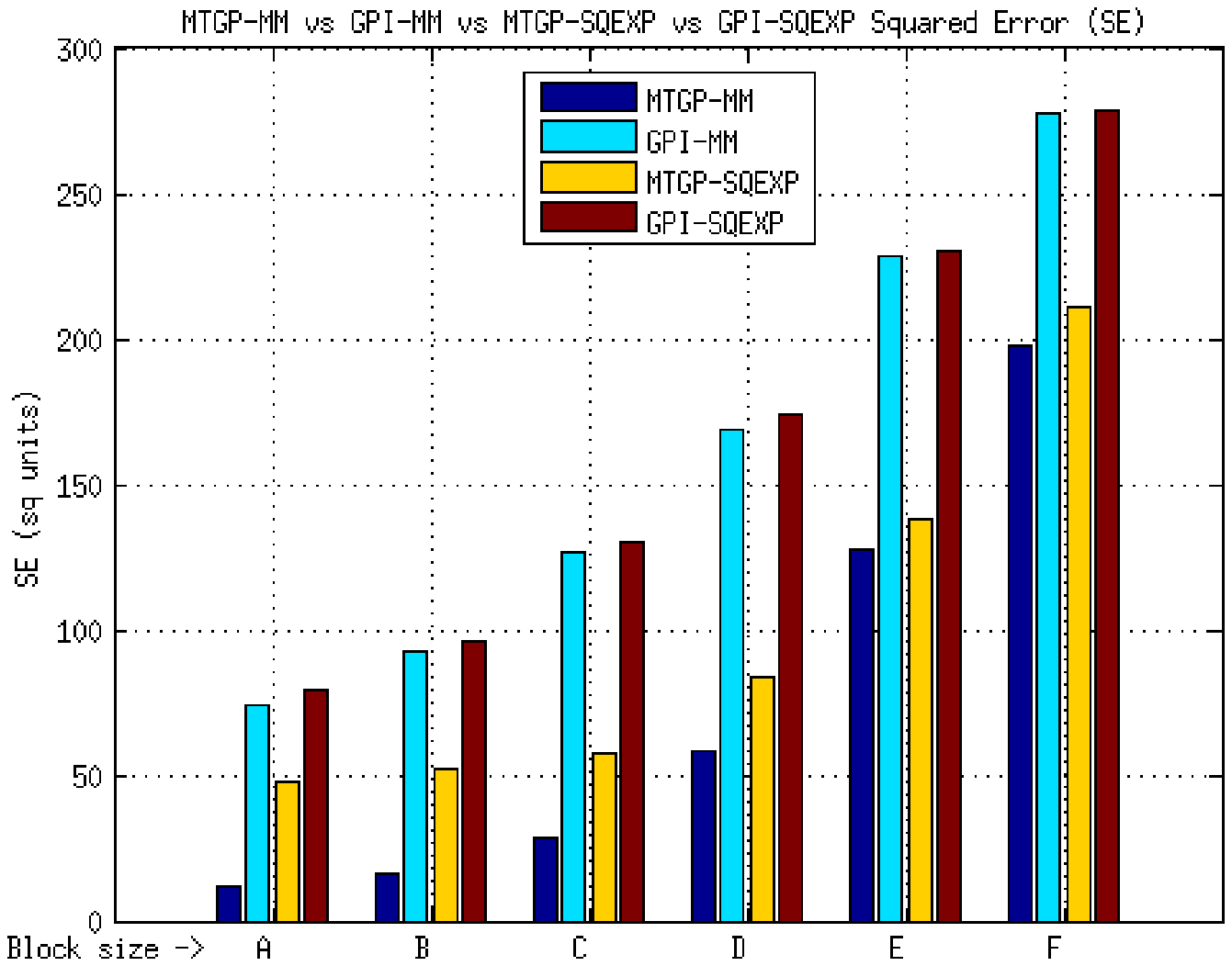}}
  \end{center}
  \caption{Element E3, MTGP vs GPI approaches, MM and SQEXP kernels, SE metric. The figure above shows the average values; the one below shows the range of values obtained considering two standard deviations about the mean. Test block sizes (m) - A (22 x 11 x 2), B (44 x 22 x 4), C (84 x 45 x 9), D (174 x 89 x 18), E (348 x 177 x 35) and F (696 x 353 x 70).}
  \label{fig:e3_mtgp_gpi_mm_sqexp_se}
\end{figure}

\begin{figure}[htbp]
  \begin{center}
    \subfigure{\includegraphics[width=0.78\columnwidth]{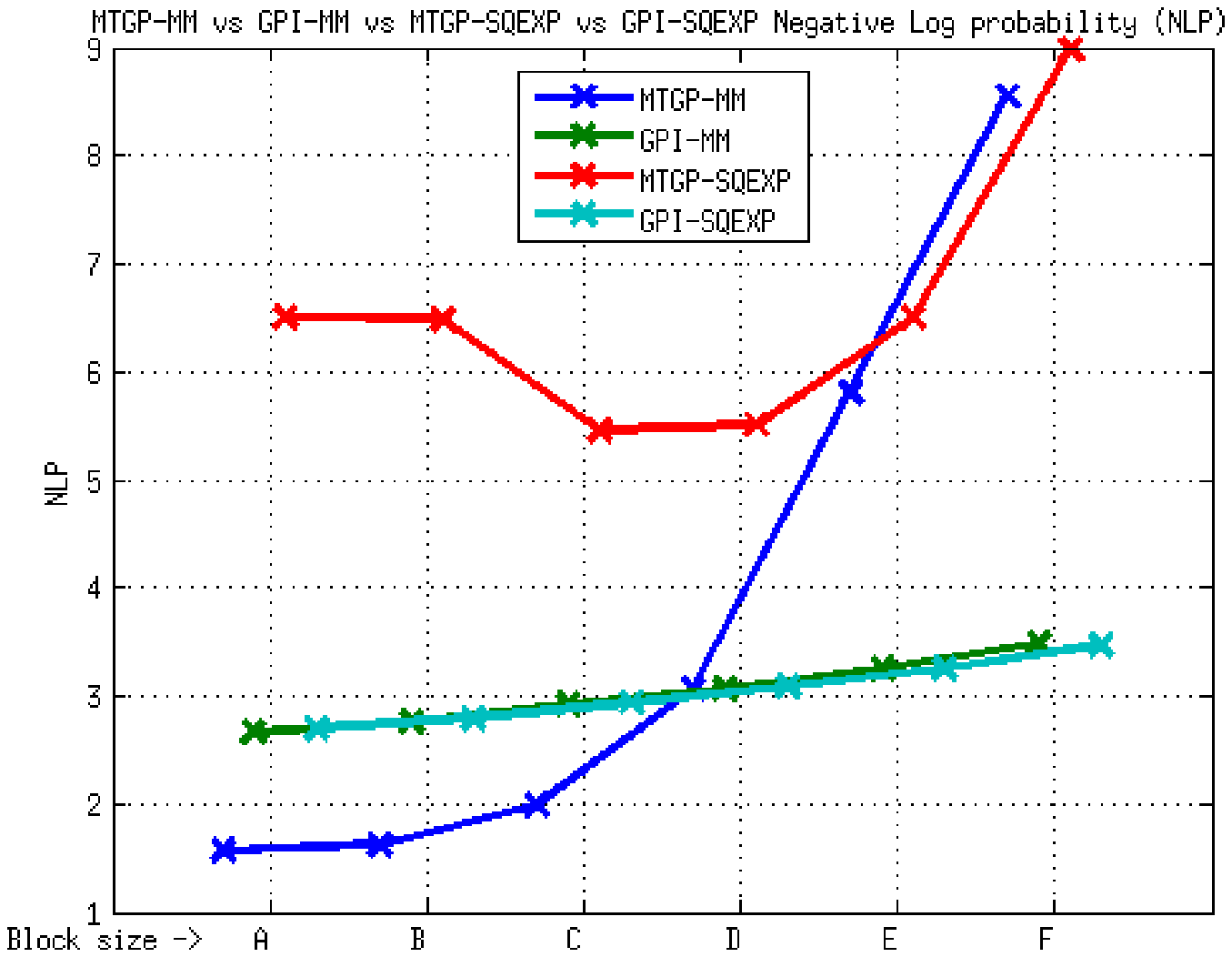}}
    \subfigure{\includegraphics[width=0.78\columnwidth]{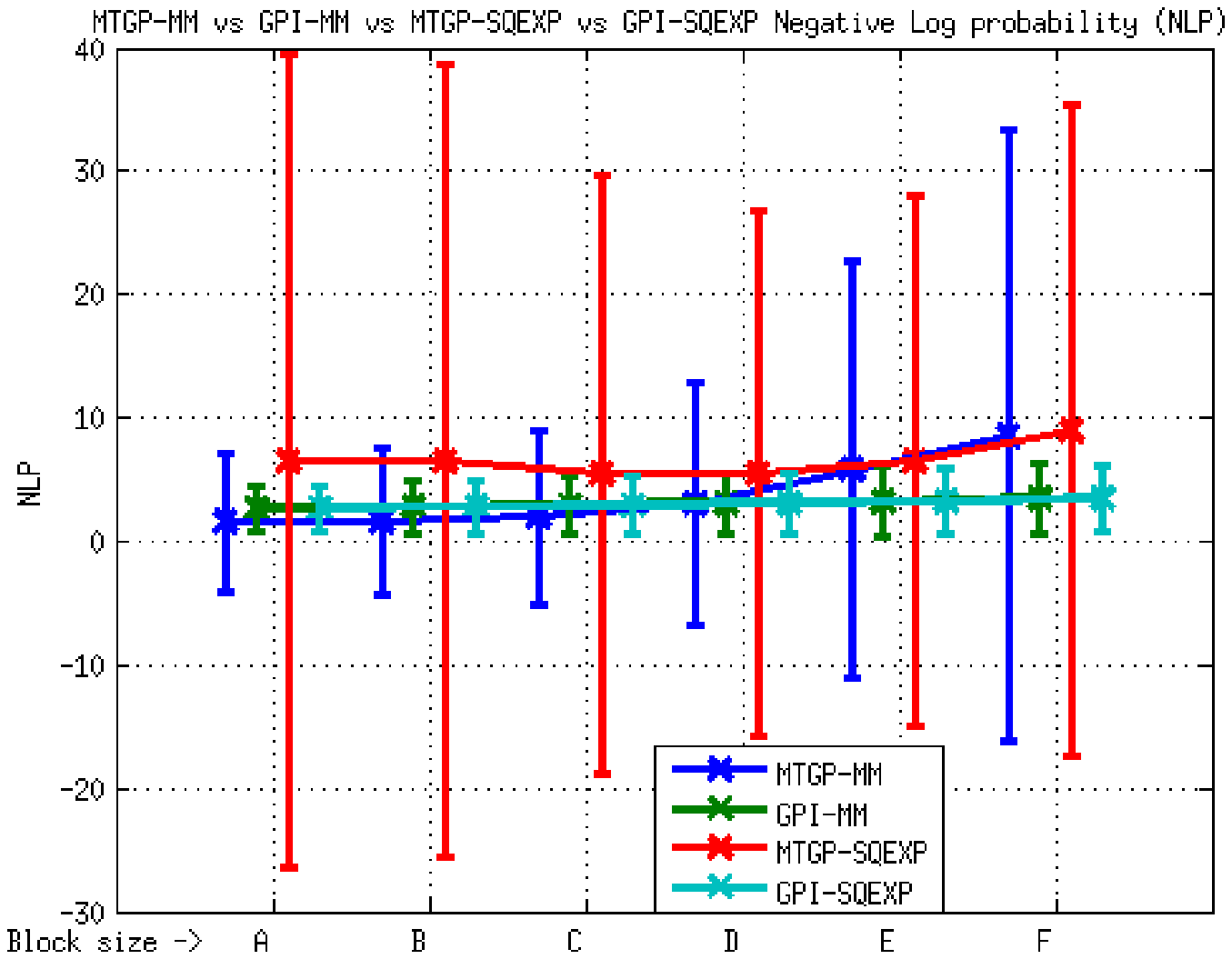}}
  \end{center}
  \caption{Element E3, MTGP vs GPI approaches, MM and SQEXP kernels, NLP metric. The figure above shows the average values; the one below shows the range of values obtained considering two standard deviations about the mean. Test block sizes (m) - A (22 x 11 x 2), B (44 x 22 x 4), C (84 x 45 x 9), D (174 x 89 x 18), E (348 x 177 x 35) and F (696 x 353 x 70).}
  \label{fig:e3_mtgp_gpi_mm_sqexp_nlp}
\end{figure}

\begin{figure}[htbp]
  \begin{center}
    \subfigure{\includegraphics[width=0.78\columnwidth]{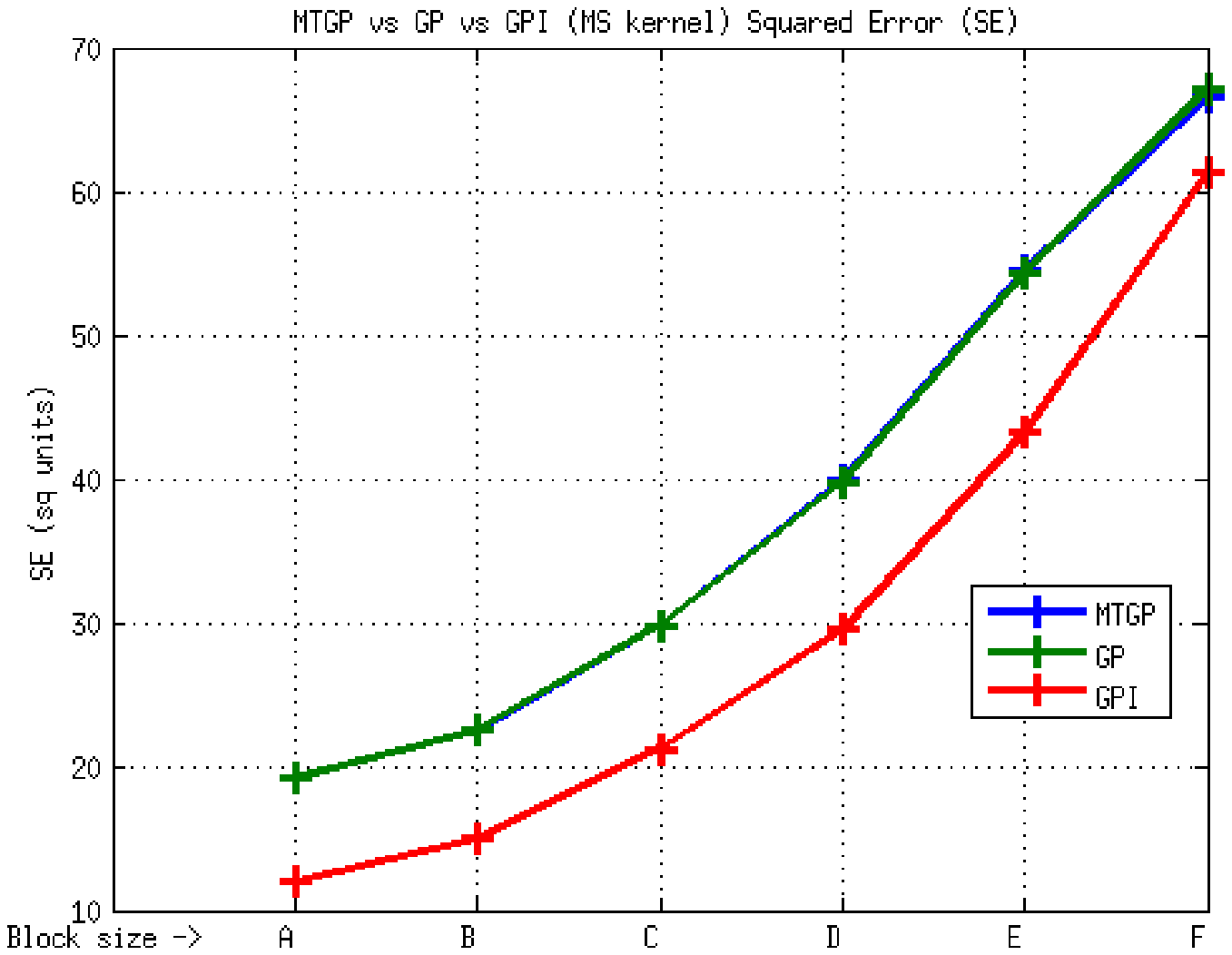}}
    \subfigure{\includegraphics[width=0.78\columnwidth]{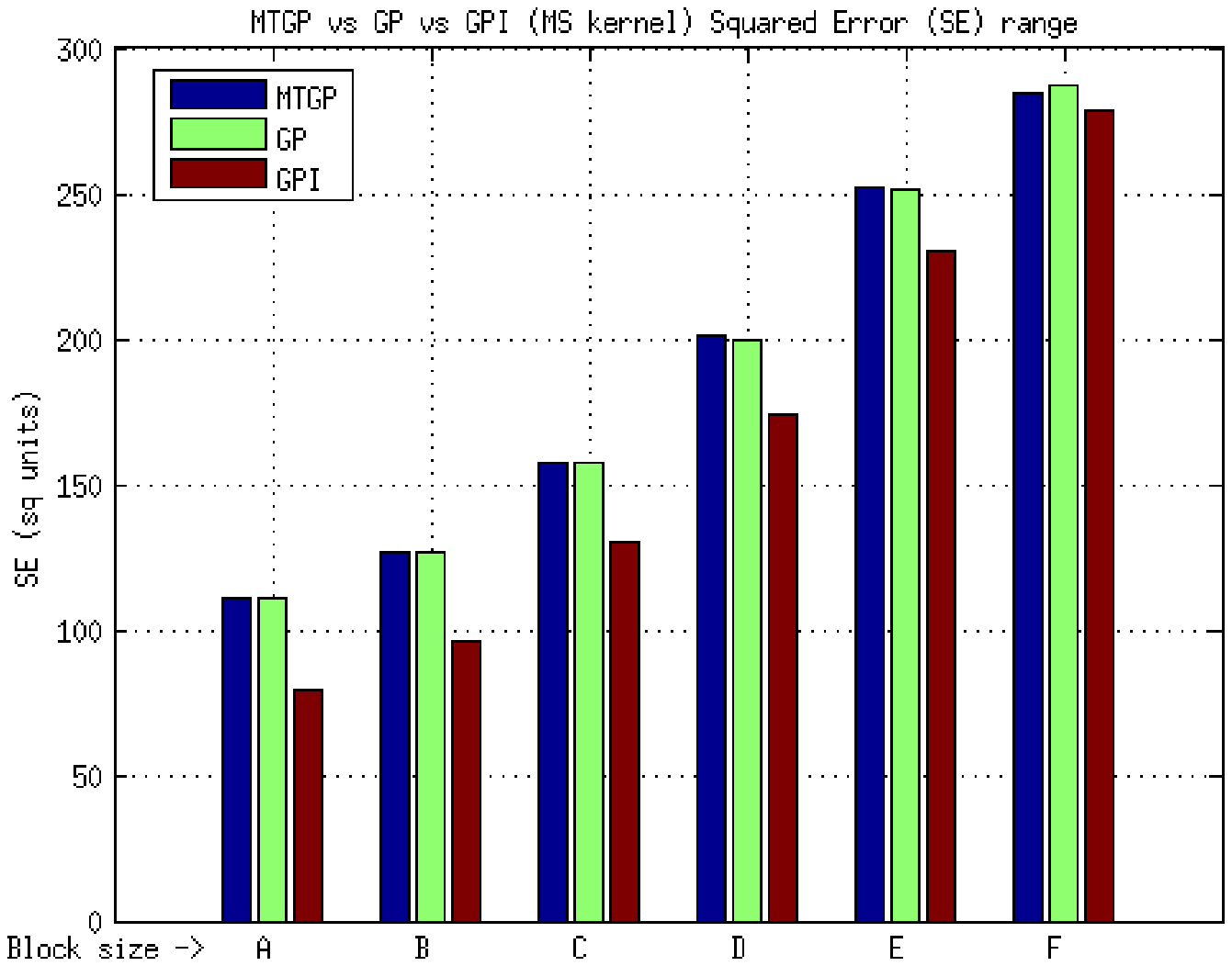}}
  \end{center}
  \caption{Element E3, MTGP vs GP vs GPI approaches, MS kernel, SE metric. The figure above shows the average values; the one below shows the range of values obtained considering two standard deviations about the mean. Test block sizes (m) - A (22 x 11 x 2), B (44 x 22 x 4), C (84 x 45 x 9), D (174 x 89 x 18), E (348 x 177 x 35) and F (696 x 353 x 70).}
  \label{fig:e3_mtgp_gp_gpi_ms_se}
\end{figure}

\begin{figure}[htbp]
  \begin{center}
    \subfigure{\includegraphics[width=0.78\columnwidth]{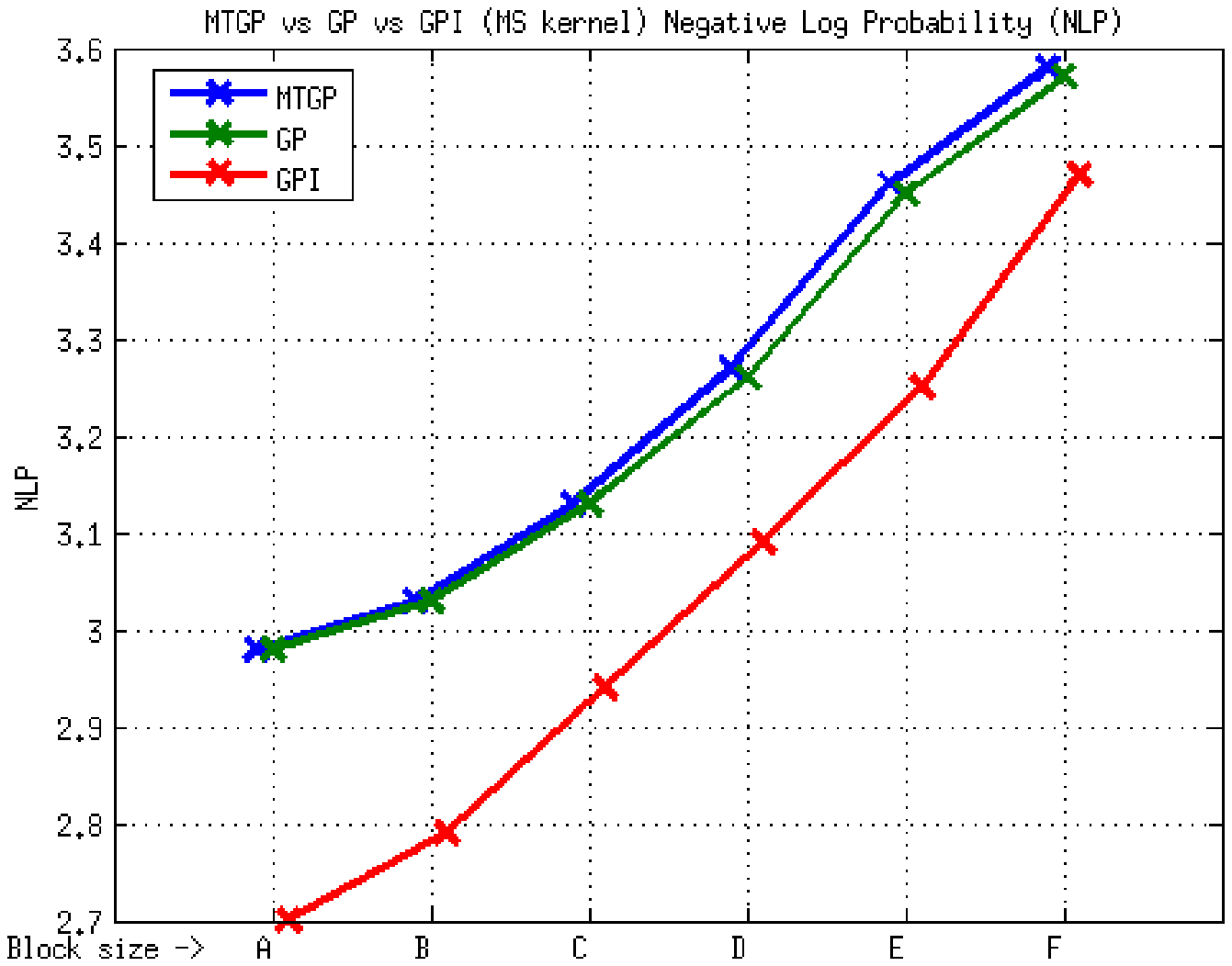}}
    \subfigure{\includegraphics[width=0.78\columnwidth]{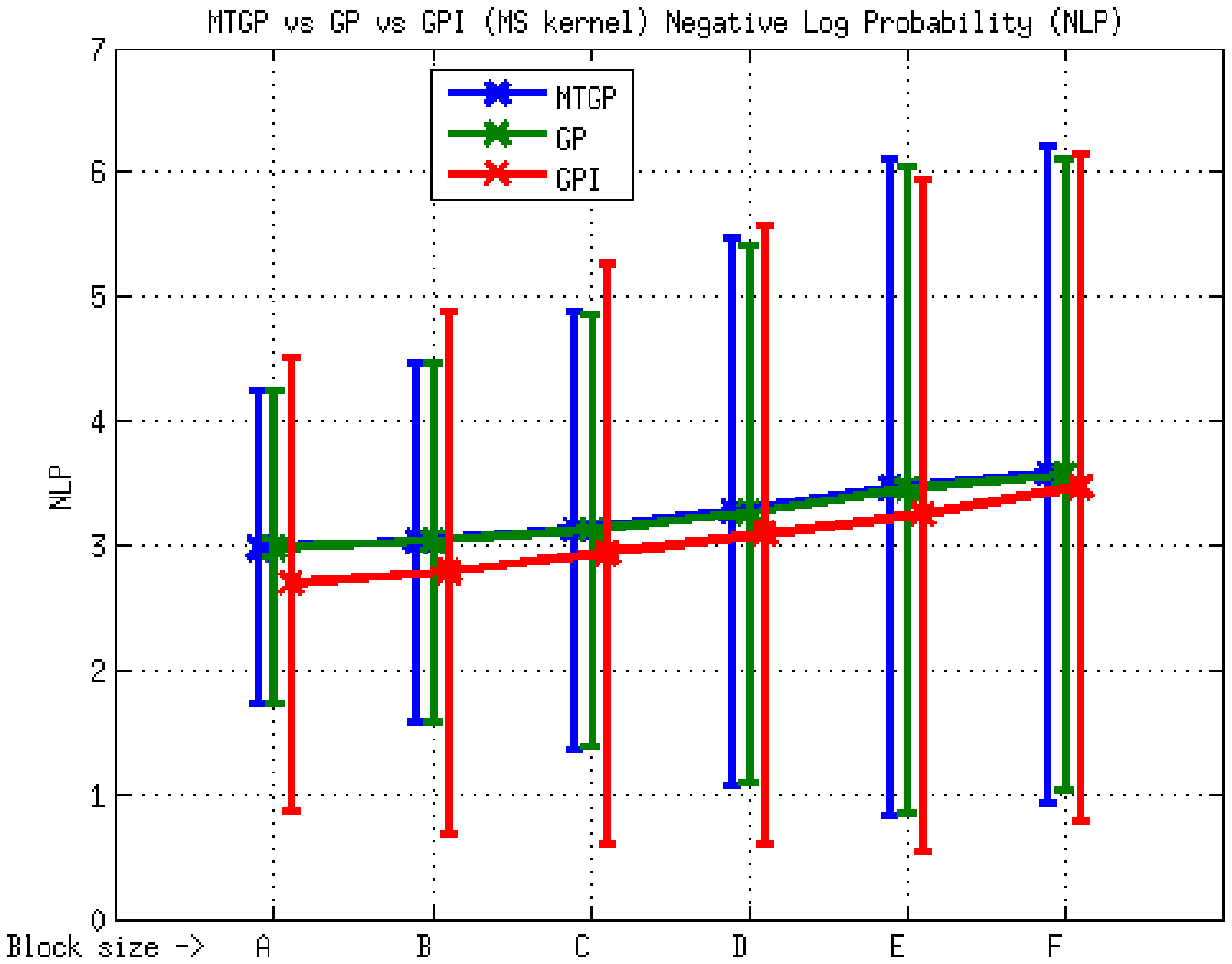}}
  \end{center}
  \caption{Element E3, MTGP vs GP vs GPI approaches, MS kernel, NLP metric. The figure above shows the average values; the one below shows the range of values obtained considering two standard deviations about the mean. Test block sizes (m) - A (22 x 11 x 2), B (44 x 22 x 4), C (84 x 45 x 9), D (174 x 89 x 18), E (348 x 177 x 35) and F (696 x 353 x 70).}
  \label{fig:e3_mtgp_gp_gpi_ms_nlp}
\end{figure}


\begin{thebibliography}{30}
  
\bibitem{vasudevan_jfr2009} S. Vasudevan, F. Ramos, E. Nettleton and H. Durrant-Whyte, ``Gaussian Process Modeling of Large Scale Terrain'', Journal of Field Robotics, volume 26(10), pages 812--840, 2009.

\bibitem{rasmussen2006} C.E. Rasmussen and C.K.I. Williams, ``Gaussian Processes for Machine Learning'', MIT Press, 2006.

\bibitem{matheron1963} G. Matheron, ``Principles of Geostatistics'', Economic Geology, volume 58, pages 1246--1266, 1963.

\bibitem{ebeltagy2001} M.A. El-Beltagy and W.A. Wright, ``Gaussian processes for model fusion'', in proc. of the International Conference on Artificial Neural Networks (ICANN), 2001.
  
\bibitem{msmith2005} R. Murray-Smith and B.A. Pearlmutter, ``Transformations of Gaussian Process priors'', chapter in Deterministic and Statistical Methods in Machine Learning, Lecture Notes in AI (LNAI) 3635, pages 110--123, Springer-Verlag, 2005.

\bibitem{girolami2006} M. Girolami, ``Bayesian data fusion with Gaussian process priors: An application to protein fold recognition'', in Workshop on Probabilistic Modeling and Machine Learning in Structural and Systems Biology (PMSB), 2006.

\bibitem{reece2011} S. Reece, S. Roberts, D. Nicholson and C. Lloyd, ``Determining intent using hard/soft data and gaussian process
  classifiers'', in proc. of the 14th International Conference on Information Fusion (FUSION), 2011.

\bibitem{vasudevan2012} S. Vasudevan, ``Data fusion using gaussian processes'', Elsevier Journal of Robotics and Autonomous Systems, Volume 60, Issue 12, December 2012, Pages 1528–1544.

\bibitem{thompson2008} D.R. Thompson and D. Wettergreen, ``Intelligent maps for autonomous kilometer-scale science survey'', in proc. of the International Symposium on Artificial Intelligence, Robotics and Automation in Space (i-SAIRAS), 2008.

\bibitem{dragiev2011} S. Dragiev, M. Toussaint and M. Gienger, ``Gaussian process implicit surfaces for shape estimation and grasping'', in proc. of the IEEE International Conference on Robotics and Automation (ICRA), pages 2845--2850, 2011.

\bibitem{vasudevan_icra2010} S. Vasudevan, F. Ramos, E. Nettleton and H. Durrant-Whyte, ``Heteroscedastic Gaussian processes for data fusion in large scale terrain modeling'', in proc. of the International Conference for Robotics and Automation (ICRA), 2010.

\bibitem{vasudevan_iros2010} S. Vasudevan, F. Ramos, E. Nettleton and H. Durrant-Whyte, ``Large-scale terrain modeling from multiple sensors with dependent Gaussian processes'', in the proc. of the IEEE/RSJ International Conference on Intelligent Robots and Systems (IROS), Taipei, October 2010.

\bibitem{vasudevan_icra2011} S. Vasudevan, F. Ramos, E. Nettleton and H. Durrant-Whyte, ``Non-stationary dependent Gaussian processes for data fusion in large scale terrain modeling'', in the proc. of the IEEE International Conference on Robotics and Automation (ICRA), Shanghai, China, 2011.

\bibitem{goldberg1998} P.W. Goldberg, C.K.I. Williams, and C.M. Bishop, ``Regression with Input-dependent Noise: A Gaussian Process Treatment'', in M.I. Jordan, M.J. Kearns, S.A. Solla and L. Erlbaum (editors), Advances in Neural Information Processing Systems (NIPS) 10, MIT Press, Cambridge, MA, 1998.

\bibitem{wabha2004} M. Yuan and G. Wabha, ``Doubly Penalized Likelihood Estimator in Heteroscedastic Regression'', Technical report, Department of Statistics, University of Wisconsin, Madison, USA, 2004.
  
\bibitem{le2005} Q.V. Le, A.J. Smola, and S. Canu, ``Heteroscedastic Gaussian Process Regression'', in proc. of the International Conference on Machine Learning (ICML), 2005.

\bibitem{kersting2007} K. Kersting, C. Plagemann, P. Pfaff and W. Burgard, ``Most Likely Heteroscedastic Gaussian Process Regression'', in proc. of the International Conference on Machine Learning (ICML), 2007.

\bibitem{Bonilla2007} E. Bonilla, K.M. Chai and C. Williams, ``Multi-task Gaussian process prediction'', in J.C. Platt, D. Koller, Y. Singer, and S. Roweis (editors), Advances in Neural Information Processing Systems (NIPS) 20, pages 153--160, MIT Press, Cambridge, MA, 2007.

\bibitem{Boyle2004} P. Boyle and M. Frean, ``Dependent Gaussian processes'', in L. Saul, Y. Weiss and L. Bottou (editors), Advances in Neural Information Processing Systems (NIPS) 17, pages 217--224, MIT Press, Cambridge, MA, 2004.

\bibitem{Higdon2002} D. Higdon, ``Space and Space-Time Modeling Using Process Convolutions'', chapter in Quantitative Methods for Current Environmental Issues, pages 37--54, Springer, 2002.

\bibitem{wackernagel2003} H. Wackernagel, ``Multivariate geostatistics: an introduction with applications'', Springer, 2003.

\bibitem{melkumyan2011} A. Melkumyan and F. Ramos, ``Multi-Kernel Gaussian Processes'', in proc. of the International Joint Conference on Artificial Intelligence (IJCAI), 2011.

\bibitem{melkumyan2009} A. Melkumyan and F. Ramos, ``A sparse covariance function for exact gaussian process inference in large data sets'', in proc. of the International Joint Conferences on Artificial Intelligence (IJCAI), volume 21, pages 1936--1942, 2009.

\bibitem{alvarez2012} M. Alvarez, L. Rosasco and N.D. Lawrence, ``Kernels for Vector-Valued Functions: A Review'', in Foundations and Trends in Machine Learning, pages 195--266, 2012.
  
\bibitem{Neal1996} R.M. Neal, ``Bayesian Learning for Neural Networks'', Lecture Notes in Statistics 118. Springer, New York, 1996.

\bibitem{Williams1998a} C.K.I. Williams, ``Computation with infinite neural networks'', Neural Computation, volume 10(5), pages 1203--1216, 1998.

\bibitem{Williams1998b} C.K.I. Williams, ``Prediction with Gaussian processes: From linear regression to linear prediction and beyond'', in M. Jordan (editor), Learning in Graphical Models, pages 599--622, Springer, 1998.

\bibitem{Hornik1993} K. Hornik, ``Some new results on neural network approximation'', Neural Networks, volume 6(8), pages 1069--1072, 1993.
  
\bibitem{alvarez2008} M. Alvarez and N.D. Lawrence, ``Sparse convolved gaussian processes for multi-output regression'', in D. Koller, D. Schuurmans, Y. Bengio and L. Bottou (editors), Advances in Neural Information Processing Systems (NIPS) 21, pages 57--64, 2009.
  
\bibitem{kohavi1995study} R. Kohavi, ``A study of cross-validation and bootstrap for accuracy estimation and model selection'', in proc. of the International Joint Conferences on Artificial Intelligence (IJCAI), volume 14, pages 1137--1145, 1995.

\end{thebibliography}
\end{document}